\documentclass{article}

\usepackage[english]{babel}
\usepackage{tikz}
\usepackage{graphicx}
\usepackage{comment}
\usepackage{caption}
\usepackage{tikz}
\usetikzlibrary{positioning}
\usepackage{authblk}
\usepackage{mathrsfs} 
\usepackage[letterpaper,top=2cm,bottom=2cm,left=3cm,right=3cm,marginparwidth=1.75cm]{geometry}

\usepackage{amsmath,amssymb,amsthm,mathtools}
\usepackage{multirow}
\usepackage{graphicx}
\usepackage[colorlinks=true, allcolors=blue]{hyperref}
\usepackage{nomencl}
\makenomenclature

\newcommand{\yadi}{\nomenclature}

\newtheorem{theorem}{Theorem}[section]

\newtheorem{corollary}{Corollary}[section]
\newtheorem{lemma}{Lemma}[section]
\newtheorem{proposition}{Proposition}[section]
\theoremstyle{definition}
\newtheorem{definition}{Definition}[section]
\newtheorem{example}{Example}[section]

\theoremstyle{remark}
\newtheorem{remark}{Remark}[section]
\DeclarePairedDelimiter{\norm}{\lVert}{\rVert}

\providecommand{\argmax}{\operatorname*{arg\ max}}

\newcommand\gattha[2]{\genfrac{}{}{0pt}{}{#1}{#2}}

\newcommand{\be}{\begin{equation}}
\newcommand{\ee}{\end{equation}}
\newcommand{\ba}{\begin{aligned}}
\newcommand{\ea}{\end{aligned}}
\newcommand{\ben}{\begin{enumerate}}
\newcommand{\een}{\end{enumerate}}
\newcommand{\bit}{\begin{itemize}}
\newcommand{\eit}{\end{itemize}}

\usepackage{multicol}
\usepackage{amsmath}
\usepackage{amsfonts}
\usepackage{amssymb}
\usepackage{graphicx}
\usepackage{color}
\usepackage[utf8]{inputenc}
\usepackage{amsfonts}
\usepackage{color}
\usepackage[normalem]{ulem}
\usepackage{nomencl}
\usepackage{tikz-cd}
\usepackage{colortbl}
\def\bhag#1{\noindent
\setcounter{equation}{0}
\section{#1}
}

\def\HH{{\mathbb H}}
\def\NN{{\mathbb N}}
\def\RR{{\mathbb R}}
\def\CC{{\mathbb C}}
\def\ZZ{{\mathbb Z}}
\def\SS{{\mathbb S}}

\def\TT{\mathbb T}

\def\a{\alpha}
\def\bs#1{{\boldsymbol{#1}}}
\def\x{\mathbf{x}}
\def\k{\mathbf{k}}
\def\y{\mathbf{y}}

\def\w{\mathbf{w}}

\def\m{\mathfrak{m}}

\def\O{{\cal O}}
\def\P{{\cal P}}

\def\C{{\mathcal C}}

\def\YY{\mathbb{Y}}

\def\ip#1#2{{\langle {#1}, {#2}\rangle}}

\def\node#1#2{x_{{#1},{#2}}}

\def\be{\begin{equation}}
\def\ee{\end{equation}}
\def\bea{\begin{eqnarray}}
\def\eea{\end{eqnarray}}
\def\eref#1{(\ref{#1})}
\def\disp{\displaystyle}

\def\binom#1#2{\scriptstyle{\left(\!\!\begin{array}{c}{#1}\\{#2}
\end{array}\!\!\right)}}
\def\donchitre#1#2{\vskip 6.5cm\noindent
\parbox[t]{1in}{\special{eps:#1.eps x=6.5cm y=5.5cm}}
\hbox to 7cm{}\parbox[t]{0.0cm}{\special{eps:#2.eps x=6.5cm y=5.5cm}}}

\def\tn{|\!|\!|}
\def\XX{{\mathbb X}}
\def\BB{{\mathbb B}}

\def\bs#1{{\boldsymbol{#1}}}

\def\gs{\gtrsim}
\def\ls{\lesssim}
\def\dist{\mathsf{dist }}

\title{\textbf{An Approximation Theory Perspective on Machine Learning}}

\author[1]{Hrushikesh N. Mhaskar}
\author[1]{Efstratios Tsoukanis}
\author[2]{Ameya D. Jagtap\thanks{Corresponding author: Ameya D. Jagtap (ajagtap@wpi.edu, ameyadjagtap@gmail.com)}}

\affil[1]{\textit{\small{Institute of Mathematical Sciences, Claremont Graduate University, CA 91711, USA}}}
\affil[2]{\textit{\small{Aerospace Engineering Department, Worcester Polytechnic Institute, Worcester, MA 01609, USA}}}

\begin{document}
\maketitle

\begin{abstract}
A central problem in machine learning is often  formulated as follows: Given a dataset\\
 $\{(x_j, y_j)\}_{j=1}^M$, which is a sample drawn from an unknown probability distribution, the goal is to construct a functional model $f$ such that $f(x) \approx y$ for any $(x, y)$ drawn from the same distribution. 
Neural networks and kernel-based methods are commonly employed for this task due to their capacity for fast and parallel computation.
 The approximation capabilities, or expressive power, of these methods have been extensively studied over the past 35 years. 
 In this paper, we will present examples of key ideas in this area found in the literature. 
  We will discuss emerging trends in machine learning including the role of shallow/deep networks, approximation on manifolds,  physics-informed neural surrogates, neural operators, and  transformer architectures.
 Despite function approximation being a fundamental problem in machine learning, approximation theory does not play a central role in the theoretical foundations of the field. 
 One unfortunate consequence of this disconnect is that it is often unclear how well trained models will generalize to unseen or unlabeled data. 
 In this review, we examine some of the shortcomings of the current machine learning framework and explore the reasons for the gap between approximation theory and machine learning practice.
We will then  review some of recent work that achieves function approximation on unknown manifolds without the need to learn specific manifold features, such as the eigen-decomposition of the Laplace-Beltrami operator or atlas construction. 
In many machine learning problems, particularly classification tasks, the labels $y_j$ are drawn from a finite set of values. 
 We summarize another recent paper that establishes a deep connection between signal separation problems and classification problems, proposing that classification tasks should be approached as instances of signal separation. 
We conclude by identifying several open research problems that warrant further investigation.
\end{abstract}

\pagebreak

\tableofcontents

\section*{List of Symbols}
\begin{list}{}{%
  \setlength{\itemsep}{6pt}
  \setlength{\parsep}{0pt}
  \setlength{\leftmargin}{3cm}
  \setlength{\labelwidth}{2.7cm}
  \setlength{\labelsep}{0.3cm}
  \renewcommand{\makelabel}[1]{\raggedright #1}
}
\item[$\mathbb{B}(x,r),\ \mathbb{B}(X,r)$] Ball, tubular neighborhood
\item[$\dist(f,\Pi_n)$] degree of approximation, \eqref{eq:degapprox}
\item[$\mathcal{D}$] decoder operator
\item[$\mathcal{D}_{q,d}$] Decoder, \eqref{eq:encoder_decoder}
\item[$\mathcal{E}$] Encoder operator
\item[$\mathcal{E}_{q,d}$] Encoder, \eqref{eq:encoder_decoder}
\item[$\mathcal{V}(G)$] variation space for kernel $G$
\item[$\mathcal{V}_N(G)$] kernel networks with kernel $G$ and $N$ neurons
\item[$\mathfrak{A}_\gamma$] Approximation space
\item[$\mathfrak{W}_r$] See Section \ref{bhag:abstracttheory}
\item[$\mathfrak{X}$] Banach space
\item[$|\nu|$] total variation of measure $\nu$
\item[$\mu,\ \nu,\ \tau$] generic measures
\item[$\mu^*$] distinguished measure
\item[$\mu^*_q$] Volume measure on $\mathbb{S}^q$, normalized
\item[$\omega_q$] Volume of $\mathbb{S}^q$
\item[$\phi$] activation function, locally supported bump function
\item[$\Phi_n$] Localized reconstruction kernel
\item[$\Pi_n$] hypothesis spaces
\item[$\Pi_n^q$] trigonometric/algebraic/spherical polynomials of degree $< n$ on $q$ dimensional spaces
\item[$\Psi_j,\ \Psi_j^*,\ \tilde{\Psi}_j$] Localized analysis kernels
\item[$\rho$] Metric, geodesic distance
\item[$\sigma_n(f)$] reconstruction operators in different contexts
\item[$\mathbb{S}^q$] Euclidean unit sphere embedded in $\mathbb{R}^{q+1}$
\item[$\tau_j,\ \tau_j^*$] frame operators in different contexts
\item[$\|\nu\|$] Marcinkiewicz-Zygmund norm of $\nu$, Definitions \ref{def:mzdef}, \ref{def:mz-measure}
\item[$\mathbb{T}$] quotient space $\mathbb{R}/(2\pi\mathbb{Z})$
\item[$\mathbb{X}$] Data space, manifold
\item[$\{\lambda_k\}$] Sequence simulating eigenvalues
\item[$\{\phi_k\}$] Sequence of orthonormal functions, simulating eigenfunctions
\item[$B_n$] Constant in Bernstein-Lipschitz inequality, Bernstein inequality
\item[$f_0$] Density of the data distribution with respect to $\mu^*$
\item[$K(\mathfrak{X},\mathfrak{W}_r;f,\delta)$] $K$-functional
\item[$p_{\ell,q}$] Orthonormalized ultraspherical polynomial of degree $\ell$, Definitions \ref{eq:sphaddformula}, \ref{eq:ultraorthogonal}
\item[$T_\ell$] Chebyshev polynomials
\item[$W_\gamma$] Smoothness class
\item[$Y_{\ell,k}$] Orthonormalized spherical harmonics
\end{list}

\section{Introduction}\label{bhag:intro}

Machine learning is increasingly influencing various aspects of modern life, from robotic vacuum cleaners to lunar landings, from life-saving surgical procedures to drones targeting enemy assets.
 The core objective of this field is to replicate and surpass human learning capabilities. 
 Human learning typically involves observing multiple examples and associating them with labels or predictions.
  For instance, we observe various objects around us, such as flowers, birds,  cats, dogs, and the moon. 
  Even without external guidance, we can differentiate between these objects, though occasionally with some confusion. 
  Later, someone teaches us the names of these objects. 
  The process of recognizing them independently is referred to as \textbf{unsupervised learning}, while the process of assigning labels to these objects is termed \textbf{supervised learning}.
   \textbf{Generalization} refers to the ability to correctly label objects that have not been seen before but are similar to those previously encountered.

\begin{figure}[h!]
\begin{center}
\includegraphics[scale=0.19, clip=true]{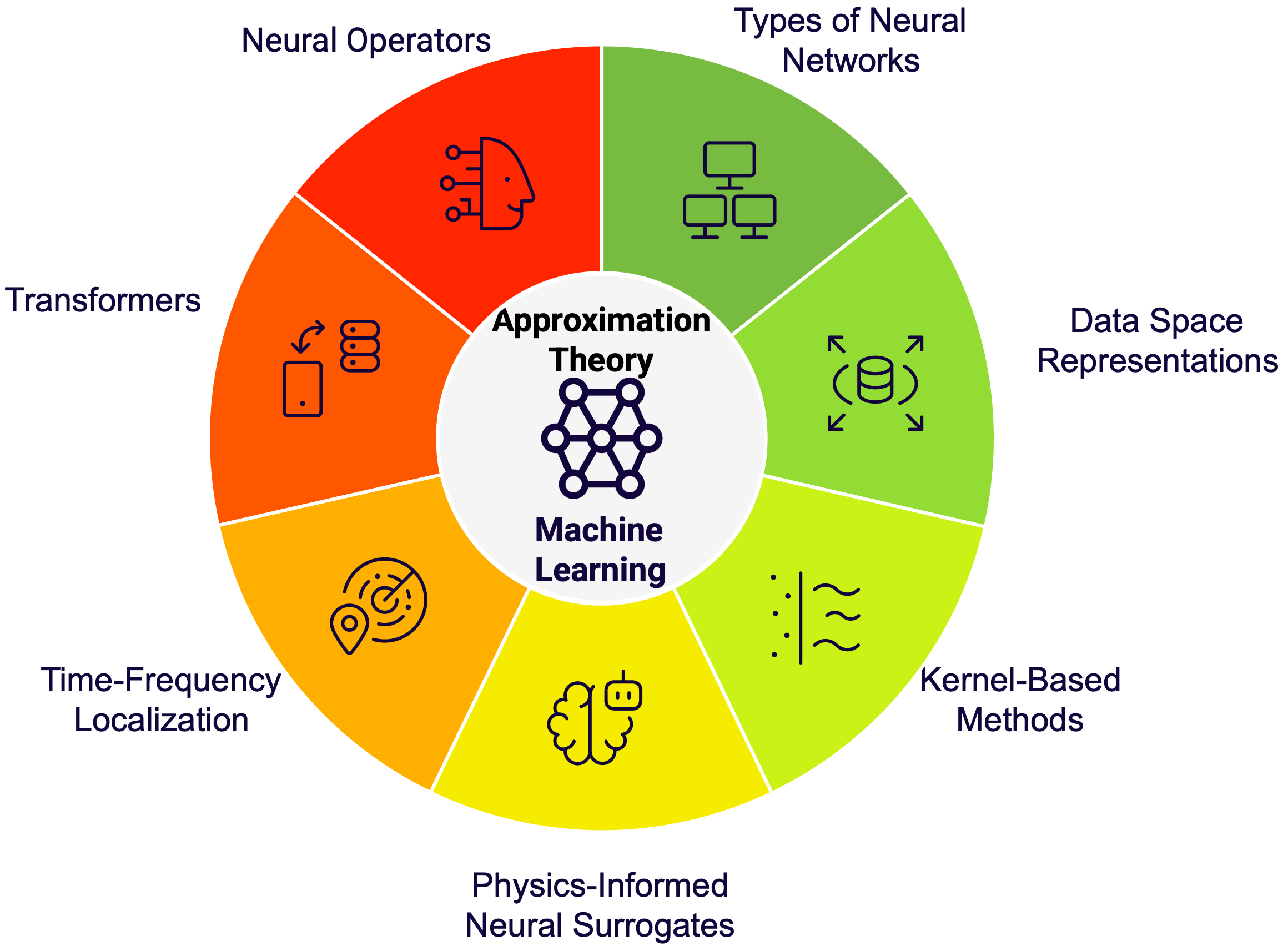} 
\end{center}
\caption{Schematic representation: Bridging approximation theory and machine learning across various fields.}
\label{fig:Schem}
\end{figure}

More formally, the central problem in machine learning can be described as follows. There exists an unknown probability distribution $\tau$ from which all objects of interest $x$ and their corresponding labels $y$ are drawn. In practice, this distribution is not known. Instead, we observe a set of samples $\{(x_j, y_j)\}_{j=1}^M$ (referred to as the \textbf{training data}) for some integer $M \geq 1$, where each $x_j$ represents a data point, and $y_j$ is the label associated with $x_j$. The challenge is to generalize from these examples in order to predict the label $y$ for a new data point $x$, typically one that is not part of the observed sample set.

Mathematically, this problem can be formulated as follows. Let $ f(x) = \mathbb{E}_\tau(y|x)$, where $ f$ is commonly referred to as the \textbf{target function}. The objective is to approximate $f$ based on the given training data. This can be expressed as $y_j = f(x_j) + \epsilon_j$, where $ \epsilon_j$ represents realizations of an unknown random variable with mean zero.
As an illustrative example, consider the well-known MNIST dataset, which consists of $M = 50,000$ training examples of $ 28 \times 28$ pixel images, each representing a handwritten digit between 0 and 9. Each image $x_j$ can be viewed as a vector in $\mathbb{R}^{784}$, and the corresponding label $y_j$ indicates the digit depicted in the image. Additionally, the dataset includes 10,000 other images, and the goal is to predict the digit contained in each of these new images.

In academic research, the problem is referred to as \textbf{classification} when the set of values that \(y\) can assume is finite. In contrast, when the set of these values forms a continuum, the problem is known as \textbf{regression}. For example, consider the task of observing the highest daily temperature, $y_j$ at a specific location,  on day $j$ over the past year. Assuming that the temperatures are measured every 15 minutes, the vector $x_j$ of temperatures during the day $j$ is a $96$ dimensional vector, there being $365$ data points of the form $(x_j,y_j)\in\RR^{96}\times\RR$.   Based on this training data, we may seek to predict the highest temperature for the following day at the same location during another year. Naturally, this problem can be generalized to account for changing locations as well. By incorporating the spatial coordinates of point $j$, we extend the feature vector to $x_j \in \mathbb{R}^{96} \times [0, 2\pi] \times [0, \pi]$, where the additional dimensions represent spatial coordinates on a domain, allowing predictions for varying locations as well as over time.
This generalization highlights the flexibility of regression models in handling both temporal and spatial variables, making them applicable to a broad range of prediction tasks.
The mathematical description of both the problems is the same; e.g., if we wish to do a $K$-class classification, i.e., $y\in \{1,\cdots,K\}$ for some integer $K\ge 2$, we define
$$
\chi_j(x)=\begin{cases}
1, &\mbox{if the label of $x$ is $j$},\\
0, &\mbox{otherwise}.
\end{cases}
$$
Then the label function $f$ can be defined by
$$
f(x)=\argmax_j \chi_j(x).
$$

Of course, there are many books, \cite{goodfellow2016deep, murphy2022probabilistic, bartlett, watanabe2009algebraic, abu2012learning, bishop2006pattern,vapnik2013nature} to point out a few, on machine learning, and each one of them makes a reference to function approximation.
However, they do not deal with approximation theory in a deep manner, referring at most to a straightforward least square fitting. 
For example, many of these books deal with polynomials expressed in terms of monomials or at best Chebyshev polynomials.
The book of Cucker and Zhou \cite{zhoubk_learning}  deals exclusively with kernel based methods in machine learning. 
There are some survey articles as well \cite{pinkus1999approximation, indiapap}, which are rather outdated by now. 
These references also do not give a background in approximation theory and explore how it can be used fruitfully for machine learning applications.

Our survey tries to bridge this gap. In this study, we will provide a detailed discussion on the connection of approximation theory with various advanced topics in scientific machine learning and neural network-based methodologies. Specifically, we will explore different types of neural networks, including shallow networks, deep networks, and $ReLU^{\gamma}$ networks, highlighting their structural properties and approximation capabilities. Additionally, we will examine data space representations, with a particular emphasis on manifold learning techniques for dimensionality reduction and data-driven modeling. The discussion will further encompass kernel-based methods, operator learning frameworks, and physics-informed neural operators, which integrate governing physical laws into neural network architectures to enhance predictive accuracy and generalizability. Moreover, we will analyze time-frequency localization techniques and the role of transformers in scientific computing and complex system modeling. A schematic representation of these topics is provided in Figure \ref{fig:Schem}.

\bhag{Traditional role of approximation theory}\label{bhag:role}

The objective of supervised machine learning is to construct a model $P$ that approximates the unknown target function $f$. Commonly employed models include shallow and deep neural networks, radial and elliptic basis function networks, as well as general kernel-based methods. The collection of such models constitutes the \textbf{hypothesis space}. A central theoretical challenge is to assess how well the model generalizes to unseen data. The traditional approach to addressing this involves defining a \textbf{loss functional} $\mathcal{L}$, which quantifies the error between the model's prediction \(P(x)\) for an input $x$ and the corresponding true output $y$. Several loss functionals are commonly used in practice, as illustrated in Table~\ref{tab:loss}, each designed to capture different aspects of prediction accuracy depending on the problem at hand.
\begin{table}[ht]
\centering
\begin{tabular}{|c|c|c|c|c|}
\hline
\textbf{Square loss} & \textbf{Exponential loss} & \textbf{0-1 loss} &\textbf{Logistic loss}  & \textbf{Hinge loss} \\
\hline
$|P(x)-y|^2$ & $\exp(-P(x)y)$& $\chi_{\{P(x)\not=y\}}$  & $\log(c+\exp(-P(x)y)$&$|1-P(x)y|$ \\
\hline
\end{tabular}
\caption{Several loss functionals are commonly used in machine learning, with the square loss being particularly suited for regression problems, while other functionals are more appropriate for classification tasks.}
\label{tab:loss}
\end{table}

The \textbf{generalization error} of $P$ is defined by 
\be\label{eq:generr}
\mbox{Generalization error}=\mathbb{E}_\tau(\mathcal{L}(P(x),y)),
\ee While we will later discuss whether this measurement is too simplistic for evaluating generalization capabilities, it remains the traditional method for assessing model performance. Ideally, given a hypothesis space $\mathcal{H}$, one aims to identify $P^\# \in \mathcal{H}$ that minimizes 
\be
\mathbb{E}_\tau(\mathcal{L}(P(x), y)).
\ee 
However, achieving this is generally infeasible because the underlying data distribution $\tau$ is unknown; we only have access to a finite set of samples $\{(x_j, y_j)\}_{j=1}^M$ drawn from $\tau$. Consequently, the focus shifts to minimizing the \textbf{empirical risk} of $P$, defined as:
\begin{equation}\label{emprisk}
\text{Empirical Risk} = \frac{1}{M} \sum_{j=1}^M \mathcal{L}(P(x_j), y_j).
\end{equation} 
The empirical risk offers a practical framework for model optimization, enabling performance evaluation based on observed data. However, during this process, it is possible that the minimizer \( P \) of the empirical risk satisfies \( P(x_j) = y_j \) for \( j = 1, \dots, M \), a phenomenon referred to as \textbf{memorization}. Models that exhibit memorization typically do not generalize well, meaning their performance on unseen test data is often poor. To address this, regularization terms are introduced into the minimization of the empirical risk \eqref{emprisk} to improve generalization. A significant portion of research on machine learning theory addresses efficient optimization methods, focusing on minimizing error surfaces, understanding the nature of these surfaces, and analyzing their local minima, saddle points, and other features. Another key research area investigates the gap between empirical risk and generalization error, as well as the behavior of their respective minimizers. Efforts have also been made to develop algorithms that intentionally produce suboptimal solutions for the optimization problem outlined in \eqref{emprisk} \cite{mukherjee2006learning}. More recently, studies have demonstrated that memorization can align with good generalization when a sufficiently large number of parameters are allowed in the hypothesis space \cite{radhakrishnan2024mechanism}. The body of literature surrounding these topics is extensive but lies outside the scope of our current work.

To establish the empirical risk minimization problem, it is necessary to first define a suitable hypothesis space. Traditionally, approximation theory plays a role in providing insight into this setup. We, therefore, consider a sequence of hypothesis spaces $\{\Pi_n\}$, where $n$ denotes the complexity of the space, such as the number of parameters in a model within $\Pi_n$. In this context, approximation theory is primarily concerned with estimating the minimal approximation error $\|f - P\|$ for all $P \in \Pi_n$ as a function of $n$. We demonstrate the link between approximation error and generalization error using the square loss.
Remembering that $f(x)=\mathbb{E}_\tau(y|x)$, it is not difficult to see that for any model $P$,
\be\label{eq:biasvar}
\int |y-P(x)|^2d\tau(x,y)=\underbrace{\int |y-f(x)|^2d\tau(x,y)}_{\mbox{variance}} +\underbrace{\int |f(x)-P(x)|^2d\mu^*(x)}_{\mbox{bias}}.
\ee
where $\mu^*$ is the marginal distribution of $x$.
Thus, in this case,  the minimizer of the bias (approximation error) is the same as the minimizer of the generalization error. This may not be the case for other loss functionals.

An inherent challenge in this framework, regardless of the chosen loss functional, is the well-known \textbf{bias-variance trade-off}. Achieving a good approximation to $f$ requires selecting a high-complexity hypothesis space, meaning the model $P$ must depend on a large number of parameters. However, this increased complexity also intensifies the optimization difficulty within the learning process, possibly leading to an increased gap between the generalization error and empirical risk.
 While advanced theoretical developments address these optimization challenges, they lie beyond the scope of this paper.

\bhag{Primer on approximation theory}\label{bhag:primer}

In approximation theory, the primary objective is to approximate a target function within a (usually) Banach space, denoted by $\mathfrak{X}$, \yadi{$\mathfrak{X}$}{Banach space} using elements from an expanding sequence of subsets of $\mathfrak{X}$, referred to as \textbf{hypothesis spaces} $\{\Pi_n\}_{n\in\mathbb{N}}$, \yadi{$\Pi_n$}{hypothesis spaces} where $\Pi_0 \subseteq \Pi_1 \subseteq \cdots$. Here, the index $ n $ typically represents the complexity of the approximation. For instance, in the approximation of a continuous function defined over $[-1,1]^q$ (where $q \geq 1$ is an integer), one might select $\Pi_n$ to be the space of all $q$-variate algebraic polynomials of total degree less than $n$.
The dimension of this space is $(\gattha{n+q-1}{q})$.
The parameter $ n $ need not necessarily be an integer. For example, if the approximants take the form $\sum_k a_k G(x, x_k) $ for certain points $ x_k $, then the complexity of the approximant can be reasonably characterized by the minimal separation $ n^{-1}$ among the points $x_k$. This minimal separation serves as a useful measure of complexity in this context.

A quantity of central interest is the so called \textbf{degree of approximation} of $f\in\mathfrak{X}$ from $\Pi_n$, defined by
\be\label{eq:degapprox}
\dist(f, \Pi_n)=\dist(\mathfrak{X}; f, \Pi_n)=\inf_{P\in \Pi_n} \|f-P\|,
\ee
where $\|\circ\|$ is the norm on $\mathfrak{X}$. \yadi{$\dist(f, \Pi_n)$}{degree of approximation, \eqref{eq:degapprox}}

The main questions in approximation theory are:
\begin{itemize}
\item (\textbf{Universal approximation property}, \textbf{density}) Does $\dist(f,\Pi_n)\to 0$ as $n\to \infty$?
\item (\textbf{Best approximation}) Does there exists a \textbf{best approximation} $P^*\in \Pi_n$ such that\\ $\dist(f, \Pi_n)=\|f-P^*\|$? 
If so, what are the properties of best approximations; e.g., uniqueness, characterization, etc.?
\item (\textbf{Smoothness classes}) What properties of $f$ ensure that $\dist(f, \Pi_n)\to 0$ at a given rate? 
For example, one seeks subspaces (\textbf{smootness classes, priors}) $W_{\gamma}$ \yadi{$W_{\gamma}$}{Smoothness class} of $\mathfrak{X}$ such that $f\in W_\gamma$ \textbf{if and only if} $\dist(f,\Pi_n)=\O(n^{-\gamma})$. 
The assertion that $f\in W_\gamma\Rightarrow \dist(f,\Pi_n)=\O(n^{-\gamma})$ is referred to as the \textbf{direct theorem}; the converse is naturally called the \textbf{converse theorem}.
\item (\textbf{Constructive approximation}) How do we construct good approximations $P_n\in \Pi_n$ and estimate $\|f-P_n\|$?
\end{itemize}

In the context of constructive approximation, the degree of approximation is defined as in \eqref{eq:degapprox}. The quantity $ \|f - P_n\|$ is referred to as the \textbf{approximation error}, while expressions such as $\mathcal{O}(n^{-\gamma})$ are known as the \textbf{rate of approximation}. Although these terms are sometimes used informally in the literature, we will adhere to this specific usage consistently throughout this paper.

\subsection{General Theory of Smoothness Classes and Construction}\label{bhag:abstracttheory}
In this section, we present a general theory addressing the problems of smoothness classes and their construction in some abstraction. This discussion is based on \cite[Chapter~7]{devlorbk}.
We will give a detailed exposition in the context of approximation of periodic funtions by trigonometric polynomials in Section~\ref{bhag:trigapprox}.\\

\noindent\textbf{Constant Convention:}\\

\textit{Throughout this discussion, $c, c_1, \cdots$ will represent generic positive constants that may depend on fixed quantities such as $\mathfrak{X}$, $\gamma$, etc. These constants may take different values in different occurrences. The notation $A \ls B$ implies $A \leq cB$, and $A \gs B$ implies $B \ls A$. Additionally, $A \sim B$ indicates $A \ls B$ and $B \ls A$.}\\

We assume that the hypothesis spaces $\Pi_n$ satisfy the following properties:
\begin{enumerate}
    \item $0 \in \Pi_n$, and $\Pi_0 = \{0\}$.
    \item $\Pi_n \subseteq \Pi_{n+1}$.
    \item $a\Pi_n = \Pi_n$ for every $a \in \RR$.
    \item $\Pi_n + \Pi_n \subseteq \Pi_{cn}$.
    \item $\bigcup_n \Pi_n$ is dense in $\mathfrak{X}$.
    \item For each $f \in \mathfrak{X}$ and $n \geq 1$, $\mathsf{dist}(f, \Pi_n) < \infty$.
\end{enumerate}
Some examples of $\Pi_n$ are: trigonometric/algebraic/spherical polynomials of degree $<n$, set of neural/RBF networks with $<n$ neurons, rational functions of numerator and denomiator degrees $<n$, etc.
If $\Pi_n$ denotes the set of all functions of the form $\sum_k a_k\exp(-|\circ-\x_k|_2^2)$, where $\x_k$'s are in $\RR^q$, $|\cdot|_2$ denotes the Euclidean norm, and $\min_{k\not=j}|\x_k-\x_j|_2 \ge 1/n$, then the property $\Pi_n+\Pi_n\subseteq \Pi_{cn}$ is not satisfied.

In the remainder of this discussion, we will assume that
$\Pi_\infty=\displaystyle \bigcup_{n>0}\Pi_n$ is dense in $\mathfrak{X}$.
The central question is as follows: 
Let $\gamma > 0$ and define
\be\label{eq:approxspace}
\mathfrak{A}_\gamma = \{f \in \mathfrak{X} : \mathsf{dist}(f, \Pi_n) \lesssim n^{-\gamma}\}.
\ee
\yadi{$\mathfrak{A}_\gamma$}{Approximation space} The goal is to determine the conditions on $f$ that are equivalent to $f \in \mathfrak{A}_\gamma$.

One way to approach this question is to select $r > \gamma$ and identify a well-defined subspace $\mathfrak{W}_r$ with $\P_\infty\subset\mathfrak{W}_r \subset \mathfrak{X}$, equipped with a seminorm $\|\cdot\|_r$, such that the following two estimates, \eqref{eq:favardineq} and \eqref{eq:berstein}, hold. \yadi{$\mathfrak{W}_r$}{See Section~\ref{bhag:abstracttheory}}

The \emph{Favard inequality} is an inequality of the form:
\be\label{eq:favardineq}
\mathsf{dist}(f, \Pi_n) \leq c_F n^{-r} \|f\|_r, \qquad f \in \mathfrak{W}_r.
\ee 

Much of the literature on machine learning examines the Favard inequality for various spaces analogous to $\mathfrak{W}_r$. A key question often raised is how to generalize these results for approximating arbitrary functions in $\mathfrak{X}$. The solution to this problem lies in the concept of $K$-functionals. 

The $K$-functional between $\mathfrak{X}$ and $\mathfrak{W}_r$ serves as a regularization criterion, defined as:\yadi{$K(\mathfrak{X}, \mathfrak{W}_r; f, \delta)$}{$K$-functional}
\be\label{eq:Kfuncdef}
K(\mathfrak{X}, \mathfrak{W}_r; f, \delta) = \inf_{g \in \mathfrak{W}_r} \{\|f - g\| + \delta^r \|g\|_r\}.
\ee

The \emph{direct theorem} states that for every $f \in \mathfrak{X}$ and $n \geq 1$,
\be\label{eq:directtheo}
\mathsf{dist}(f, \Pi_n) \leq c_F K(\mathfrak{X}, \mathfrak{W}_r; f, 1/n).
\ee

Therefore, any Favard inequality suffices to estimate the degree of approximation for an arbitrary function in $\mathfrak{X}$.

A persistent gap often exists between theory and practice, with practical performance frequently exceeding theoretical guarantees. A natural question arises: is this due to the design of clever algorithms, or does it stem from the inherent properties of functions that permit a higher rate of approximation? In approximation theory, this question is addressed by proving a so-called converse theorem. We first recall the \emph{Bernstein inequality}, which has the form
\be\label{eq:berstein}
\|P\|_r\le c_B n^r\|P\|, \qquad P\in \Pi_n.
\ee
Here, $c_F$, $c_B$ are constants as usual in $\lesssim$, but we formulate it this way to make the direct and converse theorems more precise.

The \emph{converse/inverse} theorem states that for any $\delta\in (0,1/2)$ and $n$ being the largest integer not exceeding $\log_2(1/\delta)$,
\be\label{eq:conversetheo}
K(\mathfrak{X}, \mathfrak{W}_r;f,\delta)\le 2^{3r+1}c_B\delta^r\left\{\|f\|+\sum_{k=1}^{2^n} k^{r-1}\mathsf{dist}(f, \Pi_k)\right\}.
\ee
In particular, we have the \emph{equivalence theorem}:\\
\be\label{eq:equivtheo}
\|f\|+\sup_{n\ge 0}2^{n\gamma}\mathsf{dist}(f, \Pi_{2^n})\sim \|f\|+
\sup_{n\ge 0}2^{n\gamma}K(\mathfrak{X}, \mathfrak{W}_r;f,2^{-n}).
\ee

We observe that the left-hand side expression in \eqref{eq:equivtheo} is finite if and only if $f \in \mathfrak{A}_\gamma $ and is independent of the choice of $ r > \gamma$. Consequently, the right-hand side expression is also independent of $ r $, aside from the constants involved in $\sim$. For any approximation process, the key objectives are to identify the spaces $\mathfrak{W}_r$ and to obtain an `understandable' expression for the $K$-functional. Typically, one first establishes the direct theorem and, where feasible, also the converse theorem, which is often considerably more challenging.

We now come to the second major problem in approximation theory:  \textbf{Constructive approximation}. Construct a sequence of linear operators $\sigma_n$, each $\sigma_n(f)$ \yadi{$\sigma_n(f)$}{reconstruction operators in different contexts} based on a  finite amount of information on $f$ (typically, values of $f$ or coefficients in an orthogonal expansion of $f$, with the number corresponding to the complexity of $\Pi_n$), such that $\sigma_n(f)\in \Pi_n$, and for some $\beta\in (0,1]$,
\be\label{eq:goodapprox}
\|f-\sigma_n(f)\|\lesssim \mathsf{dist}(f, \Pi_{\beta n}).
\ee
Necessarily, $\sigma_n(P)=P$ if $P\in\Pi_{\beta n}$ and
\be\label{eq:sigmaopbd}
\|\sigma_n(f)\|\lesssim \|f\|, \qquad f\in\mathfrak{X}.
\ee
In the general framework of this section, an explicit construction cannot be provided; however, we will outline how such an operator facilitates the establishment of both direct and converse theorems, as well as the so-called Littlewood-Paley expansion. Using the Favard and Bernstein inequalities, it is not difficult to show that
\be\label{eq:realization}
K(\mathfrak{X}, \mathfrak{W}_r;f,1/n)\sim \|f-\sigma_n(f)\|+n^{-r}\|\sigma_n(f)\|_r.
\ee
Let \yadi{$\tau_j$, $\tau_j^*$}{frame operators in different contexts}
\be\label{eq:tauopdef}
\tau_j(f)=\begin{cases}
\sigma_1(f), &\mbox{ if $j=0$},\\
\sigma_{2^j}(f)-\sigma_{2^{j-1}}(f), &\mbox{ if $j=1,2,\cdots$}.
\end{cases}
\ee
Obviously,
\be\label{eq:lpseries}
f=\sum_{j=0}^\infty \tau_j(f).
\ee

We conclude with the following theorem:
\begin{theorem}
\label{theo:characterization}
Let $f\in \mathfrak{X}$, $r>\gamma>0$, and the Favard and Bernstein inequalities be satisfied. 
Then the following are equivalent.
\begin{itemize}
\item[(a)] $$f\in \mathfrak{A}_\gamma.$$
\item[(b)] $$\sup_{n\ge 0}2^{n\gamma}K(\mathfrak{X}, \mathfrak{W}_r;f,2^{-n})<\infty.$$
\item[(c)] $$\sup_{n\ge 0}2^{n\gamma}\|f-\sigma_{2^n}(f)\| <\infty.$$
\item[(d)] $$\sup_{j\in\NN} 2^{j\gamma}\|\tau_j(f)\| <\infty.$$
\end{itemize}
Norm equivalences analogous to \eqref{eq:equivtheo} hold.
\end{theorem}
We note that the condition in part (a) is intrinsic to the problem. Condition (b) depends on the choice of $r$ and, therefore, requires prior knowledge of the smoothness of $f$. Conditions (c) and (d) depend on the construction of $\sigma_n$ but do not necessitate prior knowledge of smoothness of $f$. Each of these conditions can be used to define the notion of smoothness for $ f $. We favor condition (a) due to its intrinsic nature, and condition (d) for its utility in establishing estimates in harmonic analysis. The constants involved in such estimates will differ accordingly.
 
We conclude this section by discussing an additional key concept, known as widths, which offers a criterion weaker than a converse theorem. While the converse theorem demonstrates that, for any \textit{individual function}, the rate of convergence implies membership in a specific smoothness class, the width result instead addresses the worst-case error.

There are many notions of width available in the literature, e.g.,  \cite{pinkusbk, devore1989optimal, cohen2022optimal, petrova2023lipschitz}. 
Our objective is to discuss the notion of the so called curse of dimensionality which is defined in terms of widths. 
So, we will focus on the nonlinear/manifold width.

Let $K\subset \mathfrak{X}$ be a compact set. 
Any approximation process can be described as a composition of two processes. 
Let $N\ge 1$ be an integer.
The \textbf{parameter selection/information operator/encoder} is a mapping $\mathcal{E}: K\to \RR^N$. \yadi{$\mathcal{E}$}{Encoder operator} \yadi{$\mathcal{D}$}{decoder operator}
An \textbf{algorithm/reconstruction/decoder} is a mapping
$\mathcal{D} :\RR^N\to \mathfrak{X}$.

\begin{tikzpicture}[node distance=2cm, auto]
    \node (f) {$f \in K$};
    \node (P) [right of=f, node distance=4cm] {$\mathcal{E}(f) \in \mathbb{R}^N$};
    \node (A) [right of=P, node distance=5cm] {$\mathcal{D}(\mathcal{E}(f)) \in \mathfrak{X}$};
    
    \draw[->] (f) to node {Encoder: $\mathcal{E}$} (P);
    \draw[->] (P) to node {Decoder: $\mathcal{D}$} (A);
\end{tikzpicture}

For any $f\in K$, the approximation is $\mathcal{D}(\mathcal{E}(f))$. 
So, the worst case error given only the prior assumption about $f$ that $f\in K$ is defined by
$$
\mathsf{wor}(K; \mathcal{D}, \mathcal{E})=\sup_{f\in K}\|f-\mathcal{D}(\mathcal{E}(f))\|.
$$
The width of $K$ measures the absolute minimum worst case error that one expects even if one allows any clever way of designing an encoder/decoder pair. 
For technical reasons, one has to restrict the encoder to be a continuous function.
Then the width \textbf{(nonlinear/manifold) width} of $K$ is defined by
\be\label{eq:widthdef}
\mathsf{width}_N(K)=\inf_{\mathcal{D}, \mathcal{E}}\mathsf{wor}(K; \mathcal{D}, \mathcal{E}),
\ee
where the infimum is taken over \emph{all} decoders and all \emph{continuous} encoders.

An important theorem in this connection is the following theorem \cite{devore1989optimal}.
\begin{theorem}
\label{theo:width}
Let $\|\circ\|_*$ be a semi-norm defined on a subspace $\mathfrak{Y}$ of $\mathfrak{X}$ and $K=\{g\in \mathfrak{Y} : \|g\|_*\le 1\}$ be a compact subset. 
For integer $N\ge 1$, let $X_N$ be a linear subspace of $\mathfrak{X}$ with dimension $N+1$ such that the following Bernstein-type estimate holds.
\be\label{eq:widthbern}
\|P\|_*\le B_N\|P\|, \qquad P\in X_N.
\ee
Then
\be\label{eq:widthlowest}
\mathsf{width}_N(K) \ge B_N^{-1}.
\ee
\end{theorem}
Typically, we choose $K$ to be the unit ball of $\mathfrak{W}_r$. Consequently, when the sets $\Pi_n$ are linear subspaces, the Bernstein inequality also provides a result for widths. 
The distinction lies in the requirements for establishing a converse theorem. Specifically, a Bernstein inequality is needed for the particular choice of $\Pi_n$ used in a direct theorem. In contrast, to establish a lower bound on widths, it suffices to use the Bernstein inequality for any $(N+1)$-dimensional subspace $ X_N$. To the best of our knowledge, there is no known example where $X_N$ differs from one of the $\Pi_n$'s.

\subsection{Trigonometric approximation}\label{bhag:trigapprox}
We address these questions within the framework of trigonometric polynomial approximation.\\
 Let $q\ge 1$ be an integer, $\TT=\RR/(2\pi\ZZ)$. \yadi{$\TT$}{quotient space $\RR/(2\pi\ZZ)$}
 In this section, $|x|=|x\mod (2\pi)|$. For $\x\in\TT^q$, $|\x|=\max_{1\le k\le q}|x_k|$, $|\x|_p$ denote the $\ell^p$ norm of $(|x_1|,\cdots,|x_q|)$. \yadi{$|\cdot|_p$}{$\ell^p$ norm} \yadi{\|\cdot\|_p}{$L^p$ norm}
 The measure $\mu^*$ is the Lebesgue measure on $\TT^q$, normalized to be a probability measure. \yadi{$\mu^*$}{distinguished measure} \yadi{$\mu$, $\nu$, $\tau$}{generic measures}
 Let $\mathfrak{X}=X^p$ denote $L^p(\mu^*)$ if $1\le p<\infty$, and $C(\TT^q)$ if $p=\infty$. 
 
 For the space $\Pi_n$, we take \yadi{$\Pi_n^q$}{trigonometric/algebraic/spherical polynomials of degree $<n$ on $q$ dimensional spaces}
 $$
 \Pi_n^q=\mathsf{span}\{\x\mapsto e^{i\k\cdot\x} : \k\in\ZZ^q, \ |\k|_2<n\}.
 $$ 
In view of the well known Fej\'er's theorem \cite{timanbk,trigub2012fourier}, the universal approximation property holds in this context.
The existence and uniqueness  of best approximation follows from a compactness argument \cite{Singer1970BestApproximation}.  
Characterizations of these are also studied in the literature; e.g., in the case when $q=1$, $T^*\in \HH_n$ is a best approximation to $f$ if and only if there are $2n$ points  $x_1<x_2<\cdots x_{2n}$ such that $f(x_j)-T^*(x_j)=(-1)^j\|f-T^*\|$ (or $f(x_j)-T^*(x_j)=(-1)^{j+1}\|f-T^*\|$) for $j=1,\cdots, 2n$.

Let $r\ge 1$ be an integer. 
 For the space $\mathfrak{W}_r$, we take  the space $X^p_r$ of functions in $X^p$ which admit all derivatives of order $\le r$  in $X^p$. 
 The semi-norm of $X^p_r$ is defined as usual:
 $$
 \|f\|_{X^p_r}=\sum_{|\k|_1 \le r}\|D^\k f\|_p.
 $$
 Then the Favard and Bernstein  inequalities hold with the constant $c_F$ possibly dependent of $q$, and $c_B$ dependent only on $r$ \cite[Section~5.3]{timanbk}. 
 
 Next, we come to the question of the construction of a good approximation.

The Fourier coefficients of $f$ are defined by
\be\label{eq:fourcoeff}
\hat{f}(\k)=\int_{\TT^q} f(\y)\exp(-i\k\cdot\y)d\mu_q^*(\y), \qquad \k\in\ZZ^q.
\ee 
We will use a low pass filter $h$; i.e., $h\in C^\infty(\RR)$ is an even function, with $h(t)=1$ for $|t|\le 1/2$ and $h(t)=0$ if $|t|\ge 1$. 
This function is fixed in this section, and its mention is omitted from the notation. 
All constants may depend upon the choice of $h$.
We define
\be\label{eq:trigsigmadef}
\sigma_n(f)(\x)=\sum_{\k\in\ZZ^q}h\left(\frac{|\k|_2}{n}\right)\hat{f}(\k)\exp(i\k\cdot\x).
\ee
It is observed in \cite[Theorem~3.1]{eugenenevai} that
\be\label{eq:trig_good_approx}
\mathsf{dist}(X^p(\TT^q) ;f, \Pi_n^q) \le \|f-\sigma_n(f)\|_p \lesssim \mathsf{dist}(X^p(\TT^q) ;f, \Pi_{n/2}^q).
\ee

We note that with
\be\label{eq:kerneldef}
\Phi_n(\x)=\sum_{\k\in\ZZ^q}h\left(\frac{|\k|_2}{n}\right)\exp(i\k\cdot\x),
\ee
we have \yadi{$\Phi_n$}{Localized reconstruction kernel}
\be\label{eq:kernelexp}
\sigma_n(f)(\x)=\int_{\TT^q}f(\y)\Phi_n(\x-\y)d\mu_q^*(\y).
\ee
Likewise, we may write $g(t)=h(t)-h(2t)$, 
\be\label{eq:taukern}
\Psi_n(\x)=\sum_{\k\in\ZZ^q}g\left(\frac{|\k|_2}{n}\right)\exp(i\k\cdot\x),
\ee
and observe that \yadi{$\Psi_j$, $\Psi_j^*$, $\widetilde{\Psi}_j$}{Localized analysis kernels}
\be\label{eq:taukernexp}
\tau_j(f)(\x)=\int_{\TT^q}f(\y)\Psi_{2^j}(\x-\y)d\mu_q^*(\y).
\ee
Next, we define $\tilde{g}(t)=h(t/2)-h(4t)$,
\be\label{eq:psidef}
\widetilde{\Psi}_j(\x)=\sum_{\k\in\ZZ^q}\tilde{g}\left(\frac{|\k|_2}{2^j}\right)\exp(i\k\cdot\x),
\ee
and observe that $g(t)=g(t)\tilde{g}(t)$ to deduce that
\be\label{eq:taureprod}
\tau_j(f)(\x)=\int_{\TT^q}\tau_j(f)(\y)\widetilde{\Psi}_j(\x-\y)d\mu_q^*(\y).
\ee
Recalling that for any $N\ge 1$,
\begin{equation} \label{eq:fft}
\frac{1}{N^q}\sum_{\substack{\m \in \mathbb{Z}^q \\ \|\m\|_{\infty} < N}} T\left(\frac{2 \pi \m}{N}\right) = \int_{\mathbb{T}^q} T(\y) \, d\y, \qquad T \in \Pi_N^q.
\end{equation}
we deduce from \eqref{eq:taureprod} that
\be\label{eq:taureprodbis}
\tau_j(f)(\x) = \frac{1}{2^{j+3}} \sum_{\substack{\m \in \mathbb{Z}^q \\ \|\m\|_{\infty} < 2^{j+3}}} \tau_j(f)\left(\frac{2 \pi \m}{2^{j+3}}\right) \widetilde{\Psi}_j\left(\x - \frac{2 \pi \m}{2^{j+3}}\right).
\ee
Thus, we have a wavelet-like expansion
\be\label{eq:lpwaveexp}
f(\x) = \sum_{j=0}^\infty \frac{1}{2^{j+3}} \sum_{\substack{\m \in \mathbb{Z}^q \\ \|\m\|_{\infty} < 2^{j+3}}} \tau_j(f)\left(\frac{2 \pi \m}{2^{j+3}}\right) \widetilde{\Psi}_j\left(\x - \frac{2 \pi \m}{2^{j+3}}\right).
\ee
with the series converging in the sense of $X^p$.
We note that, unlike the classical Littlewood-Paley expansion, the series converges in the norm of $ X^p $, even for $p = 1$ and $p = \infty$. 

In the case $ p = \infty$, mesh-free constructions can be performed based on function evaluations, which we now discuss. Since we are interested in the construction based on function evaluations, we will now restrict $\mathfrak{X}=C(\TT^q)$; and the norm refers to the uniform norm.

Henceforth, the term `measure' will refer to a positive or signed measure with bounded total variation on $\mathbb{T}^q$. For any such measure $\nu$, $|\nu|$ will denote its total variation measure. If $\nu$ is any measure, we may define
\be\label{eq:gensigmadef}
\sigma_n(\nu;f)(\x)=\int_{\TT^q}f(\y)\Phi_n(\x-\y)d\nu(\y).
\ee

With suitable choices of the measures $\nu$, the expression $\sigma_n(\nu; f)$ can serve as a substitute for $ \sigma_n $ in the context described above. To formalize this, we introduce the following Definition - \ref{def:mzdef}.

\begin{definition}\label{def:mzdef}
Let $n\ge 1$. A measure $\nu$ will be called a \textbf{quadrature measure} of order $n$ if
\be\label{eq:quadrature}
\int_{\TT^q} P(\x)d\nu(\x)=\int_{\TT^q} P(\x)d\mu_q^*(\x), \qquad P\in \Pi_{2n}^q.
\ee
The measure $\nu$ will be called a \textbf{Marcinkiewicz-Zygmund (MZ) measure} of order $n$ if
\be\label{eq:mzineq}
\int_{\TT^q} |P(\x)|d|\nu|(\x)\le \tn\nu\tn\int_{\TT^q} |P(\x)|d\mu_q^*(\x), \qquad P\in \Pi_{2n}^q,
\ee
where $\tn\nu\tn$ is understood to be the smallest constant that works in \eqref{eq:mzineq}.
The set of all MZ quadrature measures or order $n$ will be denoted by $MZQ(n)$.
A sequence $\bs\nu=\{\nu_n\}$ of measures will be called an MZQ sequence if each $\nu_n\in MZQ(n)$ with $\tn\nu_n\tn\lesssim 1$. 
\end{definition} 
 \yadi{$\tn\nu\tn$}{Marcinkiewicz-Zygmund norm of $\nu$, Definitions~\ref{def:mzdef}, \ref{def:mz-measure}} 

If $\C\subseteq \TT^q$, the mesh norm of $\C$ is defined by
\be\label{eq:meshnormdef}
\delta(\C)=\sup_{\x\in\TT^q}\inf_{\y\in\C}|\x-\y|.
\ee
The following theorem in \cite{indiapap} is particularly interesting when $\C$ is a finite subset of $\TT^q$.
\begin{theorem}
\label{theo:mzqtheo}
There exists a constant $c_q>0$ with the following property: If $n\ge 1$ and $\C\subseteq\TT^q$ satisfies $\delta(\C)\le c_q/n$, then there exists a measure $\nu\in MZQ(n)$ supported on $\C$, with $\tn\nu\tn\lesssim 1$.
\end{theorem}
\yadi{$\mid \nu \mid$}{total variation of measure $\nu$}

Let $\bs\nu$ be an MZQ sequence. 
We define
\be\label{eq:disctaudef}
\tau_j(\bs\nu;f)=\begin{cases}
\sigma_1(\nu_1;f), &\mbox{ if $j=1$},\\
\sigma_{2^j}(\nu_{2^j};f)-\sigma_{2^{j-1}}(\nu_{2^{j-1}};f), &\mbox{ if $j=1,2,\cdots$}.
\end{cases}
\ee

Analogous to \eqref{eq:lpwaveexp}, we have the fully discrete wavelet-like expansion
\be\label{eq:supwaveexp}
f(\x) = \sum_{j=0}^\infty \frac{1}{2^{j+3}} \sum_{\substack{\m \in \mathbb{Z}^q \\ |\m| < 2^{j+3}}} \tau_j(\boldsymbol{\nu}; f)\left(\frac{2 \pi \m}{2^{j+3}}\right) \widetilde{\Psi}_j\left(\x - \frac{2 \pi \m}{2^{j+3}}\right),
\ee
with the series converging uniformly.
\begin{theorem}
\label{theo:trigcharacterization}
Let $f\in C(\TT^q)$, $r>\gamma>0$, $\bs\nu$ be an MZQ sequence.
Then the following are equivalent.
\begin{itemize}
\item[(a)] $$f\in \mathfrak{A}_\gamma.$$
\item[(b)] $$\sup_{n\ge 0}2^{n\gamma}\|f-\sigma_{2^n}(\nu_{2^n};f)\| <\infty.$$
\item[(c)] $$\sup_{j\in\NN} 2^{j\gamma}\|\tau_j(\bs\nu;f)\| <\infty.$$
\item[(d)] $$\sup_{j\in\NN}\sup_{|\m|<2^{j+3}}2^{j\gamma}\left|\tau_j(\bs\nu;f)\left(\frac{2\pi\m}{2^{j+3}}\right)\right| <\infty.
$$
\end{itemize}
Norm equivalences analogous to \eqref{eq:equivtheo} hold.
\end{theorem}

Using concentration inequalities, it can be shown that the following theorem provides a complete solution to the problem of machine learning. This result eliminates the need for any optimization, provided the data $\{x_j\}$ is sampled from the distribution $\mu_q^*$.

\begin{theorem}
\label{theo:directsoltrig}
Let $\{(x_j, y_j)\}_{j=1}^M$ be sampled from unknown distribution $\tau$, with $x_j$'s distributed according to $\mu_q^*$ (marginal distribution).
We define
$$
\widetilde{\sigma_N}(z;x)=\frac{1}{M}\sum_{j=1}^M y_j\Phi_N(x-x_j).
$$
Let $\gamma>0$, $f\in W_\gamma$, $N\gtrsim  1$,
 $M\gtrsim  N^{{q+2\gamma}}\log N$ (equivalently, $N\lesssim  (M/\log M)^{1/(q+2\gamma)})$. Then
with probability tending to $1$ as $M\to\infty$ (respectively $N\to\infty$),
\be\label{eq:directtrigest}
\sup_{x\in\TT^q}\left|\widetilde{\sigma_N}(z;x)-f(x)\right| =\O\left(\left(\frac{\log M}{M}\right)^{\gamma/(q+2\gamma)}\right).
\ee
\end{theorem} 

  Theorem~\ref{theo:directsoltrig} is a special case of Corollary~\ref{cor:ddrapprox} of  Theorem~\ref{theo:probapprox} below.
 See Remark~\ref{rem:trigcase}.

\subsection{Curse of dimensionality}\label{bhag:curse}
For the trigonometric case, the dimension of $\Pi_n^q$ is $\sim n^q$. Consequently, the Bernstein inequality for $\mathfrak{W}_r$ leads to the following estimate on the widths of the unit ball $\mathfrak{B}_r$ of $\mathfrak{W}_r$:
\be\label{eq:curseofdim}
\mathsf{width}_N(\mathfrak{B}_r)\gs N^{-r/q}.
\ee
Equivalently, this implies that in order to obtain an accuracy of $\epsilon$ in the approximation of $f$, given only that $f\in \mathfrak{B}_r$, one requires at least $\Omega(\epsilon^{-q/r})$ evaluations of $f$.
Since this quantity tends to infinity exponentially fast in the dimension of the input space, \eqref{eq:curseofdim} is referred to as \textbf{curse of dimensionality}. 
Similar conclusions arise in many other contexts, especially when approximating smooth functions in terms of their derivatives (and corresponding $K$-functionals). 
Avoiding the curse of dimensionality is an active topic of research in machine learning. 
We would like to make some salient points in this connection.
\begin{itemize}

\item The curse of dimensionality is independent of the approximation process used. 
Instead, it arises entirely from the process of parameter selection.
In particular, within the context of machine learning, one must handle function evaluations as parameter selection, and assume $\mathfrak{X}$ to allow this process to be continuous.  
The specific choice of approximation method, whether it is neural networks with a small number of neurons, radial basis function (RBF) networks, deep networks, or any other form of approximation, is inconsequential.
\item Every smoothness class is associated with a width, but it may not necessarily be subject to the `curse of dimensionality'. If the sole assumption about the target function is its membership in a smoothness class that is afflicted by the curse of dimensionality, then this limitation is unavoidable, regardless of how ingenious the approximation methods may be.
The only way to circumvent the curse of dimensionality is to consider a different smoothness class where the \textit{curse} is inherently absent, or to impose additional structural assumptions on the target functions, such as a manifold structure or a compositional structure.

\end{itemize}

\bhag{Approximation by shallow networks}\label{bhag:shallow}
A neural network with $N$ neurons is a function of the form $x\mapsto \sum_{k=1}^N a_k\phi(w_k\cdot x +b_k)$, for $x$ and $w_k$'s in some (same) Euclidean space, $a_k$, $b_k$ are real numbers and $\phi:\RR\to\RR$ is the so called activation function. 
A RBF network with $N$ neurons is a function of the form $x\mapsto\sum_{k=1}^N a_k\phi(|x-w_k|)$ with the same interpretations as above. \yadi{$\phi$}{activation function, locally supported bump function}

\subsection{Universal approximation}\label{bhag:universal}
The \textbf{universal approximation property} for a class $\mathcal{G}$ of networks (neural or RBF) means that for any integer $q\ge 1$, any compact subset $K\subset \RR^q$,  any continuous $f: K\to\RR$, and for any $\epsilon>0$, there exists $G\in\mathcal{G}$ such that $\|f-G\|_{\infty, K}<\epsilon$.
In view of the Tietze extension theorem, $f$ can be extended to $\RR^q$ as a continuous function. 
By rescaling, we may assume that $K\subseteq [-1,1]^q$, so that it is enough to assume that $K=[-1,1]^q$.
Alternately, we may multiply $f$ (extended to $\RR^q$) by a $C^\infty$ function which is equal to $1$ on $[-1,1]^q$ and equal to $0$ outside of $[-\pi/2,\pi/2]^q$, and then extend the resulting function as a $2\pi$-periodic function. 
In this way, one can use the theory of trigonometric approximation for the study of the universal approximation property.

For example, Funahashi \cite{funahashi1989} employed a discretization of an integral formula from \cite{irie1988} to establish the universal approximation property for certain sigmoidal activation functions $\sigma$. A related result was proven by Hornik, Stinchcombe, and White \cite{hornik1989} using the Stone–Weierstrass theorem, while Cybenko \cite{cybenko1989} demonstrated a similar theorem by leveraging the Hahn–Banach and Riesz Representation theorems. Chui and Li \cite{chuili1992} provided a constructive proof based on approximation via ridge functions, with an accompanying algorithm for implementation detailed in a subsequent work \cite{chui1993realization}.  
We refer to \cite{pinkus1999approximation, chui_deep} for further information on the history of this subject.
In this context, we focus on discussing two theorems of a more general nature.

It is proved in \cite{corominas1954condiciones} that if $\phi\in C^\infty(\RR)$ is not a polynomial, then there exists $b\in\RR$ for which $\phi^{(m)}(b)\not=0$ for any integer $m\ge 0$. 
Hence, for any multi-integer $\k\in\ZZ_+^q$ and $\x\in\RR^q$, we have
\be\label{eq:monomialnet}
\x^k=\frac{1}{\phi^{|\k|}(b)}\frac{\partial^{|\k|}}{\partial^\k \w}\phi(\w\cdot\x+b)\Bigg|_{\w=\bs 0}.
\ee
Thus, by utilizing divided differences, any monomial can be approximated to an arbitrary degree of accuracy by a neural network employing such an activation function. 
A routine convolution argument was used in \cite{leshnolinpinkus} to show that the statement can be generalized in the case when $\phi$ is a locally bounded function with the property that the closure of the set of discontinuities of $\phi$ has  Lebesgue measure equal to $0$.

In \cite{mhasmich}, we proved a more general result except for the conditions on $\phi$. 
We considered a continuous function $\phi : \RR^d\to\RR$ for some $d\le q$, which is assumed  to satisfy for \emph{some} $N>0$,
\be\label{eq:mhasmichcond}
\sup_{\x\in\RR^d}|\phi(\x)|(1+|\x|)^{-N} <\infty.
\ee
Then $\phi$ is a tempered distribution.
Let $\mathcal{M}_{d,q}$ be the set of all $d\times q$ matrices.
We proved that  the set $\{\phi(A(\circ)-\w): A\in \mathcal{M}_{d,q}, \ \w\in\RR^d\}$ is fundamental in $C(K)$ for every compact subset $K\subset \RR^q$ \emph{if and only if}  $\phi$ is not a polynomial. 
In particular, this result serves as an analogue of the celebrated Wiener Tauberian theorem in the case $d = q$ and, notably, provides necessary and sufficient conditions for RBF networks to act as universal approximators. For $d = 1$, it establishes the necessary and sufficient conditions for the universal approximation property of neural networks.

Finally, we remark that theorems asserting existence of neural networks with just one neuron are also known; e.g., \cite{guliyev2016single}.

\subsection{Degree of approximation}\label{bhag:degapprox}

As expected, there exists an extensive body of literature on estimating the degree of approximation using neural and radial basis function (RBF) networks. It is not feasible to provide a comprehensive review of all the related works. Instead, we will focus on three primary concepts commonly employed in deriving such results and illustrate these concepts through a selection of key papers.
One such concept is to demonstrate that certain classes of approximants, for which the degree of approximation is already established, belong to the closure of the space of neural and RBF networks. In this section, we will refer to this approach as the \emph{closure idea}.
One example of this is given in  \cite{optneur}. 
In this discussion, let $\Pi_n^q$ be the space of all $q$-variate polynomials of coordinatewise degree $<n$. 
The dimension of this space is $n^q$. 
A number of results on the degree of approximation of functions in $C([-1,1]^q)$ from $\Pi_n^q$  are well known (e.g., \cite{timanbk}).
We describe a stable basis for this space.
We define the Chebyshev polynomials for $x=\cos\theta\in [-1,1]$ by \yadi{$T_\ell$}{Chebyshev polynomials}
\be\label{eq:unichebdef}
T_\ell(\cos\theta)=\begin{cases}
\sqrt{2}\cos(\ell\theta), &\mbox{if $\ell\not=0$},\\
1, &\mbox{if $\ell=0$}.
\end{cases}
\ee
Mutivariate Chebyshev polynomials are defined by tensor product: for multi-integer $\m\in\ZZ_+^q$, 
$$
T_\m(\x)=\prod_{j=1}^q T_{m_j}(x_j).
$$
Then 
\be\label{eq:multicheb_monomial}
T_\m(\x)=\sum_{\k: |\k|_\infty \le |\m|_{\infty}} A_{\k,\m}\x^\k.
\ee
Any polynomial $P\in\Pi_n^q$ can be written in the form
\be\label{eq:chebexpansion}
P(\x)=\sum_{\m : |\m|_\infty <n} \hat{P}(\m)T_\m(\x),
\ee
where the following stability condition holds:
$$
\int_{[-1,1]^q} |P(\x)|^2\frac{d\x}{\pi^q\prod_{j=1}^q \sqrt{1-x_j^2}}=\sum_{\m : |\m|_\infty <n} |\hat{P}(\m)|^2
$$
Moreover, explicit constructions for good approximations by polynomials expressed in this basis are well known.

In this discussion, let $\phi$ be an infinitely differentiable, non-polynomial activation function and $b$ be such that
$\phi^{(m)}(b)\not=0$ for any integer $m\ge 0$.  
We let
$$
\mathcal{N}_{n,q}=\{\x\mapsto \sum_{j=1}^{n^q} b_j\phi(\w_j\cdot\x +b)\}
$$
We may then express each $\x^\k$ using \eqref{eq:monomialnet}, and approximate the derivatives involved with divided differences. 
For any $\epsilon>0$, we then obtain  a network of the form
$$
\tilde{T}_{\m,n,\epsilon}(\x)=\sum_{\k: |\k|_\infty \le \m} A_{\k,\m}\mathbb{G}_{n,\epsilon}(\w_{\k,n,\epsilon}\cdot \x +b)\in \mathcal{N}_{n,q}, \qquad |\m|_\infty <n,
$$ 
such that $\|T_\m-\tilde{T}_{\m,n,\epsilon}\|_{\infty, [-1,1]^q} <\epsilon$.
We note  that the weights $\w_{\k,n,\epsilon}$ (as well as the threshold $b$) are fixed depending only on $n$ and $\epsilon$. Thus, $\tilde{T}_{\m,n,\epsilon}$ are \textbf{pre-fabricated networks}. 
By selecting an appropriate value of $\epsilon$, we can translate results from the theory of polynomial approximation to corresponding results for approximation by neural networks.
For example, if $f$ has all derivatives of order up to $r$ which are continuous on $[-1,1]^q$, then there exists $G\in 
\mathcal{N}_{n,q}$ such that
\be\label{eq:sobolapprox}
\|f-G\|_{\infty, [-1,1]^q} \lesssim n^{-r/q}\max_{\k : |\k|_\infty \le r}\|D^\k f\|_{\infty, [-1,1]^q}.
\ee
Likewise, if $f$ is analytic on $[-1,1]^q$, then there exists $G\in \mathcal{N}_{n,q}$ such that
\be\label{eq:analapprox}
\|f-G\|_{\infty, [-1,1]^q} \lesssim_f \exp(-cn),
\ee
for some positive constant $c$ depending only on the ellipsoid of analyticity of $f$.
We note that these networks are constructed using the pre-fabricated networks; only the coefficients are dependent on $f$, and explict expressions for these can be obtained from the theory of polynomial approximation.

The other ideas stem from the following observation.
We may write a neural network of the form $\sum_{k=1}^N a_k\phi(\w\cdot\x+b_k)$ in the integral form
\be\label{eq:nnintform}
\sum_{k=1}^N a_k\phi(\w\cdot\x+b_k)=\int_{\RR^q\times\RR}\phi(\w\cdot\x+b)d\nu(\w,b),
\ee
where $\nu$ is a measure that associates the mass $a_k$ with the point $(\w_k, b_k)\in\RR^q\times\RR$, $k=1,\cdots,N$.
Writing $\bs v=(\w,b)$, $\y=(\x,1)$,  we may express the integral as an integral of the kernel $\phi(\bs v 
\cdot\y)$ with respect to a discrete measure $\nu$ defined on $\RR^{q+1}$. 
Generalizing, we obtain the following definition.
\begin{definition}
    \label{def:variationspace}
Let $\XX$ and $\YY$ be metric measure spaces. A function $G:\XX\times \YY\to \RR$ will be called a \textbf{kernel}.
The \textbf{variation space (generated by $G$)}, denoted by $\mathcal{V}(G)$, is the set of all functions of the form
$ x\mapsto \int_\YY G( x, y)d\tau( y)$ for some signed measure $\tau$ on $\YY$ whenever the integral is well defined.
For integer $N\ge 1$, the set $\mathcal{V}_N(G)$ is defined by
\be\label{eq:manifolddef}
\mathcal{V}_N(G)=\left\{\sum_{k=1}^N a_k G( \circ,  y_k) : a_1,\cdots, a_N\in\RR,\  y_1,\cdots, y_N\in \YY\right\}.
\ee
An element of $\mathcal{V}_N(G)$ will be called a \textbf{$G$-network (with $N$ neurons)}.
\end{definition}
In the literature, the variation space is also referred to as a \emph{native space} or \emph{reproducing kernel Banach space}, and an element of this space is referred to as a continuous (or infinite) network. 
An element of $\mathcal{V}_N(G)$ is referred to as a kernel based network.

One approach to determining the degree of approximation of a function in the variation space is to treat the infinite network as the expectation of a family of random variables of the form $ |\tau|_{TV} G(\mathbf{x}, \circ) h(\circ) $, with respect to the probability measure $\frac{|\tau|}{|\tau|_{TV}} $, where $ h$ is the Radon-Nikodym derivative of $ \tau $ with respect to $ |\tau|$. This framework allows the use of concentration inequalities to derive a discretized kernel-based network.

All of these results have the form
\be\label{eq:genest}
\inf\limits_{P\in \mathcal{V}_N(G)}\|f-P\|_{\infty,\XX}\le CN^{-s}\|\tau\|_{TV},
\ee
for $f\in \mathcal{V}(G)$ subject to various conditions on $G$ and the measure $\tau$ defining $f$.
Here, $s>0$ and  $C>0$ are  constants independent of $f$ (and hence, $\tau$) but may depend in an unspecified manner on $G$, $\XX$, and $\YY$, and the conditions on $\tau$. 
In particular, when $\XX$ and $\YY$ are subsets of a Euclidean space, they may depend upon the dimension of these spaces. 
The bound \eqref{eq:genest} is called \emph{dimension independent} if $s$ is greater than some positive number independent of the dimension, and \emph{tractable} if in addition, $C$ depends at most polynomially on the dimension.
We will refer to this approach as the \textit{probability theory approach.}

We give some examples, especially involving shallow neural networks with sigmoidal and ReLU-type activation functions.

In \cite{barron1993}, Barron introduced a space of functions, which is now known as the \textbf{Barron space}, which is defined more generally as the space of integrable functions $f :\RR^q\to\RR$ whose Fourier transform $\hat{f}$ satisfies, for some $s>0$,
\be\label{eq:barronspace}
\|f\|_{B_s}=\int_{\RR^q} (1+|\w|^2)^{s/2}|\hat{f}(\w)|d\w <\infty.
\ee
Barron demonstrated that $B_1$ is contained within the variation space of a sigmoidal function, thereby proving that for any $f \in B_1$, there exists a neural network $G$ with $N$ neurons such that the $L^2$ norm of $f - G$ on the unit ball of $\mathbb{R}^q $ satisfies $\mathcal{O}(1/\sqrt{N})$. A periodic version of this result was established in \cite{dimindbd}, where dimension-independent bounds for shallow periodic neural networks were derived for the class of continuous periodic functions whose Fourier coefficients are summable. These results are unimprovable in terms of the network width. As to the problem of approximating functions in the variation space by linear combination of elements in the dictionary, DeVore and Temlyakov \cite{devore1997nonlinear} showed that the rate $O(N^{-1/2})$ also holds for Hilbert spaces generated by orthogonal dictionaries using the greedy algorithm.
 K\r{u}rkov$\acute{a}$ and Sanguineti \cite{kurkova1,kurkova2} proved the rate $O(N^{-1/2})$ for Hilbert spaces generated by dictionaries with  conditions weaker  than orthogonality. The constants corresponding to these results are tractable. While $\mathcal{O}(N^{-1/2})$ represents the optimal rate for general dictionaries, this rate can be improved in specific cases. However, in most of the literature discussing these improved rates, the associated constant terms are not necessarily tractable.

For the $L^2$-approximation, Xu \cite{xu2020finite} considered  the approximation of spectral Barron spaces using ReLU$^\gamma$ neural networks, $\gamma\ge 1$ integer, where the corresponding constant is tractable. 
Also, the sharp rate $\tilde O(N^{-\frac{1}{2}-\frac{2\gamma+1}{2d}})$ for the variation space generated by ReLU$^\gamma$ network is proved by Siegel and Xu \cite{siegel2022sharp} without tractable constants. 
Here and in this context, $\tilde{O}(N^{-s})$ means $\O((\log N)^mN^{-s})$ for some $m>0$. 
The improved uniform approximation rates are also studied. Klusowski and Barron \cite{Barron2018} proved the rate $\tilde O(N^{-\frac{1}{2}-\frac{\gamma}{d}})$ for approximating functions from spectral Barron spaces of order $\gamma=\{1,2\}$ by shallow ReLU$^\gamma$ neural networks. They also proved absolute constants in this work. Using the covering number argument as in  \cite{makovoz1998uniform}, Ma, Siegel, and Xu \cite{ma2022uniform} recently obtained the uniform approximation rate $\tilde O(N^{-\frac{1}{2}-\frac{\gamma-1}{d+1}})$ for approximating functions in spectral Barron spaces of order $\gamma$ by ReLU$^\gamma$ networks with unspecified constants. 
All these results are applicable in the case $\gamma\ge 1$ is an integer.

In \cite{tractable}, dimension-independent bounds of the form $\tilde{O}(N^{-1/2})$ are derived for general $G$-networks on non-tensor product Euclidean domains, including neural, radial basis function, and zonal function networks. 
In these bounds, the constants involved depend polynomially on the dimension. 
The paper also explores a duality between the tractability of quadrature and approximation from the closed, convex, symmetric hulls of dictionaries. 
These bounds are based solely on the boundedness properties of the dictionaries, without considering the smoothness properties of the kernels. 
Sharper bounds  under different assumptions involving both boundedness and smoothness properties of $G$ are obtained in \cite{mhaskar2020dimension}, valid in particular for ReLU$^\gamma$ networks, where $\gamma$ is not required to be an integer.
 Naturally, the results in \cite{mhaskar2020dimension} are sharper in cases where both their findings and those in \cite{tractable} are applicable, although the constant terms in the latter remain unspecified.
In \cite{mhaskar2024tractability}, we have obtained tractable bounds which are the state of the art under certain additional conditions.
For example, we may consider the question of approximation by ReLU networks on the Euclidean space as an equivalent question of networks of the form $\x\mapsto \sum_{k=1}^N a_k|\x\cdot\w_k|$ where $\x, \w_k$'s are in $\SS^q$, the unit sphere embedded in $\RR^{q+1}$. 
For the native space of the corresponding kernels, we reproduced the result in \cite{mhaskar2020dimension} giving a rate of approximation $\O(N^{-(q+3)/q})$, but with tractable constants.
Results in \cite{siegel2023optimal} show that these bounds are optimal up to a logarithmic factor.

We note that all the theorems derived through probabilistic arguments are purely existence results; they do not provide insight into the specific information that the resulting networks use regarding the target function, nor do they clarify whether this information is continuous in the native space. Consequently, it is misleading to suggest that such theorems overcome the curse of dimensionality. In fact, it remains unclear whether the native spaces themselves exhibit a curse of dimensionality.

Given data of the form $\{(x_j, f(x_j))\}_{j=1}^N$, it is tempting to find a network in $\mathcal{V}_N(G)$ \yadi{$\mathcal{V}(G)$}{variation space for kernel $G$} \yadi{$\mathcal{V}_N(G)$}{kernel networks with kernel $G$ and $N$ neurons} such that \eqref{eq:genest} is satisfied, for example, by solving an optimization problem of the form

\begin{equation}\label{eq:temptregularization}
\text{Minimize} \quad \sum_{j=1}^M \left( f(x_j) - \sum_{k=1}^N a_k G(x_j,y_k) \right)^2 + \lambda \sum_{k=1}^N \left( |a_k| + |y_k|_1 \right).
\end{equation}

This strategy might not work. 
The optimization problem assumes that the network must be constructed using function evaluations.
It is not clear that estimates such as \eqref{eq:genest} will continue to hold under this extra assumption.
We have shown in \cite{mhaskar1997smooth} that if the activation function of a sequence of neural networks converging uniformly to a function $f$ is smoother than $f$ in terms of number of derivatives then either the coefficients or the weights cannot be bounded; i.e., the regularization term in \eqref{eq:temptregularization} cannot work in this context.

Another approach to obtaining estimates on the approximation error for networks constructed explicitly using known information about the target functions is to employ quadrature formulas in an integral representation similar to \eqref{eq:nnintform}, which is valid for a known class of approximants. We refer to this approach as the \emph{classical approximation theory approach}. To illustrate, consider a $2\pi$-periodic continuous activation function \( \phi \) and networks in the class
$$
\mathcal{N}_{n,m}=\sum_{|\k|_2<n, 0\le j\le 2m+1}a_{\k,j}\phi(\k\cdot\x-t_j).
$$
We note that the number of parameters in $\mathcal{N}_{n,m}$ is $\O(n^qm)$, and the weights need to be multi-integers in order to preserve the periodicity.
We will assume that $\hat{\phi}(1)\not=0$.
It is not difficult to observe \cite{dimindbd,mhaskar1995degree} that
\be\label{eq:trig_as_neur}
\exp(i\k\cdot\x)=\frac{1}{2\pi\hat{\phi}(1)}\int_{\TT} \phi(\k\cdot\x-t)\exp(it)dt.
\ee
In this section, we will denote by $\Pi_n^q$ the span of $\{\x\mapsto \exp(i\k\cdot\x), \ |\k|_2 \le n\}$. 
Then \eqref{eq:trig_as_neur} shows that each $T\in\Pi_n^q$ is in the variation space of $\phi$; indeed,
\be\label{eq:trigpoly_as_neur}
T(\x)=\frac{1}{2\pi\hat{\phi}(1)}\sum_{|\k|_2<n}\int_{\TT} \hat{T}(\k)\phi(\k\cdot\x-t)\exp(it)dt, \qquad T\in\Pi_n^q.
\ee
We discretize this integral using discrete Fourier transform to obtain a network.
Let
\be\label{eq:trigasneunet}
\mathbb{G}_m(\phi, T)(\x)= \frac{1}{(2m+1)\hat{\phi}(1)}\sum_{j=0}^{2m}\exp\left(\frac{2\pi i j}{2m+1}\right)\left(\sum_{\k\in\ZZ^q} \hat{T}(\k)\phi\left(\k\cdot(\circ)-\frac{2\pi  j}{2m+1}\right)\right) \in \mathcal{N}_{n,m},
\ee
where it is understood that $\hat{T}(\k)=0$ if $|\k|_2$ exceeds the degree $n$ of $T$. 
The estimate \eref{eq:trigasneunet} leads to
\be\label{eq:trigneunetapprox}
\left\|T-\mathbb{G}_N(\phi,T)\right\|_\infty
\le \frac{4}{|\hat{\phi}(1)|}\mathsf{dist}(L^1(\TT);\phi,\Pi_N^1)\sum_{\k\in\ZZ^q}|\hat{T}(\k)|.
\ee
Finally, let $\C=\{\x_1,\cdots,x_M\}$ be a set of arbitrary points in $\TT^q$ so that there exists integer $n\ge 1$ and a MZQ measure $\nu\in MZQ(n)$ supported on $\C$ (cf. Definition~\ref{def:mzdef}).
Let $f\in C(\TT^q)$, and only the values $f(\x_j)$, $j=1,\cdots, M$ be known.
We use \eqref{eq:trigneunetapprox} with $\sigma_n(\nu;f)$ in place of $T$. 
Writing 
$$
\hat{f}(\nu;\k)=\int_{\TT^q}f(\x)\exp(-i\k\cdot\x)d\nu(\x),
$$
some further estimation using Parseval identity leads to
\begin{proposition}\label{prop:periodic_relu_prop}
Let $\phi\in C^*(\TT)$, $\hat{\phi}(1)\not=0$, $f\in C^*$.  Then for $n,N\ge 1$, we have
\be\label{eq:neural_approx}
\begin{aligned}
\|f-\mathbb{G}_m(\phi,\sigma_n(f))\|&=\left\|f  -\frac{1}{(2m+1)\hat{\phi}(1)}\sum_{j=0}^{2m}\exp\left(\frac{2\pi i j}{2m+1}\right)\right. \\
&\qquad\qquad\times\left.\left(\sum_{|\k|_2<n} h\left(\frac{|\k|_2}{n}\right)\hat{f}(\nu;\k)\phi\left(\k\cdot(\circ)-\frac{2\pi  j}{2m+1}\right)\right)\right\|\\
&\le c(\phi)\left\{\dist(C(\TT^q);f,\Pi_n^q)+n^{q/2}\dist(L^1(\TT);\phi,\Pi_n^1)\|f\|_\infty\right\}.
\end{aligned}
\ee
\end{proposition}

\begin{example}\label{uda:periodic_relu_rmk}
{\rm For example, we consider  the smooth ReLU function $t\mapsto \log(1+e^t)=t_++\O(e^{-|t|})$. Then the function
$\disp\psi(t)=\log\left(\frac{(1+e^{t+\pi})(1+e^{t-\pi})}{(1+e^t)^2}\right)$
is integrable on $\RR$. 
The periodization 
\be\label{eq:smoothreludef}
\phi(t)=\sum_{j\in\ZZ}\psi(t+2\pi j), \qquad t\in\RR,
\ee
is an analytic function on $\TT$. So, Bernstein approximation theorem \cite[Theorem~5.4.2]{timanbk} shows that there exists $\rho_1<1$ with $\dist(L^1(\TT);\phi,\Pi_n^1)\le \rho_1^N$ for all $N\gs c(\phi)$. 
In Proposition~\ref{prop:periodic_relu_prop}, if $f$ satisfies
$\dist(C(\TT^q);f,\Pi_n^q)=\O(n^{-\gamma})$ for some $\gamma$, then we may choose $m\sim \log n$ to get a network with $\O(n^q\log n)$ neurons 
to obtain an estimate $\O(n^{-\gamma})$ on the right hand side of \eref{eq:neural_approx}.
\qed}
\end{example}

This idea has been generalized to various other settings, and algorithms are available that find approximations to the target function using the training data, \textbf{without assuming any prior knowledge about the target function} (see, for example, \cite{indiapap} for an early construction). However, directly formulating the problem as the minimization of the supremum norm error between the function and the neural network model may not be effective. The theory suggests specific relationships between the coefficients, weights, and thresholds.

Recently, a combination of the approximation theory approach and probability theory approach has been studied to get estimates for approximation of smooth functions (in the sense of approximation theory) by neural and other networks in many different contexts  
\cite{mao2023approximating, mhaskar2025approximation}.
In this approach, the target function is first approximated by a carefully constructed trigonometric polynomial, for which error estimates are known. This polynomial is then expressed as an element of the variation space, to which probabilistic techniques are applied. However, it is important to note that this approach leads only to existence theorems, rather than providing algorithms for constructing the networks.

\subsection{Interpolation by neworks}\label{bhag:interpolation}
Colloquially, the term interpolation of the data $\{(x_j,y_j)\}_{j=1}^M$ refers to predicting a value of $y$ for $x\not\in\{x_j\}_{j=1}^M$ based on this data.
In approximation theory, this term is used in a more restricted way. 
We have a family of functions $\{G_k\}_{k=1}^M$. The intepolation problem is to find coefficients $a_k$ such that
\be\label{eq:interp_problem}
\sum_{k=1}^M a_k G_k(x_j)=y_j, \qquad j=1,\cdots, M.
\ee
For example, when $x_j$'s are distinct points on the real line, and $G_k(x)=x^k$, then this is the very classical polynomial interpolation problem studied for centuries.
When $x_j$'s are in $\RR^2$ then the polynomial interpolation problem may not have a solution.
HNM recalls the excitement in the approximation theory community with the result of Micchelli \cite{micchelli1984interpolation} which states the following.  
\begin{theorem}\label{theo:micchellirbf}
Let $F \in C^\infty([0,\infty)$, $(-1)^\ell F^{(\ell)}(x) \le 0$ for all $x\in (0,\infty)$ and integer $\ell\ge 1$.
Assume that $F$ is not constant and $F(x)>0$ for all $x\in (0,\infty)$. Then for any integer $q\ge 1$, and arbitrary distinct points $x_j\in\RR^q$,
\be\label{eq:rbfdet}
(-1)^M \det(F(|x_j-x_k|^2) >0.
\ee
In particular, the interpolation problem with $G_k(x)=F(|x-x_k|^2)$ has a unique solution.
\end{theorem}
This theorem motivated a great deal of interest and research in the theory of radial basis functions. 
An analogue for neural networks is investigated by Ito  \cite{ito1996nonlinearity}.

Even though the sequence of polynomials $L_n(f)$ of degree $<n$ does not in general converge to $f$ for every continuous $f$ even in one variable, Erd\H os asked whether a sequence of polynomials of degree $<cn$ for some $c\ge 1$ exists to interpolate $f$ at $n$ points, and provides a good approximation to $f$ in the uniform norm on $[-1,1]$. 
This question was answered in the positive by Szabados \cite{szabados1978some}, provided that the degree of the polynomial interpolating at points $\cos\theta_1,\cdots, \cos\theta_M$ is at least a constant multiple of the reciprocal of the minimal separation among the $\theta$'s.
In  \cite{approxint2002}, we gave a very general theorem that states that a similar statement is true anytime a Favard estimate holds for the class of interpolants. In particular, our theorem applies to spjherical basis function networks.
These theorems are existence theorems.
In  \cite{bdint, birkhoffpap}, we have proposed a constructive method called minimum Sobolev norm interpoation to compute a higher degree interpolant, not just prescribing the values, but also various derivatives of the interpolanat.

In the context of machine learning, the subject has attracted a renewed interest along two lines of research. The theory of \emph{extreme learning machines} seeks to fix the inner weights of a neural (or more general) network, and compute an interpolant or least squares approximation based on the coefficients of the network. 
A recent survey of the topic is given by Huang et al.  \cite{huang2015trends}.
As in \cite{bdint}, it is shown in   \cite{auricchio2024accuracy} that this procedure avoids the so-called Runge phenomenon; i.e., the divergence of the interpolants even for an analytic function.
The generalization errors of the extreme learning machines is studied recently in 
\cite{kim2023theoretical}.

Another direction of research is the observation in  \cite{doi:10.1137/20M133607} that an over-parametrized nerual network can achieve both zero training error (i.e., exact interpolation of the training data), as well as an improvement in the generalization error, contrary to received wisdom.
This seminal paper has given rise to a great deal of research, mostly focusing on the optimization problems to accomplish such over-paremetrized training.
Leaving aside the question of the optimization processes, we have investigated in  \cite{mhaskar2018analysis,mason2021manifold} the question of intrinsic characteristics of the data that allows this phenomenon in the first place, especially in the presence of noise.
As expected, the minimal separation among the data points plays a central role in this theory.

\bhag{Approximation by deep networks}\label{bhag:neural}

In recent years, deep learning has been a topic of great research, with many eye-popping applications. 
A \textbf{shallow neural network} is a function of the form
\be\label{eq:shallowneuraldef}
x\mapsto \sum_{k=1}^N a_k\phi(w_k\cdot x +b_k), \qquad x\in\RR^q,
\ee
where $\phi$ is a nonlinear function, called an \textbf{activation function}, $w_k$'s are in $\RR^q$, $a_k$' and $b_k$'s are real numbers. 
A deep network with $L$ hidden layers is defined usually as a function
\be\label{eq"deepneuraldef}
x\mapsto W_L\phi(W_{L-1}\phi(W_{L-2}\phi(\cdots(\phi(W_0x +b_0)+b_1+\cdots+b_{L-1}),
\ee
where $W_j$'s are rectangular matrices, $b_j$'s are vectors of different dimensions, and the notation such as $\phi(W_0x+b_0)$ is understood as a mapping from one Euclidean space to another where $\phi$ is evaluated componentwise.
A straightforward way to visualize this is as a tree, where each non-leaf node evaluates a neural network (referred to as a \textbf{channel} in the new terminology) on the inputs represented by the leaves of the tree. From this perspective, each channel could evaluate any function. An alternative and potentially more advantageous approach is to think of a deep network as a directed acyclic graph (DAG). If the target function itself exhibits a compositional structure, suggested by the DAG, the functions evaluated at each channel are referred to as \textbf{constituent functions}.

\begin{figure}[ht]
\begin{center}
\includegraphics[width=\textwidth]{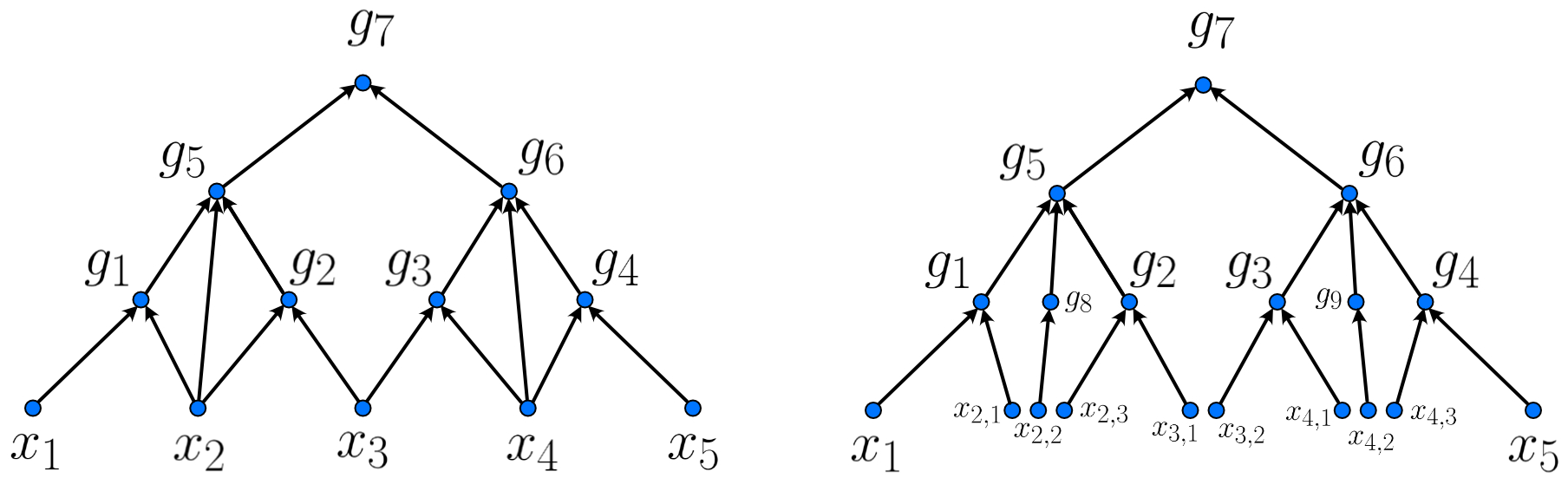} 
\end{center}
\caption{Left: a deep network as a directed acyclic graph, Right: An equivalent tree representation. Each non-leaf channel evaluates a neural network to approximate a function at that node.}
\label{fig:deepnet}
\end{figure}

There is an equivalence between DAGs and trees, as demonstrated in Figure~\ref{fig:deepnet}. 
For example, the left figure in Figure~\ref{fig:deepnet} shows a DAG representing a target function
$$
g=g_7\left(g_5(g_1(x_1,x_2),x_2,g_2(x_2,x_3)), g_6(g_3(x_3,x_4),x_3,g_4(x_4,x_5))\right).
$$
The function $g$ is obviously a function of $5$ variables.
The constituent functions are $g_1,\cdots, g_7$.
An equivalent tree representation on the right makes $g$ to be a function of $10$ variables instead  leaving the true structure to the construction of the matrices $W$.

We first recall some older results regarding deep networks.
In some sense, the oldest result on ``deep networks'' is perhaps, the Kolmogorov-Arnold-Lorentz theorem \cite[Chapter~17, Theorem~1.1]{lorentz_advanced}:
\begin{theorem}
\label{theo:kolmogorov}
There exist $q$ positive constants $\lambda_1,\cdots,\lambda_q$, with $\sum_{k=1}^q\lambda_k \le 1$, $2q+1$ continuous, strictly increasing functions $\phi_1,\cdots,\phi_{2q+1}$, with each $\phi_k :[0,1]\to [0,1]$ with the following property: for each $f\in C([0,1]^q)$, there exists $g\in C[0,1]$ (depending on $f$) such that
\be\label{eq:kolmogorov}
f(\x)=\sum_{k=1}^{2q+1}g\left(\sum_{j=1}^q \lambda_j\phi_k(x_j)\right).
\ee
\end{theorem}
In machine learning terms, there exist activation functions $\phi_j$ and $g$ such that any $ f \in C([0,1]^q)$ can be represented \textbf{exactly} as a deep network with two hidden layers, as shown in \eqref{eq:kolmogorov}. The functions $\phi_j$ are referred to as the \textbf{inner functions}, while the function $g$ is known as the \textbf{outer function}.
Several efforts have been made to construct deep networks based on this theorem (see \cite{liu2024kan} and the references therein). For example, Lai and Shen \cite{lai2021kolmogorov, lai2024optimal} implement the proof from the book \cite{lorentz2005approximation} by Lorentz,  and obtain approximation results based on certain assumptions on both the inner and outer functions. 
It is claimed that these networks use certain judiciously chosen samples of the target function, yet overcome the curse of dimensionality.
Clearly, the set of target functions so approximated cannot have the curse of dimensionality to begin with.  Liu et al.\ \cite{liu2024kan} assume that the target function is a superposition of several layers (rather than two) of univariate functions, with each being smooth.
Results of similar nature are also given by Schmidt-Hieber \cite{schmidt2021kolmogorov}.

However, these functions cannot yield optimal or constructive approximation results. A theorem of Vitushkin in \cite[Chapter~17, Theorem~3.1]{lorentz_advanced} states the following.
\begin{theorem}
\label{theo:vitushkin}
For any $r\ge 1$, there exist $r$ times continuously differentiable functions on $[0,1]^q$ which cannot be represented by $r$ times continuously differentiable functions of fewer variables. 
\end{theorem}
In \cite{chui1994neural}, it was demonstrated that it is not possible to approximate the characteristic function of a cube in $\mathbb{R}^q$ (with $q \geq 2$) arbitrarily well using shallow neural networks with a fixed number of neurons, each employing the Heaviside function as an activation function. A similar, yet stronger, result was established in \cite{chui1996limitations}.
For any $m\ge 1$, the error in approximating the characteristic function of a  cube of side $<3^{-m}$ by a network of the form $\sum_{k=1}^m a_k \phi_k(\w_k\cdot\x+b_k)$ in the $L^1$ norm is at least $1$.
We note that each neuron in this network is allowed to use a different activation function $\phi_k$, provided that each $\phi_k$ is locally integrable. However, it is not difficult to see that if $\phi$ is the Heaviside function, then
\be\label{eq:charfn}
\phi\left(\sum_{k=1}^q \phi(1+x_k)+\sum_{k=1}^q \phi(1-x_k)-2q+1/2\right)=\begin{cases}
1, &\mbox{if $\x\in [-1,1]^q$},\\
0, &\mbox{otherwise.}
\end{cases}
\ee
Thus, a network with two hidden layers can be easily constructed to provide a piecewise constant approximation to any function. Similarly, it is well known \cite{chuili1992} that any $q$-variate polynomial of total degree less than \( 2^m \) can be expressed in the form $\sum_{k=1}^M (\mathbf{w}_k \cdot \mathbf{x} + b_k)^m $, where $ M$ is the dimension of the space, i.e., $ M = \binom{2^m + q}{q} $.
It is observed in \cite{multilayer} that since 
$$
x^{2^m}=\left(\left(\cdots \left(x^2)^2\right)^2\right)^2\cdots\right),
$$
and $x^2=(\max(x,0))^2+(\max(-x,0))^2$, any such polynomial can be expressed exactly as a deep network with $m$ hidden layers with $2Mm$ neurons altogether, each evaluating the ReLU$^2$ activation function $x_+^2=(\max(x,0))^2$.
A similar observation was made in the context of synthesizing tensor product splines. As a result, approximation theory results for either free knot tensor product splines or polynomials can be easily translated into results on approximation using deep networks, with each neuron employing the ReLU$^2$ activation function.

In recent years, the ReLU activation function $x_+= \max(x,0)$ is used ubiquitously because of the ease of computing the gradient at points except at the origin. 
Yarotsky, \cite{yarotsky2016error} has shown that the function $(x_+)^2$ can be approximated arbitrarily well on $[0,1]$ by a deep ReLU network. 
This result is sharpened by many authors, such as Schimdt-Hieber \cite{schmidt2020nonparametric}. 
We quote the result as in the paper \cite{bolcskei2019optimal} of B\"olcskei in a slightly modified form:
\begin{proposition}\label{prop:yarotsky}
Let $\epsilon>0$. 
There exists a constant $C>0$ and an ReLU network $\mathbb{G}_\epsilon$ with at most $C\log(\epsilon^{-1})$ hidden layers, each with $3$ neurons, and all weights and thresholds bounded in absolute value by $1$, such that $\mathbb{G}_\epsilon(0)=0$ and
\be\label{eq:yartosky}
\max_{x\in [-1,1]}|(x_+)^2-\mathbb{G}(x)|\le \epsilon.
\ee
\end{proposition}
In \cite{bolcskei2019optimal}, the result is stated for approximation of $x^2$ on $[0,1]$.
Since $\mathbb{G}(x_+)$ is also an ReLU network with one additional layer, the statement holds as stated.
With the help of Proposition~\ref{prop:yarotsky}, all the results in \cite{multilayer}  can be translated to the approximation properties of ReLU networks as well.
A number of consequences of this idea are studied in a plethora of recent papers, too many to cite, but including \cite{perekrestenko2018universal}.

A lot of work has focused on improving our understanding of approximation by deep networks under geometric and structural priors. In particular in \cite{kohler2021rate,kohler2022estimation,kohler2023estimation} is established the minimax-optimal convergence rates for fully connected deep neural networks in regression tasks, under assumptions such as low local intrinsic dimensionality or manifold-constrained target functions. 
Complementarily, in ~\cite{burkholz2021existence} is explored the existence of so-called \emph{universal lottery tickets}—subnetworks capable of generalizing well across tasks—highlighting the role of sparsity, initialization, and overparameterization in neural architectures.
Cloninger and Klock \cite{cloninger2021deep} have studied approximation by deep ReLU networks on tubular neighborhoods of manifolds and obtained rates of approximation for functions with special structures, which depend only the dimension of the manifold. 

In \cite{dingxuanpap}, we investigate why and when deep networks exhibit better approximation properties than shallow networks. To discuss these findings, we consider the directed acyclic graph (DAG) representation. For clarity, let us assume the setting of periodic functions, with all functions mentioned in this discussion being Lipschitz continuous. While a more natural setting would involve the sphere, we defer this aspect to Section~\ref{bhag:sphere}, after developing the basic theory of approximation on the sphere.

In the periodic setting and the network of Figure~\ref{fig:deepnet},  each of the bivariate functions $g_7, g_1, g_2, g_3, g_4$ can be approximated up to an accuracy $\epsilon$ using  $\O(\epsilon^{-1/2})$ samples of its input, and each of the functions $g_5, g_6$ can be approximated using $\O(\epsilon^{-1/3})$ samples.
 So, a deep network that follows the DAG structure will require only $\O(\epsilon^{-1/3})$ samples to achieve an accuracy of $\epsilon$.
A shallow network is not able to exploit this compositional structure, and hence, will require $\Omega(\epsilon^{-1/5})$ number of samples in theory.

\begin{remark}\label{rem:isitbetter}
Deep networks are not always superior for approximation. 
First,  shallow networks can do an equivalent job if the class of functions does not suffer a curse of dimensionality to begin with.
Another example refers to our discussion above.
The function $g_5$, which is a function of three variables, will require $\Omega(\epsilon^{-1/3})$ samples regardless of its compositional structure, meaning that the choice between a shallow or deep network does not affect the sample complexity. Moreover, it is not difficult to construct directed acyclic graphs (DAGs) where a deep network may yield even worse results compared to a shallow network, for example, if $g_5$ had the structure
$$
g_5(x_1, g_1(x_1,x_2), x_3, g_2(x_2, x_3), x_3),
$$
making it a function of $5$ variables, rather than a non-compositional point of view, where it would be a function of $3$ variables only.
\qed
\end{remark}

\begin{remark}\label{rem:turing}
Clearly, all Turing-computable functions can be considered compositional, corresponding to the state diagram of the Turing machine involved, with the back arcs unrolled. \qed
\end{remark}

\begin{remark}\label{rem:compositional}
The compositional structure is not unique. 
A simple example is given by the identity
\be\label{eq:toyexample}
\tan^{-1}x-\tan^{-1}y=\tan^{-1}\left(\frac{x-y}{1+xy}\right)=\cos^{-1}\left(\frac{1+xy}{\sqrt{(1+x^2)(1+y^2)}}\right)
\ee
which can be viewed in the form $f(x,y)=f_1(x)-f_1(y)$ or as $f(x,y)=f_2(f_3(x,y))$ or $f(x,y)=f_4(f_5(f_6(x,y),f_7(x,y)))$. 
Thus, different DAGs can represent the same function. 
Of course, we may view it is a single function of two variables as well.  
Thus, different DAGs can be used to represent the same function.
In particular, it is not meaningful to ask whether a function is compositional or not.

A more reasonable question is whether a given function can be represented by a specific DAG. In theory, such an effort is explored in \cite{filom2020pde}. 
Another way to approach the problem is to consider it for a class of functions. In this context, the DAG can be used to approximate a variety of functions and their derivatives, depending on the number of input variables. The DAG structure imposes constraints on the Jacobians of the resulting maps between Euclidean spaces of the same dimension.
Even if this question is resolved, the constituent functions are not uniquely determined within a single DAG, as demonstrated by the second identity in \eqref{eq:toyexample}.
\qed
\end{remark}

\begin{remark}\label{rem:reduction}
A side effect of our observations above show that the approximation power of a deep network is analyzed trivially once we understand the approximation power of shallow networks. 
 \qed
\end{remark}

Notwithstanding Remark~\ref{rem:reduction}, the question of function approximation is still relevant if one wants to impose extra restrictions on the deep networks such as convolutional networks.
We describe a recent work by D.-X. Zhou \cite{zhou2020universality} as an example. 

A \textbf{filter mask with stride $s$} is a doubly infinite sequence $(w_j)_{j=-\infty}^\infty$ where $w_j=0$ for $j\not\in\{0,\cdots, s\}$.
A convolution of two sequences $\w$ and $\x$ is defined by
$$
(\w\ast \x)_k =\sum_{j=-\infty}^\infty w_{k-j}x_j=\sum_{j=-\infty}^\infty w_jx_{k-j},
$$
whenever the sum is well defined.
A shallow network is convolutional if it has the form
$$
\x\mapsto \sum_{k=1}^N a_k\phi((\w\ast \x)_k+b_k),
$$
where $\x\in\RR^q$ is considered to be a doubly infinite sequence whose components are \textit{``zero padded}''; i.e., $x_k=0$ if $k\le 0$ or $k>q$.
A deep network is referred to as a \textbf{convolutional neural network (CNN)} if each channel evaluates a convolutional shallow network. Zhou observed that if $a\in \RR^n$ is zero-padded, its \textit{symbol} $\sum_k a_k z^k$ is a polynomial of degree at most $n$, which can be factored into real quadratic and linear factors. By applying the Cauchy-Leibnitz formula for the product of polynomials, any such sequence can be expressed as convolutions of filter masks with a stride of at most 2. Therefore, Zhou proved that any shallow network can always be approximated (if not explicitly expressed) as a CNN.
In particular, all theorems concerning approximation by shallow networks can be translated into those regarding approximation by CNNs. This program was developed in conjunction with a result by Barron and Klusowski on the approximation power of shallow ReLU networks, which led to the assertion that CNNs are universal approximators.

Recent theoretical developments by Zhou and his colloborators have shown that deep convolutional neural networks (CNNs) can  exploit structural properties of target functions even without explicitly encoding such features into the architecture. In particular, in \cite{song2023approximation} demonstrated that CNNs can approximate ridge functions efficiently, achieving approximation rates comparable to those for univariate functions of similar smoothness, despite the absence of explicit ridge constraints. Similarly, in \cite{mao2021theory} was shown that deep CNNs can exploit radial symmetry, approximating radial functions at rates comparable to univariate approximation. These results highlight the inherent ability of deep architectures to adapt to hidden low-dimensional structures within high-dimensional data.

\bhag{Manifold learning}\label{bhag:manifoldlearning}

The constructive approximation results discussed thus far pertain to high-dimensional Euclidean domains, where the function classes often exhibit the curse of dimensionality. As a result, there exists a noticeable gap between the theoretical foundations of machine learning and its practical applications. The theory of manifold learning seeks to bridge this gap by providing a framework that better captures the intrinsic structure of data in high-dimensional spaces.

\begin{definition}
\label{def:manifold_hyp}
The \textbf{manifold hypothesis} states that the data $\{x_j\}\subset\RR^Q$ belongs to some (unknown) compact, smooth, orientable submanifold $\XX$ of the ambient space $\RR^Q$, with dimension $q\ll Q$.	
It is also customary to assume that the distribution of this data has the form $f_0d\mu^*$ where $\mu^*$ is the Riemannian volume measure of $\XX$, normalized to be a probability measure, and $f_0\sim 1$.
\end{definition}
Several studies have focused on evaluating the manifold hypothesis; for example, see \cite{fefferman2016testing} for an in-depth analysis and \cite{fefferman2023fitting} on how to fit a manifold on noisy data.

A great deal of research is devoted to the estimation of various quantities related to the unknown manifold $\XX$
from the data.
Some of these important quantities are the (negative) Laplace-Beltrami operator $\Delta$, its eigendecomposition: $\Delta\phi_k=\lambda_k^2\phi_k$, $k\in \ZZ_+$, and the heat kernel defined by
\be\label{eq:heatkerndef}
\mathcal{K}_t(x,y)=\sum_{k=0}^\infty \exp(-\lambda_k^2t)\phi_k(x)\phi_k(y), \qquad x, y\in \XX.
\ee
We note that $\mathcal{K}_t$ is the Green function for the heat equation $\Delta  u+(\partial u/\partial t)=0$.
The heat kernel defines a distance on $\XX$ (known as \textbf{diffusion metric}):
\be\label{eq:diffusionmetric}
\rho_{\mbox{diffusion}, t}^2(x,y)=\mathcal{K}_t(x,x)+\mathcal{K}_t(y,y)-2\mathcal{K}_t(x,y).
\ee
The Varadhan formula \cite{varadhan1967behavior} states that
\be\label{eq:varadhan}
\lim_{t\downarrow 0} 4t\log(\mathcal{K}_t(x,y))=-\rho(x,y)^2,
\ee
where $\rho$ is the geodesic distance metric on $\XX$.

In the remainder of this section, $\rho$ will denote the geodesic distance on $\XX$, and $\mu^*$ the Riemannian volume measure, normalized to be a probability measure.
The dimension of the manifold is denoted by $q$, that of the ambient Euclidean space by $Q$, and the Euclidean norm on $\RR^d$ is denoted by $|\cdot|_{2,d}$.

The next result illustrates the stability of the heat kernel by showing that it defines a mapping which approximately preserves geodesic distances within a local neighborhood of a manifold.

\begin{theorem}[\cite{jones2010universal} Theorem~2.2.7]\label{theo:jmstheo}
 Let $\x_0^*\in\XX$. There exist constants  $R>0$, $c_1,\cdots, c_6>0$ depending on $\x_0^*$ with the following property. Let $\mathbf{p}_1,\cdots,\mathbf{p}_q$ be $q$ linearly independent vectors in $\RR^q$, and $\x_j^*\in\XX$ be chosen so that $\x_j^*-\x_0^*$ is in the direction of $\mathbf{p}_j$, $j=1,\cdots,q$, and 
$$
c_1R\le \rho(\x_j^*,\x_0^*)\le c_2R, \qquad j=1,\cdots,q,
$$ 
and $t=c_3R^2$. Let $B\subset\XX$ be the geodesic ball of radius $c_4R$, centered at $\x_0^*$, and 
\be\label{eq:jmsphidef}
\Phi_{\mbox{{\rm jms}}}(\x)=R^q(\mathcal{K}_t(\x,\x_1^*), \cdots, \mathcal{K}_t(\x,\x_q^*)), \qquad \x\in B.
\ee
Then
\be\label{eq:jmsdistpreserve}
\frac{c_5}{R}\rho(\x_1,\x_2)\le \|\Phi_{\mbox{{\rm jms}}}(\x_1)-\Phi_{\mbox{{\rm jms}}}(\x_2)\|_{2,q}\le 
\frac{c_6}{R}\rho(\x_1,\x_2), \qquad \x_1,\x_2\in B.
\ee
\end{theorem}
Since the paper \cite{jones2010universal} deals with a very general manifold, the mapping $\Phi_{\mbox{jms}}$ is not claimed to be a diffeomorphism, although it is obviously one--one on $B$.
Other efforts to learn the atlas include \cite{chui_deep,schmidt2019deep,shaham2018provable}.
Once the atlas is estimated, classical approximation theory methods can be used to approximate a function on $\XX$.

Another relevant area in this context is diffusion geometry, which focuses on estimating the eigensystem $\lambda_k$, $\phi_k$ of the Laplace-Beltrami operator or similar second-order elliptic operators. An early introduction to this subject can be found in the special issue \cite{achaspissue} of Applied and Computational Harmonic Analysis.
The following theorem of Belkin and Niyogi, \cite{belkinfound} plays a fundamental role in this theory.

\begin{theorem}\label{theo:belkin_niyogi}
For a smooth function $f : \XX\to\RR$,
\be\label{eq:belkinbd}
\lim_{t\to 0}\frac{1}{t(4\pi t)^{q/2}}\int_\XX \exp\left(-\frac{|\x-\y|_{2,Q}^2}{t}\right)(f(\y)-f(\x))d\mu^*(\y) = \Delta(f)(\x)
\ee 
uniformly for $\x\in\XX$.
Equivalently, uniformly for $\x\in\XX$, we have
\be\label{eq:belkin_saturation}
\left|\frac{1}{(4\pi t)^{q/2}}\int_\XX \exp\left(-\frac{|\x-\y|_{2,Q}^2}{t}\right)(f(\y)-f(\x))d\mu^*(\y)-t\Delta(f)(\x)\right|=o(t)
\ee
as $t\to 0+$.
\end{theorem}

In practice, the integral expression in \eqref{eq:belkinbd} is discretized to obtain the so called graph Laplacian.
The following discussion is based on  \cite{matrixpap}.
For $t>0$ and $x, y\in\mathbb{R}^Q$, let
\begin{equation}\label{eq:affinityfn}
W^t(x,y):=\exp\left(-\frac{|x-y|_{2,Q}^2}{t}\right).
\end{equation}
We consider the data points $\{x_i\}_{i=1}^M$ as vertices of an undirected graph, where the edge weight between $x_i$ and $x_j$ is given by $W^t(x_i, x_j)$. This defines a weighted adjacency matrix, referred to as the \textbf{affinity matrix}, denoted by $\mathbf{W}^t$.
We define $\mathbf{D}^t$ to be the diagonal matrix with the $i$-th entry on the diagonal given by $\displaystyle \sum_{j=1}^M W^t(x_i,x_j)$.
The \textbf{graph Laplacian} is defined by the matrix
\begin{equation}\label{eq:graphlaplacian}
\mathbf{L}^t=\frac{1}{M}\left\{\mathbf{D}^t-\mathbf{W}^t\right\}.
\end{equation}
We note that for any real numbers $a_1,\cdots,a_M$,
$$
\sum_{i,j=1}^M a_ia_j L^t_{i,j} = \frac{1}{2M}\sum_{i,j=1}^M W^t_{i,j}(a_i-a_j)^2.
$$
We conclude that the eigenvalues of $\mathbf{L}^t$ are all real and non-negative, and therefore, can be ordered as
\begin{equation}\label{eq:graphlaplacian_eigvalues}
0=\lambda^t_0\le \lambda^t_2 \le \cdots\le \lambda^t_{M-1}.
\end{equation}
We may consider the eigenvector corresponding to $\lambda^t_k$ to be a function on $\{x_j\}_{j=1}^M$ rather than a vector in $\mathbb{R}^M$, and denote it by $\phi^t_k$, thus,
\be\label{eq:eigvectordef}
\lambda^t_k\phi_k^t(x_i)=\sum_{j=1}^M L^t_{i,j}\phi^t_k(x_j) = \frac{1}{M}\left(\phi_k^t(x_i)\sum_{j=1}^M W^t(x_i,x_j)-\sum_{j=1}^M W^t(x_i,x_j)\phi_k^t(x_j)\right), \qquad k=0,\cdots,M-1.
\ee
\begin{theorem}\label{theo:grapheig}
{\rm \cite[Theorem~2.1]{Bel_Niy_2007}} Let $\{x_i\}_{i=1}^M$ be chosen randomly from the distribution $\mu^*$,
and let $\phi_k^t$ denote the Nystr\"om extension of the eigenfunction $\phi_k^t$ originally defined on the sample points via \eqref{eq:eigvectordef}. That is,
\begin{equation}\label{eq:eigen_function}
\phi_k^t(x)=\frac{\sum_{j=1}^M  W^t(x,x_j)\phi_k^t(x_j)}{\sum_{j=1}^M  W^t(x,x_j)-M\lambda_k^t}, \quad x \in \mathbb{R}^Q,
\end{equation}
for all $x$ where the denominator is nonzero.

Let $\phi_k$ be the eigenfunction of the Laplace-Beltrami operator on $\XX$ corresponding to the eigenvalue $\lambda_k$ and sharing the same index $k$ as $\phi_k^t$. Then, there exists a sequence $t_n \to 0$ such that
\begin{equation}\label{eigvaluelimit}
\lim_{n\to\infty} \frac{1}{t_n^{1+q/2}}|\lambda_k^{t_n}-\lambda_k| =0,
\end{equation}
and
\begin{equation}\label{eigfnlimit}
\lim_{n \to \infty} \| \phi_k^{t_n} - \phi_k\| = 0,
\end{equation}
where the norm is the $L^2$ norm and the limits are taken in probability with respect to $\mu^*$.
\end{theorem}

Variants of these theorems are proved by Lafon \cite{lafon} and Singer \cite{singer2006graph}, giving rates of convergence.
There are different variants of the graph Laplacian as well; e.g., by renormalizing the affinity matrix \cite{sule2023sharp,coifmanlafondiffusion,hein2005intrinsic} and many heuristics about how to choose this denominator \cite{coifmanlafondiffusion,von2008consistency}.

The approximation of the Laplace-Beltrami operator by the graph Laplacian establishes a connection between spectral graph theory \cite{chung1997spectral} and manifold learning. For instance, a two-class classification problem can be framed as a task of partitioning the dataset into two subsets, one with label $c_i = 1$ and the other with label $c_i = -1$. By treating the data points as nodes on a graph, this problem becomes equivalent to a graph cut problem:\\
\begin{center}
\textit{Minimize $\sum_{(i,j)\in E}(c_i-c_j)^2$, subject to $\sum_i c_i=0$, and $c_i\in \{-1,1\}$,}
\end{center}
where $E$ is the set of edges.
A relaxation of this problem is to replace the condition $c_i\in \{-1,1\}$ by $\sum_i c_i^2=1$, and allow $c_i\in\RR$. 
The Raleigh-Riesz theorem \cite{horn_johnson_book} leads to the fact that optimal solution of this problem is the eigenvector of the graph Laplacian correspoinding to the second (least positive) eigenvalue $\lambda_1^t$, known as the \textbf{Fiedler vector}.

The signs of the components of this vector provide the desired classification. This process can be iteratively repeated to achieve multi-class classification, as described in \cite{irion2015applied}. Alternatively, Lafon and Lee \cite{lafon2006diffusion} propose a different procedure for achieving the same objective.

\bhag{Approximation on data spaces}\label{bhag:dataspaces}

\subsection{Preparatory notes}\label{bhag:preparatory}
While the research outlined in Section~\ref{bhag:manifoldlearning} primarily focuses on studying the properties of the manifold that encapsulates the features in the data, we initiated a study in \cite{mauropap} on the approximation of functions on this manifold, using only the eigendecomposition of the Laplace-Beltrami operator. Subsequent developments in \cite{heatkernframe, mauropap, eignet,  compbio, modlpmz} demonstrated that the full strength of the differentiability structure on the manifold is not necessary for this purpose. After several efforts to extract the abstract features required for developing localized kernels and operators, which are essential for deriving analogues of the results presented in Section~\ref{bhag:trigapprox}, our current understanding of the setup is as follows. Much of the discussion in this section draws from \cite{mhaskar2020kernel, mhaskar2023ryan}, although we will focus on compact spaces in this context.

Let $\XX$ be a connected,  compact, metric space with metric $\rho$. For $r>0$, $x\in\XX$, we denote
$$
\mathbb{B}(x,r)=\{y\in\XX : \rho(x,y)\le r\}, \ \Delta(x,r)=\mathsf{closure}(\XX\setminus \mathbb{B}(x,r)).
$$
If $K\subseteq \XX$ and $x\in\XX$, we write as usual $\rho(K,x)=\inf_{y\in K}\rho(y,x)$. 
It is convenient to denote the set\\
 $\{x\in\XX; \rho(K,x)\le r\}$ by $\mathbb{B}(K,r)$.
The diameter of $K$ is defined by $\mathsf{diam}(K)=\sup_{x,y\in K}\rho(x,y)$.

We fix a non-decreasing sequence $\{\lambda_k\}_{k=0}^\infty$, with $\lambda_0=0$ and $\lambda_k\uparrow\infty$ as $k\to\infty$. 
We also fix  a probability measure $\mu^*$ on $\XX$, and a system of orthonormal  functions $\{\phi_k\}_{k=0}^\infty\subset  C_0(\XX)$, such that $\phi_0(x)\equiv 1$ for all $x\in \XX$. 
We define
\be\label{eq:diffpolyclass}
\Pi_n=\mathsf{span}\{\phi_k : \lambda_k <n\}, \qquad n>0.
\ee
It is convenient to write $\Pi_n =\{0\}$ if $n\le 0$ and $\Pi_\infty =\bigcup_{n>0}\Pi_n$.
It will be assumed in the sequel that $\Pi_\infty$ is dense in $C_0$ (and hence, in every $L^p$, $1\le p <\infty$). 
The notation $X^p$ will mean $L^p$ if $1\le p<\infty$ and $C_0$ if $p=\infty$.
We will often refer to the elements of $\Pi_\infty$ as \textbf{diffusion polynomials} in keeping with \cite{mauropap}.
\yadi{$\XX$}{Data space, manifold}
\yadi{$\rho$}{Metric, geodesic distance}
\yadi{$\{\lambda_k\}$}{Sequence simulating eigenvalues}
\yadi{$\{\phi_k\}$}{Sequence of orthonormal functions, simulating eigenfunctions}
\yadi{$\BB(x,r)$, $\BB(X,r)$}{Ball, tubular neighborhood}
\begin{definition}\label{def:ddrdef}
The  tuple $\Xi=(\XX,\rho,\mu^*, \{\lambda_k\}_{k=0}^\infty, \{\phi_k\}_{k=0}^\infty)$ is called a \textbf{(compact) data space} if 
each of the following conditions is satisfied.
\begin{enumerate}
\item For each $x\in\XX$, $r>0$, $\mathbb{B}(x,r)$ is compact.
\item (\textbf{Ball measure condition}) There exist $q\ge 1$ and $\kappa>0$ with the following property: For each $x\in\XX$, $r>0$,
\be\label{eq:ballmeasurecond}
\mu^*(\mathbb{B}(x,r))=\mu^*\left(\{y\in\XX: \rho(x,y)<r\}\right)\le \kappa r^q.
\ee
(In particular, $\mu^*\left(\{y\in\XX: \rho(x,y)=r\}\right)=0$.)
\item (\textbf{Gaussian upper bound}) There exist $\kappa_1, \kappa_2>0$ such that for all $x, y\in\XX$, $0<t\le 1$,
\be\label{eq:gaussianbd}
\left|\sum_{k=0}^\infty \exp(-\lambda_k^2t)\phi_k(x)\phi_k(y)\right| \le \kappa_1t^{-q/2}\exp\left(-\kappa_2\frac{\rho(x,y)^2}{t}\right).
\ee
\end{enumerate}
We refer to $q$ as the \textbf{exponent} for $\Xi$.
\end{definition}

The primary example of a data space is, of course, a Riemannian manifold, satisfying \eqref{eq:ballmeasurecond}. 
We have proved in  \cite[Appendix~A]{mhaskar2020kernel} that the Gaussian upper bound is satisfied. 

We define the smoothness class $W_{p;\gamma}$ to be the class of all functions $f\in L^p$ for which
\be\label{eq:dataspacesobol}
\|f\|_{p;\gamma}=\|f\|_p+\sup_{j\ge 0}2^{j\gamma}\mathsf{dist}(L^p;f, \Pi_{2^j}) <\infty.
\ee

\subsection{Approximation and analysis of functions}\label{bhag:manifold_approx}
Next, we define various operators of reconstruction and analysis.
For any compactly supported function $H:\RR\to\RR$, we write
\be\label{eq:dataspace_kern_gen}
\Phi_n(H;x,y)=\sum_{k=0}^\infty H\left(\frac{\lambda_k}{n}\right)\phi_k(x)\phi_k(y), \qquad n>0,\ x, y\in\XX,
\ee
and the corresponding operator
\be\label{eq:dataspace_operator_gen}
\sigma_n(H;f)(x)=\int_\XX \Phi_n(H;x,y)f(y)d\mu^*(y), \qquad f\in L^1,\ x\in\XX.
\ee
Clearly, $\sigma_n(H;f)\in\Pi_n$.
Writing
\be\label{eq:dataspace_fourcoeff}
\hat{f}(k)=\int_\XX f(y)\phi_k(y)d\mu^*(y),
\ee
we have
\be\label{eq:dataspace_operator_bis}
\sigma_n(H;f)(x)=\sum_{k: \lambda_k<n} H\left(\frac{\lambda_k}{n}\right)\hat{f}(k)\phi_k(x), \qquad f\in L^1, \ n>0,\ x\in\XX.
\ee
In \cite{tauberian}, we have proved a general theorem, which implies in this case, the following localization estimate for the kernels.
\begin{theorem}\label{theo:dataspacelocal}
Let $H\in C^\infty(\RR)$ be an even, compactly supported function. 
Then for any $S\ge q+2$, we have
\be\label{eq:dataspacelocal}
|\Phi_n(H;x,y)|\ls \frac{n^q}{\max(1, (n\rho(x,y))^S)}, \qquad n\ge 1, \ x,y\in\XX.
\ee
Consequently,
\be\label{eq:dataspacelebesgue}
\sup_{x\in\XX}\int_{\XX}|\Phi_n(H;x,y)|d\mu^*(y)\ls 1.
\ee
In both of the above estimates,  the constant involved in $\ls$ depends upon $H$ and $S$ (but not on $n, x,y$).
\end{theorem}
In the sequel, we let $h\in C^\infty(\RR)$ be an even function, satisfying $h(t)=1$ if $|t|\le 1/2$ and $h(t)=0$ if $|t|\ge 1$. 
It is also convenient to require that $h$ be non-increasing on $[0,\infty)$. 
All constants may depend upon $h$.
With the function $h$, we define $g(t)=h(t)-h(2t)$, $g^*(t)=\sqrt{g(t)}$, $\tilde{g}(t)=h(\frac{t}{2})-h(4t)$.
Correspondingly, we define
\be\label{eq:dataspace_tauop}
\begin{aligned}
\tau_j(f)&=\begin{cases}
\sigma_1(h;f), &\mbox{ if $j=0$},\\
\sigma_{2^j}(g;f)=\sigma_{2^j}(h;f)-\sigma_{2^{j+1}}(h;f), &\mbox{ if $j=1,2,\cdots$},
\end{cases}\\
\tau_j^*(f)&=\begin{cases}
\sigma_1(\sqrt{h};f), &\mbox{ if $j=0$},\\
\sigma_{2^j}(g^*;f), &\mbox{ if $j=1,2,\cdots$}.
\end{cases}
\end{aligned}
\ee
We have the following analogue of Theorem~\ref{theo:characterization} (cf. \cite{mhaskar2020kernel, mauropap}).

\begin{theorem}\label{theo:dataspace_character}
Let $1\le p\le \infty$, $f\in X^p$. We have
\begin{enumerate}
\item[(a)]
\be\label{eq:dataspace_goodapprox}
\mathsf{dist}(L^p;f, \Pi_n)\le \|f-\sigma_n(h;f)\|_p\ls \mathsf{dist}(L^p;f, \Pi_{n/2}).
\ee
\item[(b)]
\be\label{eq:dataspace_paley}
\begin{aligned}
f&=\sum_{j=0}^\infty \tau_j(f)=\int_\XX \tau_0(f)(y)\Phi_1(h;\circ,y)d\mu^*(y)+\sum_{j=0}^\infty\int_\XX \tau_j(f)(y)\Phi_{2^j}(\tilde{g};\circ,y)d\mu^*(y)\\
&=\int_\XX \tau_0^*(f)(y)\Phi_1(\sqrt{h};\circ,y)d\mu^*(y)+\sum_{j=0}^\infty\int_\XX \tau_j^*(f)(y)\Phi_{2^j}(g^*;\circ,y)d\mu^*(y),
\end{aligned}
\ee
where the convergence is in the sense of $X^p$.
\item[(c)] If $f\in L^2$ then
\be\label{eq:dataspace_frame}
\|f\|_2^2=\sum_{j=0}^\infty \|\tau_j^*(f)\|_2^2 \sim \sum_{j=0}^\infty \|\tau_j(f)\|_2^2.
\ee
\item[(d)] If $\gamma>0$ then
\be\label{eq:dataspace_char}
\|f\|_{p;\gamma}\sim \|f\|_p+\sup_{j\ge 0} 2^{j\gamma}\|f-\sigma_{2^j}(f)\|_p \sim \|f\|_p + \sum_{j= 0}^\infty 2^{j\gamma}\|\tau_j(f)\|_p \sim \|f\|_p + \sum_{j= 0}^\infty 2^{j\gamma}\|\tau_j^*(f)\|_p.
\ee
\end{enumerate}
\end{theorem}
\begin{remark}\label{rem:dataspace_char}
In Theorem~\ref{theo:dataspace_character}, part (a) shows that $\sigma_n$ is an operator giving ``good approximation''. 
The first equation in part (b) gives a Paley-Wiener type dyadic expansion of $f$. The second and third equations give  frame expansions. 
After discretization, values of $\tau_j(f)$ (respectively, $\tau_j^*(f)$) would appear as frame coefficients.
The novelty of this approach, compared to classical wavelet expansions, lies in three key aspects: 
\begin{enumerate}
\item It is applicable to a general data space rather than being restricted to $L^2$ of a Euclidean space.
\item The convergence is characterized in the $X^p$ sense for \textbf{any} $p \in [1, \infty]$, which includes uniform convergence for functions in $C_0$.
\item In wavelet analysis, one starts with a large number of samples of $f$, and calculates the wavelet coefficients using two-scale relations.
As we will see in Section~\ref{bhag:local}, the details $\tau_j(f)$ are computed for higher and higher values of $j$ using a larger and larger number of values of $f$, starting with a coarse approximation $\tau_1(f)$ based on a small number of samples.
\end{enumerate}  
Part (c) discusses the use of frame properties of the operators $\tau_j$ and  $\tau_j^*$. Part (d) provides a characterization of the spaces $W_{p;\gamma}$ in terms of the various operators. A similar characterization for Besov spaces is straightforward to prove, and we omit it here to maintain clarity in the exposition. \qed
\end{remark}

\subsection{Monte-Carlo discretization}\label{bhag:montecarlo}

In the context of machine learning, we have to deal with data of the form $\{(x_j, y_j=f(x_j)+\epsilon_j)\}$, where $x_j\in\XX$, $f:\XX\to\RR$ is continuous, and $\epsilon_j$ is a realization of a random variable with unknown distribution. 
The key distinction from classical approximation theory, as described in Section~\ref{bhag:trigapprox}, is that in manifold learning, we do not have the flexibility to choose the locations of the points $ x_j$. Instead, it is customary to assume that the points $x_j$ are distributed according to the normalized Riemannian volume measure of the manifold. The discussion in this section is based on \cite{mhaskar2020kernel}.
The notation $\|\cdot\|$ in this section denotes the uniform norm $\|\cdot\|_\infty$.

We adopt a more general setup for data spaces. Specifically, we consider a distribution on $\XX$ of the form $d\nu^* = f_0 d\mu^*$, where $f_0 \in C(\XX)$, and assume that the points $x_j$ are drawn from this distribution.
We assume that points $(x_j,y_j)$ are drawn from an  unknown distribution $\tau$ on $\XX\times\RR$ such that $\nu^*$ is the marginal distribution of $\tau$ on $\XX$. 
Then we define $f:\XX\to\RR$ by the formula
\be\label{eq:function_as_expectation}
f(x)=\mathbb{E}_\tau(y|x).
\ee
Rather than treating $y_j = f(x_j) + \epsilon_j$, it is more convenient to consider $y = \mathcal{F}(x, \epsilon)$, where $(x, \epsilon)$ is a random variable with $\tau$ as the distribution. This formulation allows for the incorporation of non-additive noise. Furthermore, it aligns with the perspective in numerical analysis, where the observed value of $y_j$ is viewed as the exact value of the target function evaluated at a different point.
In this case, an additional advantage is that this different point does not need to lie on $\XX$, although we need to know what point $x_j\in\XX$ is supposed to correspond to this perturbation.
Thus, our training data is $\mathcal{D}=\{(x_j,\mathcal{F}(x_j,\epsilon_j))\}_{j=1}^M$.  
We define
\be\label{eq:data_based_approx}
\hat{\mathcal{F}}_n(\mathcal{D})(x)=\frac{1}{M}\sum_{j=1}^M \mathcal{F}(x_j,\epsilon_j)\Phi_n(x,x_j).
\ee
\begin{remark}
We note that our approximation $\hat{\mathcal{F}}(\mathcal{D})$
is truly universal; making absolutely no assumptions on the data in order to construct it.
Moreover, its construction is simple and does not involve any optimization.
We consider this to be a remarkable achievement in the case when the manifold assumption is satisfied and we can construct the required eigendecomposition. \qed
\end{remark}

A valuable tool for deriving the discretization of the integral operator in \eqref{eq:dataspace_operator_gen} is provided by concentration inequalities. The following theorem presents one such inequality.
\yadi{$B_n$}{Constant in Bernstein-Lipschitz inequality, Bernstein inequality}
\begin{theorem}
\label{theo:bernstein_concentraion}
(\textbf{Bernstein concentration inequality}) Let $Z_1,\cdots, Z_M$ be independent real valued random variables such that for each $j=1,\cdots,M$, $|Z_j|\le R$, and $\mathbb{E}(Z_j^2)\le V$. Then for any $t>0$,
\be\label{eq:bernstein_concentration}
\mathsf{Prob}\left( \left|\frac{1}{M}\sum_{j=1}^M (Z_j-\mathbb{E}(Z_j))\right| \ge t\right) \le 2\exp\left(-\frac{Mt^2}{2(V+Rt)}\right).
\ee
\end{theorem}

We would like to apply the Bernstein concentration inequality with $Z_j=y_j\Phi_n(x, x_j)$ for each $x\in\XX$. 
However, this would get us a bound depending upon $x$. 
To obtain a uniform bound, we need an argument based on covering numbers, for which we use the Bernstein-Lipschitz condition:

\begin{definition}\label{def:bernstein}
We say that the system $\Xi$ satisfies \textbf{Bernstein-Lipschitz condition} if for every $n>0$, there exists $B_n>0$ such that
\be\label{eq:bernstein_der}
|P(x)-P(y)|\le B_n\rho(x,y)\|P\|_\infty, \qquad x, y\in\XX, \ P\in \Pi_n.
\ee
\end{definition}
In the case when $\XX$ is a smooth compact manifold, and $\phi_n$'s are the eigenfunctions of the Laplace-Beltrami operator, this inequality is shown with $B_n\sim n$ in \cite{frankbern}.

The following theorem is a simple consequence of Theorem~8.1 \cite{mhaskar2020kernel} and Theorem~\ref{theo:dataspace_character}(a).
\begin{theorem}\label{theo:probapprox}
Let  $\tau$, $\nu^*$,  $\mathcal{F}$, $f$ be as described above. We assume the Bernstein-Lipschitz condition. 
Let $0<\delta<1$, $\gamma>0$, and $ff_0\in W_{\infty;\gamma}$.
 There exist constants $c_1, c$, such that if $M\ge c_1n^{q+2\gamma}log(cnB_n/\delta)$, and $\{(x_1,\epsilon_1), \cdots, (x_M, \epsilon_M)\}$ is a random sample from $\tau$, then
\be\label{eq:probapprox}
\mathsf{Prob}_{\nu^*}\left(\left\{\left\|\hat{\mathcal{F}}(\mathcal{D})-\sigma_n(\nu^*;f)\right\|_\infty \gs \frac{\|\mathcal{F}\|_{\infty,\XX\times\Omega}\sqrt{\|f_0\|_\XX}}{n^\gamma}\right\}\right)\le \delta.
\ee
\end{theorem}

Together with the results in Section~\ref{bhag:manifold_approx}, we get the following bounds on the approximation of $ff_0$ by $\hat{\mathcal{F}}(\mathcal{D})$.
\begin{corollary}\label{cor:ddrapprox}
With the setup as in Theorem~\ref{theo:probapprox}, we have for $M\gs n^{q+2\gamma}\log(cnB_n/\delta)$, 
\be\label{eq:ddrapprox}
\mathsf{Prob}_{\nu^*}\left(\left\{\left\|\hat{\mathcal{F}}(\mathcal{D})-ff_0\right\|_\infty \gs \frac{\|\mathcal{F}\|_{\infty,\XX\times\Omega}\sqrt{\|f_0\|_\XX}+\|ff_0\|_{\infty;\gamma}}{n^\gamma}\right\}\right)\le \delta.
\ee
\end{corollary}
\begin{remark}\label{rem:trigcase}
In the case when $\XX=\TT^q$ and $\nu^*$ is the Lebesgue measure on $\TT^q$, normalized to be a probability measure as in Theorem~\ref{theo:directsoltrig}, we have $B_n=n$, $f_0\equiv 1$. 
Then Corollary~\ref{cor:ddrapprox} reduces to Theorem~\ref{theo:directsoltrig}. \qed
\end{remark}
\begin{remark}\label{rem:probest}
If we assume that the points $x_j$ are sampled from the distribution $\mu^*$, then $f_0 \equiv 1$, and the above theorem provides an estimate on the degree of approximation of $f$. Conversely, if we apply the theorem with $f \equiv 1$, we obtain an estimate on the density $f_0$. Common methods for estimating the density typically use positive kernels for this purpose. However, it is well known in approximation theory that such methods are subject to a phenomenon called saturation; i.e., the degree of approximation cannot exceed $\mathcal{O}(n^{-2})$ unless $f_0$ is trivial. In contrast, our kernel guarantees that there is no such saturation. \qed
\end{remark}

\subsection{Local approximation}\label{bhag:local}

One of the major challenges in machine learning is the large volume of data required to train a system. A potential solution to this problem is to design systems that only utilize data in the vicinity of the point where a prediction is needed. Mathematically, this involves setting the target function to zero outside a neighborhood of the point in question. However, such hard thresholding destroys the smoothness of the function, thereby negatively impacting the degree of approximation. In our context, without relying on a differentiability structure, we define the set $C(\XX)$ of \textit{infinitely differentiable} functions by
\be\label{eq:cinfty}
C^\infty(\XX)=\bigcap_{\gamma>0} W_{\infty;\gamma}.
\ee
For  a ball $\BB\in\XX$, we will say that $\phi\in C^\infty(\BB)$ to mean that $\phi\in C^\infty(\XX)$ and is supported on $\BB$. 
We may now define the local smoothness class $W_{p;\gamma}$ at $x_0\in\XX$ to be the set of $f\in L^p(\XX)$ such that there exists $r>0$ with the property that $f\phi\in C^\infty(\BB(x_0,r))$ for every $\phi\in C^\infty(\BB(x_0,r))$. 
Theorem~\ref{theo:dataspace_character} can be applied in a straightforward manner to obtain characterization of local smoothness classes based on data in a neighborhood of $x_0$ simply by replacing $f$ by $f\phi$ for some $\phi\in C^\infty(\BB(x_0,r))$ that is equal to $1$ on, say $\BB(x_0,r/3)$, assuming that such a function exists.

A more fundamental question is whether the approximation $\sigma_n(f)$ can automatically adjust its approximation power according to the local smoothness of $f$ at each point. To investigate this question within our abstract framework, we need to introduce additional assumptions. A particularly useful property of smooth manifolds is the existence of a partition of unity. In our abstract setting, we define the following:

\begin{definition}\label{def:partionunity}
(\textbf{Partition of unity})
For every $r>0$, there exists a  countable family $\mathcal{F}_r=\{\psi_{k,r}\}_{k=0}^\infty$ of functions in $C^\infty$ with the following properties:
\begin{enumerate}
\item Each $\psi_{k,r}\in \mathcal{F}_r$ is supported on $\mathbb{B}(x_k,r)$ for some $x_k\in\XX$.
\item For every $\psi_{k,r}\in\mathcal{F}_r$ and $x\in\XX$, $0\le \psi_{k,r}(x)\le 1$.
\item For every $x\in \XX$, there exists a finite subset $\mathcal{F}_r(x)\subseteq \mathcal{F}_r$ such that
\be\label{partitionsum}
\sum_{\psi_{k,r}\in \mathcal{F}_r(x)}\psi_{k,r}(y)=1, \qquad y\in\mathbb{B}(x,r).
\ee
 \end{enumerate}
\end{definition}

We note that if the partition of unity holds, then for any $x_0\in\XX$ and $r>0$, there exists $\phi \in C^\infty(\BB(x_0,r))$ such that $\phi(x)=1$ for all $x\in \BB(x_0,r/3)$. 

We need a further technical assumption.

\begin{definition}\label{def:prodassumption}
(\textbf{Product assumption})
There exists $A^*\ge 1$ and a family $\{R_{j,k,n}\in\Pi_{A^*n}\}$ such that for every $S>0$,
\be\label{weakprodass}
\lim_{n\to \infty}n^S\left(\max_{\lambda_k,\lambda_j <n,\ p=1,\infty}\|\phi_k\phi_j-R_{j,k,n}\|_p\right) =0.
\ee
We say that an \textbf{strong product assumption} is satisfied if instead of \eref{weakprodass}, we have
for every $n>0$ and $P, Q\in\Pi_n$, $PQ\in \Pi_{A^*n}$.
\end{definition}

In \cite[Theorem~6.1]{geller2011band}, Geller and Pesenson demonstrated that the strong product assumption (and, by extension, the product assumption) holds for compact homogeneous manifolds. We extended this result in \cite[Theorem~A.1]{modlpmz} to encompass the case of eigenfunctions associated with general elliptic partial differential operators on arbitrary compact, smooth manifolds, under the condition that the coefficient functions of the operator satisfy specific technical requirements.

We proved the following theorem in \cite{mhaskar2020kernel}.

\begin{theorem}
\label{theo:dataspace_localchar}
Let the partition of unity and product assumption hold. Let  $x_0\in \XX$, $\gamma>0$, $1\le p\le \infty$ and $f\in X^p$. 
Then  the following statements are equivalent.\\
{\rm (a)} $f\in W_{p;\gamma}(x_0)$ \\
{\rm (b)} There exists a ball $\BB\ni x_0$ such that
\be\label{eq:sigma_localchar}
\sup_{j\ge 0}2^{j\gamma}\|f-\sigma_{2^j}(f)\|_{p,\BB} <\infty.
\ee
{\rm (c)} There exists a ball $\BB\ni x_0$ such that
\be\label{eq:tau_localchar}
\sup_{j\ge 0}2^{j\gamma}\|\tau_j(f)\|_{p,\BB} <\infty.
\ee
\end{theorem}

Thus, the approximation power of the operators $\sigma_{2^j}$ adapts automatically to the \textbf{local} smoothness of the target function. This local smoothness is \textbf{characterized} by the local norms of the analysis operators $\tau_j$.

To derive a discretized version of Theorem~\ref{theo:dataspace_character}, we define - similarly to Section~\ref{bhag:trigapprox} - the concept of Marcinkiewicz-Zygmund quadrature measures. 

\begin{definition}[\textbf{Marcinkiewicz-Zygmund (MZ) measures}]\label{def:mz-measure}
Let $n\ge 1$. A measure $\nu$ will be called a \textbf{quadrature measure} of order $n$ if
\be\label{eq:data_quadrature}
\int_{\XX} P(\x)d\nu(\x)=\int_{\XX} P(\x)d\mu_q^*(\x), \qquad P\in \Pi_n.
\ee
The measure $\nu$ will be called a \textbf{Marcinkiewicz-Zygmund (MZ) measure} of order $n$ if
\be\label{eq:data_mzineq}
\int_{\XX} |P(\x)|d|\nu|(\x)\le \tn\nu\tn\int_{\XX} |P(\x)|d\mu_q^*(\x), \qquad P\in \Pi_n,
\ee
where $\tn\nu\tn$ is understood to be the smallest constant that works in \eqref{eq:data_mzineq}.
The set of all MZ quadrature measures or order $n$ will be denoted by $MZQ(n)$.
A sequence $\bs\nu=\{\nu_n\}$ of measures will be called an MZQ sequence if each $\nu_n\in MZQ(n)$ with $\tn\nu_n\tn\lesssim 1$. 
\end{definition}

If $\mathcal{C} \subset \XX$, we define the \emph{mesh norm} $\delta(\mathcal{C})$ (also known as fill distance, covering radius, or density content) and the \emph{minimal separation} $\eta(\mathcal{C})$ by
\begin{equation}\label{eq:data_meshnormdef}
\delta(\mathcal{C}) = \sup_{x \in \XX} \inf_{y \in \mathcal{C}} \rho(x,y), \qquad 
\eta(\mathcal{C}) = \inf_{\substack{x, y \in \mathcal{C} \\ x \not= y}} \rho(x,y).
\end{equation}
It can be shown that if $\mathcal{C}$ is finite, then the measure that assigns the mass $\eta(\mathcal{C})^q$ to each point in $\mathcal{C}$ is an MZ measure of order $\eta(\mathcal{C})^{-q}$ (Lemma~5.3 in \cite{eignet}).

In order to prove the existence of such MZ quadrature measures in the absence of a differentiable structure, we assume the Bernstein-Lipschitz condition stated in Definition~\ref{def:bernstein}. The following result can then be established by a slight variation of the proof of Theorem~\ref{theo:mzqtheo}.

\begin{theorem}
\label{theo:data_mzqtheo}
We assume the Bernstein-Lipschitz condition. 
There exists a constant $\alpha>0$ with the following property: If $n\ge 1$ and $\C\subseteq\XX$ satisfies $\delta(\C)\le \alpha/n$, then there exists a measure $\nu\in MZQ(n)$ supported on $\C$, with $\tn\nu\tn\lesssim 1$.
\end{theorem}
We have shown in \cite{mhaskar2020kernel} that if $\mathcal{C}$ is a random sample of size $M \gs n^q\log n$, then $\delta(\mathcal{C})\ls 1/n$ with high probability.

We may now define the reconstruction and analysis operators exactly as in Section~\ref{bhag:trigapprox}.
If $\nu$ is any measure, we may define
\be\label{eq:data_sigmadef}
\sigma_n(\nu;f)(x)=\int_{\XX}f(y)\Phi_n(x,y)d\nu(y).
\ee
We note that $\sigma_n(\nu;f)$ is the same as $\sigma_n(\mu^*;f)$, except that the integrals defining the coefficients $\hat{f}(\ell)$ are evaluated approximately using the integrals with respect to $\nu$.

For a sequence $\bs\nu$ of measures, we define 
\be\label{eq:data_disctaudef}
\tau_j(\bs\nu;f)=\begin{cases}
\sigma_1(\nu_1;f), &\mbox{ if $j=1$},\\
\sigma_{2^j}(\nu_{2^j};f)-\sigma_{2^{j-1}}(\nu_{2^{j-1}};f), &\mbox{ if $j=1,2,\cdots$}.
\end{cases}
\ee 
We define $\tilde{g}(t)=h(t/2)-h(4t)$, and $\widetilde{\Psi}_j$ by
\be\label{eq:data_psidef}
\widetilde{\Psi}_j(x,y)=\sum_\ell \tilde{g}\left(\frac{\lambda_\ell}{2^j}\right)\phi_\ell(x)\phi_\ell(y).
\ee
The discretized version of Theorem~\ref{theo:dataspace_localchar} is the following.
For simplicity, we assume the strong product assumption.
Some of the following statements hold with only the weak product assumption.

\begin{theorem}\label{theo:data_disc_local_char}
We assume the partition of unity and the strong product assumption.
Let $\bs\nu$ be a sequence of measures such that each $\nu_j\in MZQ(A^*2^j)$, $f\in C(\XX)$.\\
{\rm (a)} We have
\be\label{eq:data_disc_goodapprox}
\mathsf{dist}(C(\XX));f,\Pi_{2^j})\le \|f-\sigma_{2^j}(\nu_j;f)\|_\infty\ls \mathsf{dist}(C(\XX));f,\Pi_{2^j/A^*}).
\ee
{\rm (b)} We have
\be\label{eq:data_disc_paley}
f(x)=\sum_{j=0}^\infty \int_\XX \tau_j(\nu_j;f,y)\widetilde{\Psi}_j(x,y)d\nu_j(y).
\ee
{\rm (c)} Let $\gamma>0$, $x_0\in\XX$.
The following statements are equivalent.
\begin{itemize}
\item $f\in W_{\infty;\gamma}$.
\item There exists a ball $\BB\ni x_0$ such that
\be\label{eq:data_disc_loc_approx}
\sup_{j\ge 0} 2^{j\gamma}\|f-\sigma_{2^j}(\nu_j;f)\|_{\infty,\BB} <\infty.
\ee
\item There exists a ball $\BB\ni x_0$ such that
\be\label{eq:data_disc_tau_localchar}
\sup_{j\ge 0}2^{j\gamma}\|\tau_j(\nu_j;f)\|_{\infty,\BB} <\infty.
\ee
\end{itemize}
\end{theorem}
\begin{remark}\label{rem:disc_paley}
We observe that the expansion in \eqref{eq:data_disc_paley} is a doubly discrete expansion, serving as a complete analogue of \eqref{eq:supwaveexp}. The measure $\nu_j$ depends on the points where the target function is sampled. For simplicity, we have expressed the integral with respect to $\nu_j$ without introducing additional complexity in the notation. However, it is worth noting that the integration in \eqref{eq:data_disc_paley} may well be with respect to a different measure, potentially one that allows for faster evaluation, similar to the case in \eqref{eq:supwaveexp}. \qed
\end{remark}
\begin{remark}\label{rem:num_examples}
{\rm
In the context of trigonometric polynomial approximation, Theorem~\ref{theo:data_disc_local_char} has been used in \cite{loctrigwave, trigwave} for the detection of singularities of periodic function.
Illustrations in the case of the sphere are given in \cite{quadconst}. 
The operator $\sigma_n$ does not provide the best error in the sense of either $L^2$ or supremum norm.
However, the localization of the kernels implies that if the function is smooth on all but a small subset of the sphere, then the percentage of points where the error is small is substatnially better than classical methods such as least square approximation.
for example, adapting an example in \cite{quadconst}, we consider the target function $f(x,y,z)=(x-0.7)_{+}^{5/6} + (z-0.7)_{+}^{5/6}$, $(x,y,z)\in \SS^2$.
For the least square approximation, the error is $<10^{-3}$ at $92.83\%$ test points on $\SS^2$, the corresponding percentage for the reconstruction operator is only $88.89\%$. 
The situation is drastically different when smaller errors are contemplated: least square error is $<10^{-5}$ at only 
$3.47\%$ of the test points, the corresponding percentage for the reconstruction operator is $54.31\%$. 
\qed}
\end{remark}
\subsection{Kernel based approximation}\label{bhag:kernelbased}
Kernel-based approximation is one of the oldest methods of approximation. In modern terms, one of the earliest proofs of the Weierstrass approximation theorem, attributed to Lebesgue, employs translates of a linear $B$-spline. Radial basis functions (RBFs), or RBF networks, have been extensively studied for a long time, predating the widespread popularity of approximation theory through neural networks.
A \emph{radial basis function network} has the form $\x\mapsto \sum_k a_k\phi(|\x-\x_k|)$, where $\phi$ is a univariate function.
Some of the seminal papers in this direction are the following. In \cite{micchelli1984interpolation}, Micchelli showed that under certain conditions an RBF network can interpolate any data in any Euclidean space.
In \cite{park1991universal}, Park and Sandberg showed that RBF networks with $\phi(t)=\exp(-t^2/2)$ have universal approximation property.
In \cite{mhasmich}, we obtained necessary and sufficient conditions for an activation function $\phi$ to satisfy the universal approximation property.
In \cite{girosi1990networks}, Girosi and Poggio pointed out that the RBF networks, such as Nadarya-Watson estimator developed originally for probability density estimation can also be used for function approximation.

More generally, a \emph{kernel-based approximation} takes the form $ x \mapsto \sum_k a_k G(x, x_k) $, which enables approximation on any set $\Omega $, where $ G : \Omega \times \Omega \to \mathbb{R} $. We now briefly diverge from our discussion of approximation on data spaces to introduce concepts related to kernel-based approximation in a broader context.
One important reason why such approximations are popular is that in machine learning, one traditionally wants to measure the error in approximation by a Hilbert space $L^2(\mu^*,\Omega)$  norm, but the approximation has to be constructed from values of the target function.
This forces one to restrict the attention to a subspace $\mathcal{H}$ of the Hilbert space  on which function evaluations are  continuous functionals. 
In view of the Riesz representation theorem, this gives rise to a kernel $G$ and an inner product on $\mathcal{H}$ such that $f(x)=\langle G(x,\cdot), f\rangle_\mathcal{H}$ for every $f\in\mathcal{H}$ and $x\in\Omega$ \cite{aronszajn1950theory}. 
The subspace $\mathcal{H}$ is therefore called a \emph{reproducing kernel Hilbert space} (RKHS).
The kernel $G$ is usually assumed to be positive definite; i.e., assuming a topology on $\Omega$, for every Borel measure $\mu$ on $\Omega$,
$$
\int_\Omega\int_\Omega G(x,y)d\mu(x)d\mu(y)\ge 0
$$
with equality if and only if $\mu\equiv 0$.
The Golomb-Weienberger principle \cite{golomb1959optimal} states that for any set of points $\{x_j\}_{j=1}^M\subset \Omega$, and $\{y_j\}_{j=1}^M$, there is a unique $f\in \mathcal{H}$ such that $\langle f, f\rangle_{\mathcal{H}}$ is minimized subject to the conditions $f(x_j)=y_j$, $j=1,\cdots, M$. 
Moreover, this solution is given by $f(x)=\sum_{k=1}^K a_kG(x,x_k)$ for suitably chosen $a_k$'s.

There is a cannonical embedding $f\mapsto\mathcal{D}_G(f)$, $f\in \mathcal{H}$ so that
\be\label{eq:cannonical_embedding}
f(x)=\int_\Omega G(x,y)\mathcal{D}_G(f)(y)d\mu^*(y), 
\ee
and $\langle f, f\rangle_{\mathcal{H}} =\|\mathcal{D}_G(f)\|_{\mu^*}$.
There is a natural operator associated with $G$ given by
$$
\mathcal{T}_G(g)(x)=\int_\Omega G(x,y)g(y)d\mu^*(y), \qquad g\in L^2(2,\Omega).
$$
This is best understood using the spectral theorem for this operator.
Assuming that $\Omega$ is compact and $G$ is continuous, the operator is a Hilbert-Schmidt operator, and hence, one has a formal expansion (known as Mercer expansion):
\be\label{eq:mercer}
G(x,y)=\sum_{k=1}^\infty \Lambda_k^2\phi_k(x)\phi_k(y),
\ee
where $\{\phi_k\}$ is a complete orthonormal system in $L^2(\mu^*,\Omega)$.
In terms of this expansion, we have formally:
\be\label{eq:understanding}
\mathcal{D}_G(f)(x)=\sum_k \Lambda_k^{-2}\hat{f}(k)\phi_k(x),
\ee
where 
$$
\hat{f}(k)=\int_\Omega f(y)\phi_k(y)d\mu^*(y).
$$
Moreover,
\be\label{eq:rkhs_inner_product}
\langle f, g\rangle_{\mathcal{H}}=\sum_k \frac{\hat{f}(k)\hat{g}(k)}{\Lambda_k^2}.
\ee
The book by Cucker and Zhou \cite{zhoubk_learning} provides a comprehensive treatment of kernel-based approximation from the perspective of machine learning. More recently, Belkin \cite{belkin2019reconciling, belkin2021fit} has argued that a deep understanding of kernel-based approximation is essential for comprehending deep networks and feature selection. The literature on kernel-based approximation is vast and cannot be fully summarized here.

It is important to note that the theory of Reproducing Kernel Hilbert Spaces (RKHS) is relevant primarily when approximation is performed within a Hilbert space using the values of the function. However, we adopt the perspective that it is more productive to consider approximation in the uniform norm, where function evaluations are already well defined. In this context, the formula \eqref{eq:cannonical_embedding} defines a variation space (a reproducing kernel Banach space), where $\mathcal{D}_G(f)$ is also a continuous function.
We have already discussed the approximation properties of these spaces, where $\mathcal{D}_G(f)$ is even allowed to be a measure.
These results are not constructive.
Additionally, the variation space may be too restrictive, often consisting only of very smooth functions. In this section, we explore alternative approaches that allow for explicit constructions of arbitrary target functions and establish bounds on the approximation error for functions that are less smooth than those within the variation space. Most importantly, we present a surprising converse theorem that clarifies that the complexity of kernel-based approximations should be measured in terms of the minimal separation among the centers $x_k$, rather than the number of non-linearities involved.
The following discussion is based on \cite{eignet, sloanfest, mhaskar2020kernel}.

In particular, we now resume the set up of data spaces as before.
Our discussion is easier if we rewrite the Mercer expansion in the form
\be\label{eq:dataspacemercer}
G(x,y)=\sum_k  b(\lambda_k)\phi_k(x)\phi_k(y),
\ee
for a suitable function $b :[0,\infty)\to [0,\infty)$.
To emphasize the dependence on the ``eigenstructure'', we have called a function of the form $x\mapsto\sum_j a_jG(x,x_j)$ an \emph{eignet}.
It is clear that for any $P\in\Pi_\infty$, the equation
\be\label{eq:basic_eignet_eq}
P(x)=\int_\XX G(x,y)\mathcal{D}_G(P)(y)d\mu^*(y)
\ee
holds.
The approach involves discretizing the integral using a quadrature formula while carefully tracking the associated errors. This requires some preliminary preparation in our abstraction.

We begin with the \textit{smoothness} of \( G \), which enables us to estimate the approximation error for \( G \) in terms of diffusion polynomials, and consequently, the quadrature error.
Intuitively, $G$ is a kernel of type $\beta$ if $b(\lambda_k)\sim \lambda_k^{-\beta}$. 
For approximation in $L^2$, this condition is enough.
However, we need to be more careful for uniform approximation.

\begin{definition}\label{def:eigkerndef}
Let $\beta\in\mathbb{R}$. A function $b:\mathbb{R}\to[0,\infty)$ will be called a mask of type $\beta$ if $b$ is an even, $S$ times continuously differentiable function such that for $t>0$, $b(t)=(1+t)^{-\beta}F_b(\log t)$ for some $F_b:\mathbb{R}\to\mathbb{R}$ such that  $|F_b^{(k)}(t)|\le c(b)$, $t\in\mathbb{R}$, $k=0,1,\cdots,S$,  and $F_b(t)\ge c_1(b)$, $t\in\mathbb{R}$.     A function $G:\mathbb{X}\times\mathbb{X}\to \mathbb{R}$ will be called  a kernel of type $\beta$ if it admits a formal expansion $G(x,y)=\sum_{j=0}^\infty b(\lambda_j)\phi_j(x)\phi_j(y)$ for some mask $b$ of type $\beta>0$. If we wish to specify the connection between $G$ and $b$, we will write $G(b;x,y)$ in place of $G$.
\end{definition}

The definition of a mask of type $\beta$ can be relaxed to some extent. For instance, the various bounds on $ F_b$ and its derivatives may be assumed to hold only for sufficiently large values of $|t| $, rather than for all $t \in \mathbb{R}$. In such cases, it is possible to construct a new kernel by adding a suitable diffusion polynomial (of fixed degree) to $G$, as is commonly done in the theory of radial basis functions, thereby obtaining a kernel whose mask satisfies the definition provided above.
Similarly, the requirement that $b$ must be non-negative can be relaxed if both $b$ and $|b|$ satisfy the other conditions in Definition~\ref{def:eigkerndef}. In this case, we simply apply our results once to $2|b| + b$ and once to $2|b|$.
These adjustments do not introduce any new aspects to our theory. Therefore, we adopt the more restrictive definition provided above.
We state the following lemma, asserting the existence of a kernel of type $\beta$.

\begin{lemma}\label{lemma:eignet_funda}
Let $1\le p\le\infty$, $\beta>q/p'$,  $b$ be a mask of type $\beta$.
 For every $y\in\XX$, there exists $\psi_y:=G(\circ,y)\in X^p$ such that $\ip{\psi_y}{\phi_k} =b(\ell_k)\phi_k(y)$, $k=0,1,\cdots$. We have
\be\label{eq:glpuniform}
\sup_{y\in\XX}\|G(\circ,y)\|_p \ls 1.
\ee
\end{lemma}

To describe our theorems, we need a little bit further preparation.
The first lemma deals with the pruning of a sequence of data sets such that resulting data sets are minimally separated at the same scale as their mesh norms (recall the definition \eqref{eq:data_meshnormdef}).

\begin{lemma}\label{lemma:pruning}
Let $\{\C_m\}$ be a sequence of finite subsets of $\XX$, with $\delta(\C_m)\sim 1/m$, and $C_m\subseteq \C_{m+1}$, $m=1,2,\cdots$. Then there exists a sequence of subsets $\{\tilde\C_m \subseteq \C_m\}$, where, for $m=1,2,\cdots,$ $\delta(\tilde\C_m)\sim 1/m$, $\tilde \C_m\subseteq \tilde \C_{m+1}$, $\delta(\tilde\C_m)\le 2\eta(\tilde \C_m)$. 
\end{lemma}

We now assume a nested sequence $\{\C_m\}$ of data sets such that $\delta({\C_m})\sim \eta(\C_m)\sim 2^{-m}$, and such that there is a  measure $\nu_m\in MZQ(A^*2^m)$, supported on $\C_m$ for each $m=1,2,\cdots$.
For simplicity of exposition, we will be more explicit, and write
\be\label{eq:explicitmeasure}
\int_\XX f d\nu_m =\sum_{\xi\in\C_m} w_\xi f(\xi).
\ee
We define
\be\label{eq:eignetcont}
\mathbb{G}_m(f)(x)=\sum_{\xi\in\C_m}\left\{w_\xi \mathcal{D}_G(\sigma_{2^m}(\mu^*;f))(\xi)\right\}G(x,\xi).
\ee

In the case of approximation of continuous functions, we may use a discretization of \eqref{eq:basic_eignet_eq} with $P=\sigma_{2^m}(\nu_m;f)$ using $\nu_m$ itself to obtain the network:
\be\label{eq:eignetformula}
\mathbb{G}_{m,\infty}(f)(x)=\sum_{\xi\in \C_m}\left\{w_\xi\mathcal{D}_G(\sigma_{2^m}(\nu_m;f))(\xi)\right\}G(x,\xi).
\ee
Finally, we define a space of eignets
\be\label{eq:eignetspace}
\mathcal{G}(\C_m)=\mathsf{span}\{G(\circ,\xi) : \xi\in \C_m\}.
\ee

The following theorem is a rephrasing of (Theorem~3.3, \cite{eignet}).
The proof is verbatim the same.
We note that the theorem gives a complete characterization of the spaces $W_{p;\gamma}$ for $\gamma<\beta-q/p'$ in terms of approximation from $\mathcal{G}(\C_m)$.
One interesting aspect is that it is the minimal separation among the centers $\C_m$ which is the ``correct'' way to measure complexity rather than the number $|\C_m|$ of nonlinearities.
The other interesting aspect is that the constructions given in \eqref{eq:eignetcont} and \eqref{eq:eignetformula} are asymptotically optimal in the sense of an equivalence theorem, not just in the sense of widths.

\begin{theorem}
\label{theo:data_eignet_char}
Suppose that 
\be\label{eq:heatlowbd}
\sum_{k=0}^\infty \exp(-\lambda_k^2t)\phi_k(x)^2\gs t^{-\a/2}, \qquad x\in\XX,\ t\in (0,1].
\ee
Then the following are equivalent for each $\gamma$ with $0<\gamma<\beta-q/p'$:\\
{\rm (a)} $f\in W_{p;\gamma}$.\\
{\rm (b)} $\sup_{m\ge 1}2^{m\gamma}\|f-\mathbb{G}_m(f)\|_p <\infty$.\\
{\rm (c)} $\sup_{m\ge 1}2^{m\gamma}\dist(L^p;f,\mathcal{G}(\C_m)) <\infty$.\\
In the case when $p=\infty$, each of these assertions is also equivalent to \\
{\rm (d)} $\sup_{m\ge 1}2^{m\gamma}\|f-\mathbb{G}_{m,\infty}(f)\|_\infty<\infty$.
\end{theorem}

The following stability theorem underscores the importance of minimal separation in this theory.
\begin{theorem}\label{theo:coefftheo}
We assume that \eref{eq:heatlowbd} holds. Let $1\le p\le\infty$,  $\beta>q/p'$,  $\C\subset\XX$ be a finite set,  $a_y\in\RR$, $y\in\C$, and ${\bf a}=(a_y)_{y\in\C}$. Then
\be\label{coeffineq}
\|{\bf a}\|_{\ell^p}\le c\eta(\C)^{q/p'-\beta}\left\|\sum_{y\in\C} a_yG(\circ,y)\right\|_p.
\ee
\end{theorem}

The discussion in this section primarily focuses on eigenkernels whose singular values decay polynomially. Many important kernels, including the Gaussian, do not satisfy these conditions. A detailed analysis of such kernels is provided in \cite{mhaskar2020kernel}. While the converse theorem is not explicitly stated there, the proof of the converse theorem in \cite{eignet} can be adapted with only minor and straightforward modifications.
In particular, the converse theorem for Gaussian networks was established in \cite{convtheo}. For kernels with rapidly decaying singular values, a characterization of local smoothness is also possible using eigenets, as discussed in \cite{mhaskar2020kernel}.
 
\bhag{Approximation on the sphere}\label{bhag:sphere}

The theory developed in Section~\ref{bhag:dataspaces}  applies directly to the approximation and analysis of functions on the sphere. 
In fact, this was the original context in which most elements of the theory were first developed. 
Several notable contributions to the theory of spherical approximation are detailed in \cite{dai2013approximation, michel2012lectures, locsmooth}.

We now highlight two additional aspects that specifically require the nuances of the spherical domain.
One of them is approximation by ReLU$^{\gamma}$ networks, which we will discuss in Section~\ref{bhag:relunets}. 
The other is approximation of operators, which we will discuss in Section~\ref{bhag:operatorapprox}.

\subsection{Spherical polynomials}\label{bhag:sphpolynomials}
In this section, we review some facts regarding spherical polynomials.

The unit sphere of a Euclidean space $\RR^{q+1}$ is defined by
\be\label{eq:spheredef}
\SS^q =\{(x_1,\cdots, x_{q+1})\in\RR^{q+1}: x_1^2+\cdots +x_{q+1}^2=1\}.
\ee
A sphere is a prototypical and simple example of a manifold (and hence, a data space), of dimension $q$.
The distance between $x,y\in\SS^q$ is given by $\rho(x,y)=\arccos(x\cdot y)$.
In this section, we will denote by $\omega_q$ the Riemannian volume of $\SS^q$, and by $\mu_q^*$ the Riemannian volume measure, normalized to be a probability measure. 
More explicitly,
\be\label{eq:volumerecurs}
\omega_{q}= \frac{2\pi^{(q+1)/2}}{\Gamma((q+1)/2)}=
\begin{cases}
2\pi, &\mbox{ if $q=1$},\\[1ex]
\displaystyle\sqrt{\pi}\frac{\Gamma(q/2)}{\Gamma(q/2+1/2)}\omega_{q-1}, &\mbox{ if $q\ge 2$.}
\end{cases}
\ee
\yadi{$\SS^q$}{Euclidean unit sphere embedded in $\RR^{q+1}$}
\yadi{$\mu_q^*$}{Volume measure on $\SS^q$, normalized}
\yadi{$\omega_q$}{Volume of $\SS^q$}
\yadi{$Y_{\ell,k}$}{Orthonormalized spherical harmonics}
\yadi{$p_{\ell,q}$}{Orthonormalized ultraspherical polynomial of degree $\ell$, Definitions~\ref{eq:sphaddformula}, \ref{eq:ultraorthogonal}}
 By representing a point $x\in \mathbb{S}^q$ as $(x' \sin\theta,\cos\theta)$ for some $x'\in \mathbb{S}^{q-1}$, one has the recursive formula for measures
\be\label{eq:recursivemeasure}
\frac{\omega_q}{\omega_{q-1}} d\mu^*_q(x)=\sin^{q-1}(\theta)d\theta d\mu^*_{q-1}(x'),
\ee

The eigenvalues of the  negative Laplace-Betrami operator are $\lambda_\ell^2=\ell(\ell+q-1)$, $\ell=0,1,\cdots$. 
The space $\mathbb{H}_\ell^q$ of eigenfunctions corresponding to each $\lambda_\ell$ comprises restrictions to $\SS^q$ of homogeneous harmonic polynomials of total degree $\ell$ in $q+1$ variables. 
Interestingly, the space $\Pi_n^q$ of all \emph{spherical polynomials} of total degree $<n$; i.e., the space of restrictions to $\SS^q$ of all polynomials in $q+1$ variables of total degree $<n$, can be expressed as $\Pi_\ell^q=\oplus_{\ell=0}^{n-1}\HH_\ell^q$. 
Moreover, if $\{Y_{\ell,k}\}$ is any orthonormal basis for $\HH_\ell^q$ then we have the \emph{addition formula}:
\be\label{eq:sphaddformula}
\sum_{k=1}^{\dim (\mathbb{H}^q_\ell)} Y_{\ell,k}(x)Y_{\ell,k}(y)=\frac{\omega_q}{\omega_{q-1}}p_{\ell,q}(1)p_{\ell, q}(x\cdot y),
\ee
where  $p_{\ell,q}$ is a polynomial of degree $\ell$ with positive leading coefficients. 
The sequence $\{p_{\ell, q}\}$ satisfies
\be\label{eq:ultraorthogonal}
\int_{-1}^1 p_{\ell,q}(t)p_{m,q}(t)(1-t^2)^{q/2-1}dt =\delta_{\ell,m}.
\ee
In particular, the localized kernels are given by $(x,y)\mapsto \Phi_{n,q}(x\cdot y)$, where $\Phi_{n,q}$ is a \emph{univariate} polynomial defined by
\be\label{eq:sphlockern}
\Phi_{n,q}(t)=\frac{\omega_q}{\omega_{q-1}}\sum_{\ell=0}^{n-1} h\left(\frac{\ell}{n}\right)p_{\ell,q}(1)p_{\ell, q}(t).
\ee
We note that we may use $\ell$ instead of $\sqrt{\ell(\ell+q-1)}$ here. 
The localization property of these kernels takes the form: For any $S\ge 1$,
\begin{equation}
\label{eq:sphkernloc}
    |\Phi_{n,q}(x\cdot y)|\lesssim \frac{n^q}{\max(1,n\arccos(x\cdot y))^S},
\end{equation}
where the constant involved may depend upon $S$.

\begin{figure}[ht]
\begin{center}
\begin{minipage}{0.3\textwidth}
\includegraphics[width=\textwidth]{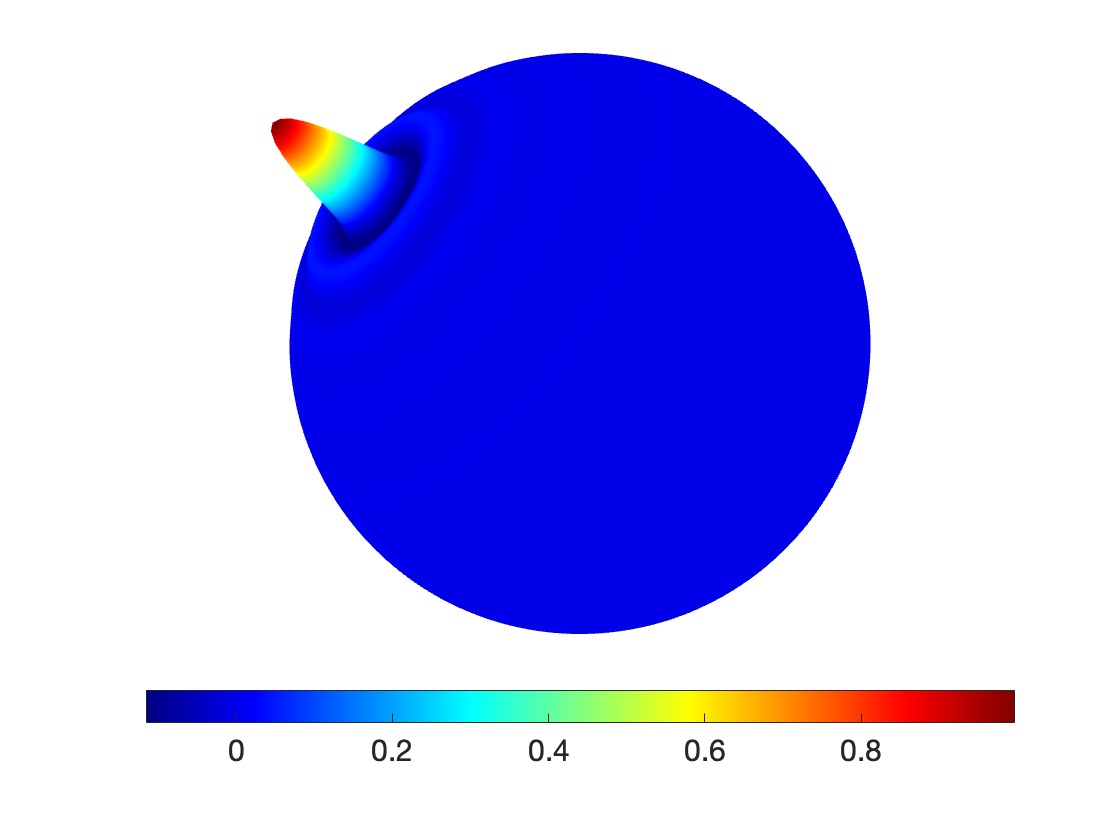} 
\end{minipage}
\begin{minipage}{0.3\textwidth}
\includegraphics[width=\textwidth]{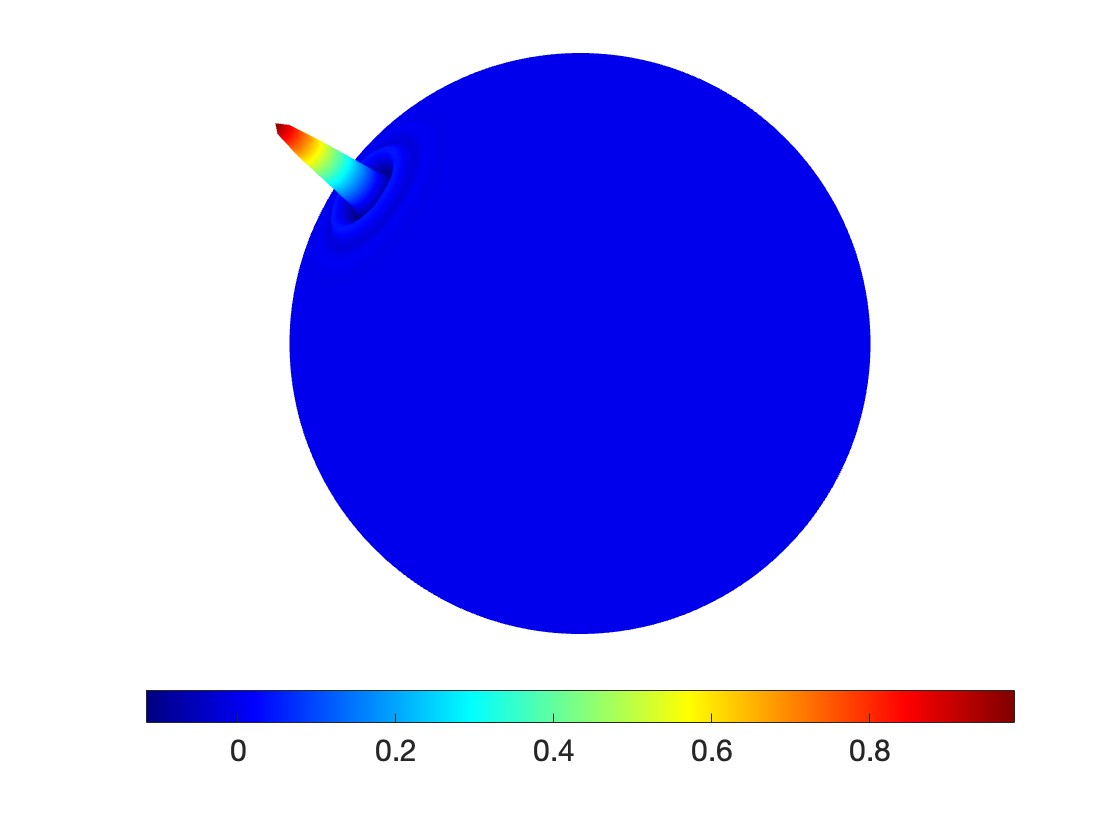} 
\end{minipage}
\begin{minipage}{0.3\textwidth}
\includegraphics[width=\textwidth]{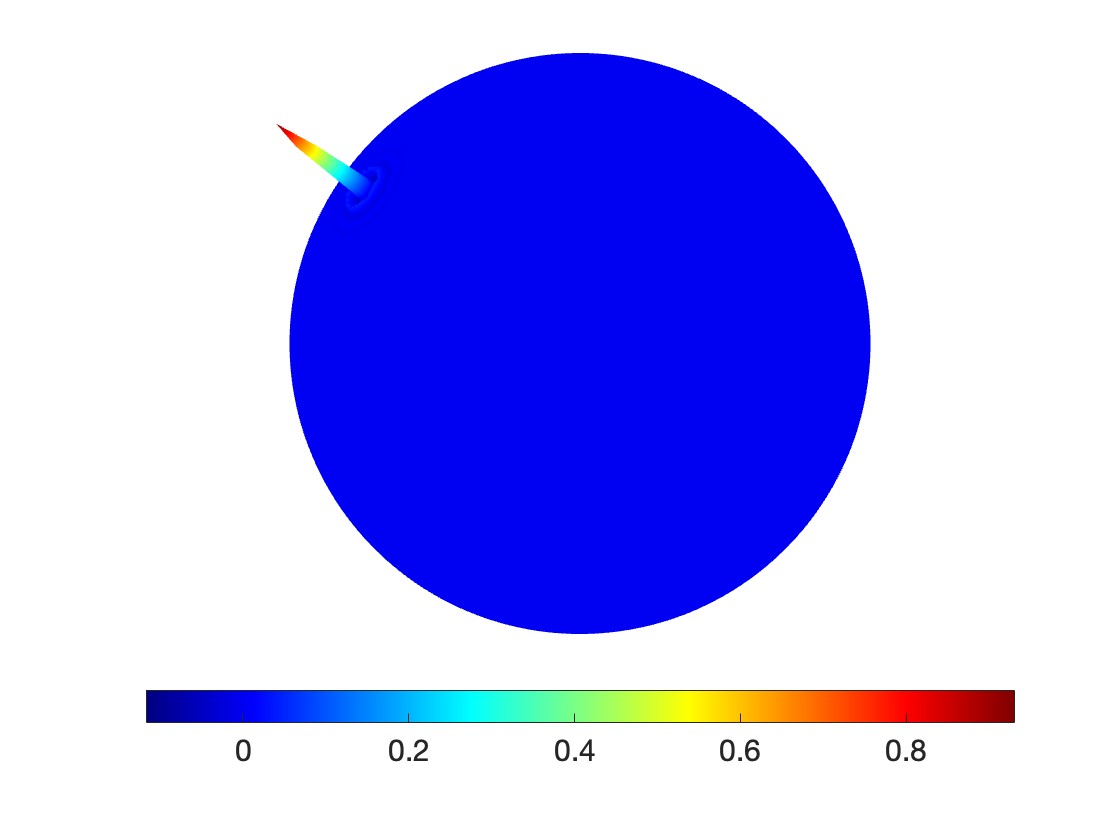} 
\end{minipage}
\end{center}
\caption{This figure illustrates the increasing localization of $\Phi_{n,q}$ as $n$ increases.
Left: $n=32$, Middle: $n=64$, Right: $n=128$.}
\label{fig:sph_loc_kern}
\end{figure}

Figure~\ref{fig:sph_loc_kern} illustrates the localized kernels for $q=2$ for different degrees.

\subsection{Approximation by \texorpdfstring{ReLU$^\gamma$}{ReLU gamma} networks}\label{bhag:relunets}

The ReLU activation function, given by $t_+= \max(t,0)$  is perhaps the most ubiquitously used function in recent times.
For mathematical analysis, it is easier to use the related activation function $|t|=t_++(-t)_+$ ($t_+=(t+|t|)/2$).
The activation function ReLU$^k$ is given by $(t_+)^k$. 
Except in the case when $k$ is an even integer, we may replace this as well (up to a polynomial of degree $k$) by $|t|^k$. 
The case when $k$ is an even integer does not involve new concepts.
More generally, we may define   the ReLU$^\gamma$ function  by $|t|^{2\gamma+1}$. 
A network with this activation function has the form
\be\label{eq:relugammanet}
\x\mapsto \sum_k a_k|\x\cdot \w_k +b_k|^{2\gamma+1}, \qquad \x\in\RR^q.
\ee 
Replacing $\x$ by $\mathbf{X}=(\x,1)/(1+|\x|^2)^{1/2}\in\SS^q$, and writing $\mathbf{W}_k=(w_k,b_k)/(|\w_k|^2+b_k^2)^{1/2}$, one obtains (apart from a weight function) a \emph{zonal function network} given by
\be\label{eq:relugammasph}
\mathbf{X}\mapsto \sum_k A_k|\mathbf{X}\cdot \mathbf{W}_k|^{2\gamma+1}.
\ee
Thus, approximation of a function on $\RR^q$ by ReLU$^\gamma$ networks is equivalent to the approximation of a even function on $\SS^q$ by networks of the form \eqref{eq:relugammasph}.

The series expansion for the kernel $|\x\cdot\y|^{2\gamma+1}$ in terms of spherical harmonics is obtained in \cite{sphrelu}. 
It is very similar to the expansion for eignets, except that only even polynomials are involved and the coefficients have alternating signs.
Nevertheless, using some special properties of ultraspherical polynomials $p_{\ell,q}$, one can carry out the agenda outlined in Section~\ref{bhag:kernelbased} for these kernels as well.

Thus, for any even spherical polynomial $P$, there is a polynomial $\mathcal{D}_\gamma(P)$ of the same degree for which
\be\label{eq:relunativespace}
P(x)=\int_{\SS^q} |\x\cdot \y|^{2\gamma+1}\mathcal{D}_\gamma(P)(\y)d\mu_q^*(\y).
\ee
The operator $\mathcal{D}_\gamma$ can be extended to a subspace of $X^p(\SS^q)$ using standard closure arguments in functional analysis.

We define for $n>0$, the (abstract)  network approximating $f$ (cf. \eref{eq:relunativespace}) by
\be\label{eq:relunetdef}
G_{\gamma;n}(\mu, \nu;f)(\x)
=\int_{\SS^q}|\x\cdot \y|^{2\gamma+1}\mathcal{D}_\gamma\left(\sigma_n(\mu;f)\right)(\y)d\nu(\y).
\ee
We will use a discretely supported measure $\nu$ in this formula, so that $G_{\gamma;n}$ is a finite linear combination of quantities of the form $|\x\cdot \y_k|^{2\gamma+1}$.
We note that $\sigma_n(\mu;f)$ is defined in terms of $\int_{\SS^q}fY_{\ell,k}d\mu$; i.e., using the values of $f$ on the support of $\mu$. 
When $\mu$ is a discrete measure, then $\disp
\mathcal{D}_\gamma\left(\sigma_n(\mu;f)\right)$ is thus well defined in terms of these values.
We will use $\mu$ to be $\mu_q^*$ in the case when $f\in L^p(\SS^q)$, $1\le p<\infty$ and in the case when $f\in C(\XX)$, may take any measure in $MZQ(2n)$ with the property that $f$ is sampled on the support of this measure. 
The following is a special case of our theorems in \cite{sphrelu}.

\begin{theorem}
\label{theo:sphrelutheo}
Let $\gamma>-1/2$, $2\gamma+1$ not be an even integer. Let $1\le p\le \infty$, $f\in X^p(\SS^q)$, $\mathcal{D}_\gamma(f)$ be defined and in $X^p(\SS^q)$,  $n\ge 1$ and $\nu$ be an MZ quadrature measure of order $2^{n+2}$.  
Then
\be\label{eq:phigammadegapprox}
\|f-G_{\gamma;2^n}(\mu_q^*,\nu;f)\|_p \lesssim 2^{-2n(\gamma+1)}(1+\tn\nu\tn)\|\mathcal{D}_{\gamma}(f)\|_p.
\ee
Moreover, the coefficients in the network  satisfy
\be\label{eq:phigammasphcoefftheory}
\int_{\SS^q}|\mathcal{D}_\gamma\left(\sigma_n(\mu_q^*;f)\right)(\y)|d|\nu|(\y)\lesssim \|
\mathcal{D}_{\gamma}(f)\|_1.
\ee
If $p=\infty$, and $\mu$ is an MZ quadrature measure of order $2^{n+2}$ then
\be\label{eq:phigammasampledegapprox}
\|f-G_{\gamma;2^n}( \mu,\nu;f)\|_\infty \le c2^{-2n(\gamma+1)}\tn\mu\tn(1+\tn\nu\tn)\|\mathcal{D}_{\gamma}(f)\|_\infty.
\ee
\end{theorem}
\begin{remark}\label{rem:no_of_terms}
The measure $\nu$ in the theorem above is supported on $\sim 2^{nq}$ points in $\SS^q$.
Thus, in terms of the number of neurons in the network, the rate of approximation in $G_{\gamma;2^n}( \mu,\nu;f)$ as well as $G_{\gamma;2^n}( \mu_q^*,\nu;f)$ in terms of the number $N$ of neurons in the network is $\O(N^{-2(\gamma+1)/q})$.
In the case of ReLU networks, $\gamma=0$. 
Yarotsky \cite{yarotsky2016error} has shown the rate $\O(N^{-2/q})$ to be unimprovable under assumptions on the number of derivatives of $f$ that ensure that $\mathcal{D}_0(f)\in C(\XX)$.  The compositional structure of these networks allows them to overcome the \emph{curse of dimensionality} by efficiently approximating functions that possess a hierarchical or compositional nature  \cite{poggio2016_cbmm58} \qed
\end{remark}
\begin{remark}\label{rem:relu_parameters}
The centers of the network in Theorem~\ref{theo:sphrelutheo} are \textbf{independent of the training data.}
Also the network can be constructed by treating the training data as coefficients of a set of fixed networks constructed independently of the target function; based entirely on the locations of the sampling points.
\qed
\end{remark}
\begin{remark}\label{rem:equivariance}
The networks $G_{\gamma;2^n}(f)$ are rotation equivariant. 
That is if $U$ is any rotation matrix on $\RR^{q+1}$ and $f_U(x)=f(Ux)$, then
\be\label{eq:sphrotinvariance}
G_{\gamma;2^n}(\mu_q^*;\nu;f_U,x)=G_{2^n}(\phi,\mu_q^*;\nu;f,Ux), \qquad x\in\SS^q.
\ee
\qed
\end{remark}
\begin{remark}\label{rem:dimindbd}
In \cite{mhaskar2019dimension}, we have proved that if $f$ is in the variation space for $|t|^{2\gamma+1}$; i.e.,
$$
f(x)=\int_{\SS^q}|t|^{2\gamma+1}d\tau(y)
$$
for some measure $\tau$ having bounded variation on $\SS^q$, then there exists a network $\mathbb{G}(f)$ with $N$ neurons which satisfies the bound
\be\label{eq:dimindbd}
\|f-\mathbb{G}\|_\infty \lesssim \|\tau\|_{TV}\frac{\sqrt{\log N}}{N^{1/2+(2\gamma+1)/q}}.
\ee
In  particular, if $\mathcal{D}_\gamma(f)\in C(\XX)$, then $f$ is in the variation space and $\|\tau\|_{TV}\le \|\mathcal{D}_\gamma(f)\|_\infty$.
Thus, we obtain both a dimension-dependent bound and a dimension-independent bound for the same function space.
For integer values of $2\gamma+1$, Siegel \cite{siegel2025optimal} has shown that the bound is optimal.
Thus, both kinds of bounds are unimprovable in some sense.
 However, the dimension-dependent bounds are derived constructively, while the dimension-independent bounds are presented as an existence theorem, without providing guidance on how the network parameters should be selected. 
In particular, the dimension-independent bounds may not be realizable if the networks are to be constructed based on the values of the target function. \qed
\end{remark}

\subsection{Numerical Comparison of Spherical Approximation Methods}
\label{subsec:numerical_experiments_sphere}

To illustrate the practical performance of the localized approximation schemes discussed in Sections 7 and 8, we present a numerical  example to compare the use of localized kernels vs. kernels with shart cutoff vs. least squares and the use of Monte Carlo vs. quadrature formulas for approximating functions defined on the unit sphere $\mathbb{S}^2 \subset \mathbb{R}^3$.

In \cite{quadconst}, we had studied five benchmark examples for this purpose. 
We review one of these:
$$
G(x)=\max\!\left(0.015 - (x_1^2 + x_2^2 + (x_3 - 1.02)^2), 0\right)
       + \exp(0.9x_1 + 1.1x_2 + x_3), \qquad (x_1,x_2, x_3)\in\SS^2.
$$
We will approximate this by polynomials in $\Pi_{64}^2$, based on $65,536$ samples of $G$, and test the quality of approximation on $20,000$ points, all the samples and test points being generated randomly from the uniform distribtution on $\SS^2$.

In \cite{quadconst},the spherical harmonic coefficients of $G$ were approximated using a quadrature formula, and a specially designed spline function of order $5$ was used as the filter $h$. 
The resulting approximation will be denoted by $QS5$. 
Instead, one could use the sharp cutoff filter $h$ so that we use the reproducing kernel rather than a localized kernel. 
This method will be denoted by $QS1$.
Finally, we could use a straightforward Monte Carlo method to approximate the coefficients.
The resulting approximations will be denoted by $MS5$ and $MS1$ respectively.
Finally, we denote the usual least square approximation by $LS$.
In Table~\ref{tab:sphere_error}, we summarize the results as percentage of points where the error $<10^{-x}$.
We note that all these methods give comparable results when the desired error is $<10^{-4}$. 
The difference is stark when the desired error is $<10^{-7}$. 
With ordinary least square approximation LS, the error is $<10^{-7}$ only at $0.92\%$ of the test points, and the results with $MS1$ and $QS1$ are similar.
With localized kernel, the percentage of points for even the Monte Carlo method $MS5$ is much greater at $3.19\%$, but the method $QS5$ using quadrature formulas and localized kernel is a clear winner with the error $<10^{-7}$ at $90.78\%$.

\begin{table}[htbp]
\centering
\renewcommand{\arraystretch}{1.15}
\begin{tabular}{|c|ccccccccc|}
\hline
$x \rightarrow$ & 10 & 9 & 8 & 7 & 6 & 5 & 4 & 3 & 2 \\
\hline
LS & 0.00 & 0.02 & 0.08 & 0.92 & 9.46 & 86.10 & 99.11 & 99.99 & 100.00 \\
\hline
MS1 & 0.00 & 0.02 & 0.08 & 0.92 & 9.46 & 86.10 & 99.11 & 99.99 & 100.00 \\
QS1 & 0.00 & 0.01 & 0.09 & 0.74 & 7.94 & 84.99 & 99.19 & 99.99 & 100.00 \\
\hline
MS5 & 0.02 & 0.06 & 0.32 & 3.19 & 33.16 & 95.66 & 99.06 & 99.98 & 100.00\\
QS5 & 0.39 & 3.34 & 41.95 & 90.78 & 94.52 & 97.19 & 99.18 & 99.97 & 100.00\\
\hline

\end{tabular}
\caption{Percentage of points at which the error is $<10^{-x}$ by different methods of approximation}
\label{tab:sphere_error}
\end{table}

\section{Operator approximation}\label{bhag:operatorapprox}
The problem of approximating an operator acting on a Banach space arises in a wide range of applications across various fields. 
For example, in the theory of partial differential equations, it is crucial for approximating solutions to complex equations, such as those governing fluid dynamics, heat transfer, and electromagnetic fields. 
In inverse problems, such as identifying material properties from experimental data or reconstructing unknown boundary conditions, operator approximation plays a key role. 
Similarly, in the modeling of time series data, operators are often used to map input sequences to output classes, and efficient approximation methods can improve accuracy of tasks such as prediction of time series. 
Additionally, in system identification, where the goal is to model a dynamical system based on observed inputs and outputs, approximating operators is essential for developing accurate system models. 
These examples illustrate the broad relevance of operator approximation in diverse domains, including machine learning, signal processing, control theory, and scientific computing.
Early works in this direction are the oft-cited papers by Chen and Chen \cite{chen1993approximations,chen1995universal}, where the authors prove that continuous (nonlinear) operators from $C(\XX)$ to $C(\YY)$ for compact sets $\XX$, $\YY$ can be approximated in the weak topology arbitrarily well by neural networks with a special structure, known now-a-days as branch and trunk networks. 
This is a pure existence result, without error bounds or explicit constructions. In \cite{mhaskar1997neural}, we presented a constructive procedure to achieve this in the case of functionals acting on $L^p([-1,1])$, and derived optimal bounds on the approximation error. The field has seen renewed interest in recent years with the development of various network architectures, such as Deep neural operators \cite{lu2021learning45}, Fourier Neural Operators (FNOs) \cite{li2020fourier}, Spectral Neural Operators (SNO) \cite{fanaskov2023spectral}, Two-step DeepOKAN \cite{zhang2026bubbleokan}, Wavelet Neural Operators (WNO) \cite{tripura2023wavelet}, RiemannONet \cite{peyvan2024riemannonets}, Convolutional Neural Operator \cite{raonic2023convolutional}, Partition-of-Unity DeepONet \cite{goswami2024learning}, U-nets, and others. However, from the perspective of approximation theory, results in this area are still underdeveloped.
A few references include \cite{mhaskar2023local,kovachki2021universal,lanthaler2022error}. Other theoretical works include providing estimates on generalization error for operator learning, such as the bounds provided in \cite{chen2023deep,liu2024deep}. They study the influence of various network parameters on the error and demonstrate that if the target operator exhibits a low complexity structure, then deep operator learning can avoid the curse of dimensionality.

In Section~\ref{bhag:pins}, we review some state-of-the-art results dealing with the use of deep networks for operator approximation.
In Section~\ref{bhag:sphere_reduce}, we review the probably non-traditional results in our paper \cite{mhaskar2022local}, where we use approximation by spherical polynomials for this purpose.

\subsection{Physics-Informed Neural Surrogates}\label{bhag:pins}
Physics-Informed Neural Surrogates (PINS) are deep learning models that incorporate physical laws, typically represented by partial differential equations (PDEs), into neural networks to improve predictive accuracy and generalization. By embedding governing equations as soft constraints or penalty terms in the loss function, these surrogates minimize reliance on large labeled datasets while ensuring physically consistent solutions. They are particularly effective for high-dimensional and computationally intensive simulations, including turbulence modeling and fluid dynamics. In the literature, PINS have been referred to by various names, such as physics-informed neural networks, physics-guided neural networks, physics-constrained neural networks, and physics-enhanced neural networks. Among all, most popular one is Physics-Informed Neural Networks (PINNs) \cite{raissi2019physics,penwarden2023unified,jagtap2022deepWW,abbasi2025history,kharazmi2021hp, jagtap2020conservative,jagtap2020adaptive,jagtap2022deepknn,menon2025anant, jagtap2020locally, abbasi2025challenges,shukla2021physics, jagtap2023important,jagtap2022physics,mao2020physics,shukla2021parallel}. Despite the widespread adoption of PINNs across various scientific applications, there is limited mathematical analysis explaining their effectiveness. This section reviews foundational publications that contribute to the theoretical understanding of PINNs, particularly their convergence properties.

PINNs have demonstrated considerable potential for solving PDEs, yet several common failure modes limit their robustness in practice. Prominent among these is loss imbalance, where physics, boundary, and data terms evolve on disparate scales, leading to slow convergence or biased solutions unless carefully weighted or adaptively rescaled \cite{wang2021loss}. Boundary and initial condition enforcement can be particularly challenging for convection-dominated or hyperbolic systems, where weak enforcement may permit unphysical solutions \cite{krishnapriyan2021failure}. Moreover, single-network formulations often struggle with sharp gradients and multiscale features, motivating domain decomposition and multi-network approaches \cite{jagtap2020conservative,jagtap2020extended}.

Error analysis for PINNs has been an active area of research, with recent studies focusing on deriving rigorous error bounds and providing comprehensive reviews of existing results; see Ryck and Mishra \cite{de2022error,de2024numerical}. These analyses typically consider various error components and their dependence on factors such as the type of partial differential equation and the dimensionality of the problem domain (Ryck and Mishra \cite{de2024numerical}).
The development of error analysis for PINNs includes frameworks for both a priori and a posteriori error estimates, particularly for linear PDEs by Zeinhofer et al. \cite{zeinhofer2024unified}. They also proposed the unified approach based on bilinear forms, where coercivity and continuity conditions yield sharp error estimates. Additionally, numerical analyses of PINNs have been conducted to establish a structured framework for assessing errors in PDE approximations. These studies examine approximation, generalization, and training errors while highlighting the influence of solution regularity and stability \cite{de2024numerical}.
Key advancements in PINN error estimates include rigorous error bounds for approximating solutions of linear parabolic PDEs, particularly Kolmogorov equations; see Ryck and Mishra \cite{de2022error}. Comprehensive analyses have provided deeper insights into the accuracy of PINNs in solving these equations. Researchers have also derived a priori and a posteriori error estimates for PINNs applied to a range of PDEs, including elliptic, parabolic, hyperbolic, and Stokes equations, as well as PDE-constrained optimization problems.
The  first comprehensive theoretical analysis of PINNs (and eXtended PINNs or XPINNs \cite{jagtap2020extended}) for a prototypical nonlinear PDE, the Navier-Stokes equations is done by Ryck et al. \cite{de2024errorEE}. 
These estimates relate total error to training error, network architecture, and the number of quadrature points used in the approximation. Hu et al. \cite{hu2021extended} developed generalization results for both PINNs and XPINNs.
Several approaches to error estimation in PINNs have been explored, including a posteriori error estimation, which is crucial for certifying PINN solutions; see Hillebrecht and Unger \cite{hillebrecht2022certified}. 

In particular, in their earlier work, Shin et al. \cite{shin2020convergence} established the consistency of PINNs for linear elliptic and parabolic PDEs, demonstrating that a sequence of neural networks minimizing Hölder-regularized losses converges uniformly to the PDE solution with probability one. Their follow-up study \cite{shin2020error} extended these results to linear advection equations. Ryck et al. \cite{de2024errorEE} provided the first comprehensive nonlinear theory for PINNs.
The PINN framework reformulates PDEs as optimization problems, raising critical theoretical questions:
\begin{enumerate}
    \item \textbf{Existence of Approximate Solutions}: Given a tolerance \(\epsilon > 0\), do neural networks exist such that both generalization and training errors remain below \(\epsilon\)? An affirmative answer ensures the minimization of the PDE residual but does not necessarily imply a small total error.
\item \textbf{Total Error Bound}: If the generalization error is small, is the total error also small, i.e., \(||u - \hat{u}_{\theta}|| < \delta(\epsilon)\), where \(\delta(\epsilon) \sim O(\epsilon)\) in a suitable norm? This question confirms whether PINN approximations remain faithful to the true PDE solution.
\item \textbf{Generalization from Training}: Given a small training error and sufficient training data, does the generalization error remain proportionately small? This ensures that trained PINNs accurately approximate the PDE solution.
\end{enumerate}

Mishra and Molinaro \cite{mishra2022estimates} leveraged PDE stability to bound total error relative to generalization error (Question 2) and related generalization to training error via quadrature accuracy (Question 3). However, Question 1 was not addressed, and their assumptions on network weights may not hold universally. Ryck and Mishra \cite{de2022error} analyzed Kolmogorov PDEs, including the heat equation and Black-Scholes equation, showing that PINNs satisfy the error bound:
\[\mathscr{R}(\hat{u}_{\theta}) \leq \left(C \hat{\mathscr{R}}(u_{\theta}) + \mathcal{O}(N^{-1/2}) \right)^{1/2}.\]
For nonlinear PDEs, Ryck et al. \cite{de2024errorEE} examined PINNs for the Navier-Stokes equations, affirmatively answering all three theoretical questions. They demonstrated the existence of neural networks approximating Navier-Stokes solutions with minimal generalization and training errors. Furthermore, they related generalization error to total error and utilized quadrature rules to link generalization to training error:
\[\mathscr{R}(\hat{u}_{\theta}) \leq \left(C \hat{\mathscr{R}}(u_{\theta}) + \mathcal{O}(N^{-1/d}) \right)^{1/2},\]
where \(d\) is the problem dimension.
Mishra and Molinaro \cite{mishra2021physics} studied high-dimensional radiative transfer equations, establishing generalization error bounds in terms of training error and sample size:
\[\mathscr{R}(\hat{u}_{\theta}) \leq \left(C \hat{\mathscr{R}}(u_{\theta})^2 + c\left(\frac{(\ln N)^{2d}}{N} \right) \right)^{1/2}.\]
This result suggests that PINNs may not suffer from the curse of dimensionality, provided that training errors remain independent of dimensionality. A good survey paper on theoretical foundations of PINNs and neural operator is given by Shin et al. \cite{shin2024theoretical}. Overall, these studies contribute to a deeper understanding of PINNs, validating their empirical success through rigorous mathematical analysis.

Scientific foundation models (SciFM) represent a new class of data-driven systems designed to capture shared structure across diverse scientific domains by learning from large, heterogeneous collections of experimental, observational, and simulated data. A rigorous definition of SciFMs, together with a comprehensive review of their scope, methodologies, and applications, is provided by Menon \emph{et al.}~\cite{menon2026scientific}.
Unlike task-specific models, SciFM are trained to produce generalisable representations of physical, chemical, biological, and environmental processes, enabling transfer across scales, modalities, and disciplines. By integrating domain knowledge with self-supervised learning and scalable architectures, these models offer a unifying computational substrate for prediction, discovery, and hypothesis generation. As SciFM increasingly inform scientific inference and decision-making, understanding their design principles, limitations, and alignment with physical laws is essential for ensuring both scientific validity and reproducibility. SciFMs extend beyond single PINN model by learning generalisable representations from diverse scientific systems and regimes, rather than optimising for a single problem setting. This confers improved data efficiency, robustness across conditions, and the ability to adapt rapidly to new tasks without problem-specific retraining.

\subsection{Reduction to the sphere}\label{bhag:sphere_reduce}
Intuitively, the problem of approximation of operators is not difficult. 
To illustrate the ideas, let $\mathcal{F} : L^2(\TT^q)\to L^2(\TT)$, and
\be\label{eq:oplipschitz}
\|\mathcal{F}(f)-\mathcal{F}(g)\|\le \|f-g\|,
\ee
where, in this discussion, $\|\cdot\|$, norm denotes the $L^2(\TT^q)$ (or $L^2(\TT)$)  norm.
Any $f\in L^2(\TT^q)$ can be approximated optimally by the partial sum of its Fourier series, effectively encoding $f$ via its Fourier coefficients of order up to $n$. 
Denoting the number of Fourier coefficients by $d\sim n^q$,  we have the encoder-decoder pair:
\be\label{eq:encoder_decoder}
\mathcal{E}_{q,d}:L^2(\TT^q)\to \RR^d, \quad \mathcal{E}_{q,_d}(f)=\{\hat{f}(\k)\}_{|\k|_1<n}, \quad \mathcal{D}_{q,d} :\RR^d\to L^2(\TT^q), \quad \mathcal{D}_{q,d}(\mathbf{a})=\sum_{\k: |\k|_1<n}a_\k\exp(i\k\cdot \x).
\ee
Assuming that $f\in W_{2;\gamma}$ for some $\gamma>0$, $\|f\|_{2;\gamma}\le 1$, we have
\be\label{eq:periodrecuperation}
\|f-\mathcal{D}_{q,d}(\mathcal{E}_{q,d}(f))\|\lesssim d^{-\gamma/q}.
\ee
Accordingly,
\be\label{eq:operatorrecup1}
\left\|\mathcal{F}(f)-\left(\mathcal{F}\circ\mathcal{D}_{q,d}\right)(\mathcal{E}_{q,d}(f))\right\|\lesssim d^{-\gamma/q}.
\ee
\yadi{$\mathcal{E}_{q,d}$}{Encoder, \eqref{eq:encoder_decoder}}
\yadi{$\mathcal{D}_{q,d}$}{Decoder, \eqref{eq:encoder_decoder}}
In this discussion, let us denote $\disp \left(\mathcal{F}\circ\mathcal{D}_{q,d}\right)(\mathcal{E}_{q,d}(f))$ by $g\in L^2(\TT)$.
We may use the encoder-decoder pair based on the Fourier coefficients of $g \in L^2(\TT)$, say $\mathcal{E}_{1,m}$, $\mathcal{D}_{1,m}$. 
If $g\in W_{2;\Gamma}$ for some $\Gamma>0$ and $\|g\|_{2;\Gamma} \le 1$, then this leads to
\be\label{eq:operatorrecup2}
\left\|\mathcal{F}(f)-\mathcal{D}_{1,m}(\mathcal{E}_{1,m}(g))\right\|\lesssim d^{-\gamma/q}+m^{-\Gamma}.
\ee
Thus, the task reduces to approximating each of the Fourier coefficients $\hat{g}(\ell)$, $|\ell|<m$ as a function of $\mathcal{E}_{q,d}(f)\in \RR^d$; i.e., $2m+1$ functions on $\RR^d$, where $d\sim n^q$.
Figure~\ref{fig:operatorreduce} is a graphical illustration of this reduction of the problem of approximation of an operator to that of the problem of many real valued functions of several real variables.

\begin{figure}[ht]
\[
\begin{tikzcd}
	f
	\arrow[r, "\mathcal{E}_{q,d}"]
	\arrow[d, "\mathcal{F}"',  end anchor={[xshift=-10pt]}]
	&
	\mathcal{E}_{q,d}(f) \in \mathbb{R}^d
	\arrow[d]
	\\
	\mathcal{F}(f) \in L^2(\mathbb{T})
	&
	g = (\mathcal{F} \circ \mathcal{D}_{q,d})(\mathcal{E}_{q,d}(f)) 
	\arrow[l, "\mathcal{D}_{1,m} \circ \mathcal{E}_{1,m}"'] 
	\arrow[d, "\mathcal{E}_{1,m}"]
	\\
	&
	\{\hat{g}(\ell)\}_{|\ell| < m} \in \mathbb{R}^{2m+1} 
	\arrow[d, "\mathcal{D}_{1,m}"]
	\\
	&
	\mathcal{D}_{1,m}(\mathcal{E}_{1,m}(g))
	\arrow[uuuu, to=2-1, "\approx", end anchor={[xshift=-20pt]}]
\end{tikzcd}
\]
\caption{The reduction of operator approximation to approximation of many real valued functions of finite variables.}
\label{fig:operatorreduce}
\end{figure}

In principle, one can use any of the approximation methods discussed above for this purpose.
Thus, a universal approximation theorem is quite trivial.

If we wish to keep track of the approximation error, then there is a formidable challenge.
In order to get a good approximation to $f$, we need a large $d$, and likewise, we need a large $m$. 
On the other hand, the main task of approximating the $2m+1$ functions on $\RR^d$ using $N$ parameters will lead to approximation errors of the form $C(d)/N^{r/d}$, where $C(d)$ is typically a constant depending upon $d$ in an unspecified manner.
One solution to this problem is to assume that each of the $2m+1$ functions belong to a class of functions on $\RR^d$ (or at least $[-1,1]^d$) which does not have a curse of dimensionality, or at least a very weak curse; e.g., that these are holomorphic functions. 
This is the approach taken by Addock et. al \cite{adcock2020deep,adcock2021learning}. 

The following discussion is based on \cite{mhaskar2023local}.
In \cite{mhaskar2023local}, we took up a more ambitious program, where we assume that these functions belong to a class which does have a curse of dimensionality, but develop approximation methods to ensure that the constant $C(d)$ depends only sub-polynomially on $d$ ($\O(d^{1/6})$ to be precise). 
For this purpose, we need to define smoothness of functions carefully, guided by the classical definition by Stein \cite{stein2016singular}.
Also, we use specially designed localized kernels so that the training data, consisting of functions  in $L^2(\TT^q)$, can be chosen only on a small neighborhood of the input function $f$.

In the general setting of approximation of an operator from one Banach space to another, we assume that the ``right'' encoder-decoder pairs are selected depending upon the domain knowledge about the application in which this problem arises. 
Further, the values of the $\RR^d$-valued encoder can be mapped to $\SS^d$, using a suitable stereographic projection similar to what is described in Section~\ref{bhag:relunets}.
Hence, we focus on the approximation of a continuous function on $\SS^d$, in particular, the dependence of the various constants on $d$. 
Rather than complicating the notation, we will denote in the rest of this section the target function on $\SS^d$ by $f$, not to be confused with the $f$ in the discussion so far.

We need three main ingredients: (1) the smoothness class, (2) a more refined version of MZ quadrature measures, and (3) specially designed kernels and the corresponding operators.

First, we discuss the smoothness of $f$, adapted from Stein's definition.
We recall our convention that for \emph{any} $r>0$ (integer or non-integer), $\Pi_r^d$ is the class of all spherical polynomials of degree $<r$.
\begin{definition}\label{def:smoothness}
Let $f\in C(\SS^d)$, $r>0$ and $\x\in\SS^d$. The function $f$ is said to be $r$-smooth at $\x$ if there exists $\delta=\delta(d;f,\x)>0$  such that
\be\label{eq:smoothnessdef}
\|f\|_{d;r,\x}:=\|f\|_\infty+\min_{P\in \Pi_r^d}\max_{\y\in \mathbb{B}(\x,\delta)}\frac{|f(\y)-P(\y)|}{|\x-\y|_{d+1}^r} <\infty.
\ee
The class of all $f\in C(\SS^d)$ for which $\|f\|_{d;r,\x}<\infty$ will be denoted by $W_{d;r,\x}$.
The class $W_{d;r}$ will denote the set of all $f\in C(\SS^d)$ for which 
\be\label{eq:globalsmoothessdef}
\|f\|_{d;r}=\sup_{\x\in \SS^d}\|f\|_{d;r,\x} <\infty.
\ee
We note that for $f\in W_{d;r}$, we may choose $\delta(d;f)$ in \eqref{eq:smoothnessdef} to be independent of $\x$.
\end{definition}

The notion of the MZ quadrature measures needs to be refined in order to keep track of the constants. 
We propose the following modification, adopted for the current application.
\begin{definition}\label{def:quadmeasure}
Let $n\ge 1$. A  measure $\nu$ on $\SS^d$ is called a \textbf{quadrature measure of order $n$} if for every $P\in\Pi_n^d$,
\be\label{eq:op_quadrature}
\int_{\SS^d} Pd\nu =\int_{\SS^d}Pd\mu^*. 
\ee
A measure $\nu$ is called a \textbf{Marcinkiewicz-Zygmund  measure of order $n$} (abbreviated by $\nu\in \mathsf{MZ}(d;n)$) if  for every $P\in\Pi_{n/2}^d$:
\be\label{eq:op_mzineq}
 \int_{\SS^d} |P|^2d|\nu| \le c\int_{\SS^d}|P|^2d\mu^*,
\ee
for some positive constant $c$. The infimum of all such constants will be denoted by $\tn\nu\tn_{d;n}$.
A measure $\nu$ is called a \textbf{Marcinkiewicz-Zygmund quadrature measure of order $n$} (abbreviated by $\nu\in \mathsf{MZQ}(d;n)$) if both \eqref{eq:quadrature} and \eqref{eq:mzineq} hold for every $P\in \Pi_n^d$.
\end{definition}

The Tchakaloff theorem \cite{rivlin2020chebyshev} implies that MZ quadrature measures of oreder $n$ exist, supported on exactly $\mathsf{dim}(\Pi_n^d)$ points.
At this time, we do not know an explicit construction of such formulas.
Theorem~\ref{theo:mzqtheo} states the existence of formulas based on arbitrary data, but the constants involved depend upon $d$ in an unspecified manner.

Finally, we define the localized kernels, motivated by \cite{filbir2008polynomial}.

Let $K_{d,n}$ denote the reproducing kernel for $\Pi_n^d$; i.e.,
$$
K_{d,n}(x\cdot y)=\sum_{\ell=0}^{n-1}\sum_{k=1}^{\dim (\mathbb{H}^q_\ell)} Y_{\ell,k}(x)Y_{\ell,k}(y).
$$
The kernel $K_{d,n}$ can be expressed explicity in terms of Jacobi polynomials.
A Jacobi polynomial $p_k^{(\alpha,\beta)}$ is a polynomial of degree $k$ with positive leading coefficient such that the sequence $\{p_k^{(\alpha,\beta)}\}$ satisfies the orthogonality relation
$$
\int_{-1}^1 p_k^{(\alpha,\beta)}(t)p_\ell^{(\alpha,\beta)}(t)(1-t)^\alpha (1+t)^\beta =\delta_{k,\ell}.
$$
Our specially designed kernel is defined by
\be\label{eq:op_kerndef}
\Phi_{d;n,r}(x)=\widetilde{\Phi}_{d;n,r}(x)\left(\frac{1+x}{2}\right)^{n}= K_{d;(d+2)n}(x)\frac{p_{dn}^{(d/2+r,d/2-2)}(x)}{p_{dn}^{(d/2+r,d/2-2)}(1)}\left(\frac{1+x}{2}\right)^n.
\ee
If $\nu$ is a measure on $\SS^d$ having bounded total variation and $f$ is integrable with respect to $\nu$, we define the corresponding operator by
\be\label{eq:sphoperatordef}
\sigma_{d;n,r}(\nu, f)(\x)=\int_{\SS^d}f(\y)\Phi_{d;n,r}(\x\cdot\y)d\nu(\y), \qquad n >0, \ \x\in\SS^d.
\ee

In the remainder of this section, the constants involved in $\lesssim$, $\gtrsim$, and $\sim$ are understood be independent of the dimension $d$, although they may depend upon other fixed parameters (such as the smoothness $r$) as usual.

An important property of these operators is given in the following theorem, which is similar to the previous theorems of this kind, but the dependence on the constants is described explicitly.
In interpreting these theorems in terms of operator approximation, one has to keep in mind that each point $\x\in\SS^d$ corresponds to the encoding of some function in our domain $\mathfrak{X}$ of the operator.

\begin{theorem}\label{theo:op_goodsphapprox}
Let $d\ge 3$, $r\ge 0$, $n\ge 2(d+1)$,  and $f\in C(\SS^d)$. If $\nu\in\mathsf{MZQ}(d;2(d+2)n)$, then
\be\label{eq:op_degapproxstrongbis}
E_{d;2(d+2)n}(f)\le \|f-\sigma_{d;n,r}(\nu,f)\|_\infty \ls d^{1/6}\tn\nu\tn_{d;2(d+2)n}E_{d;n}(f).
\ee 
\end{theorem}

We end this section with a theorem regarding local approximation.
\begin{theorem}\label{theo:op_locsphapprox}
Let $d\ge 3$,   $\x\in\SS^d$, $r=r(\x)>0$,and $f\in W_{d;r,\x}$. Let $nd\ge (d+r+1)^2$,  $\nu\in\mathsf{MZQ}(d;2(d+2)n)$.
If $n$ is large enough so that
 \be\label{eq:near_ball_def}
\delta_n=\sqrt{\frac{16r\log n}{n}}\le \delta(d;f,\x),
\ee
then
\be\label{eq:loc_deg_wholeint}
|f(\x)-\sigma_{d;n,r}(\nu,f)(\x)|\ls \frac{d^{1/6}}{\mathsf{dim}(\Pi_{2(2+d)n}^d)^{r/d}}\|f\|_{W_{d;r,\x}}\tn\nu\tn_{d;2(d+2)n}.
\ee
Moreover,
 \be\label{eq:loc_deg_approx}
 \left|f(\x)-\int_{\BB(\x,\delta_n)}\Phi_{d;n,r}(\x\cdot\y)f(\y)d\nu(\y)\right|\ls \frac{d^{1/6}}{\mathsf{dim}(\Pi_{2(d+2)n}^d)^{r/d}}\|f\|_{W_{d;r,\x}}\tn\nu\tn_{d;2(d+2)n}.
 \ee
 All the constants involved in $\ls$ depend only upon $r(\x)$ but are otherwise independent of $f$, $\x$, $n$, and $d$.
\end{theorem}

\bhag{A new paradigm}\label{bhag:newparadigm}
\subsection{Some comments on the current paradigm}\label{bhag:shortcomings}

The prevailing paradigm in machine learning, selecting the model class based on approximation estimates, optimizing the model using data-driven processes, and subsequently analyzing its generalization error, has demonstrated remarkable success across various applications. However, from a theoretical perspective, this approach exhibits several shortcomings that warrant closer examination.

\begin{enumerate}
\item The use of degree of approximation in order to guide the choice of the model complexity might be misleading. 
As observed in Section~\ref{bhag:relunets}, ReLU networks provide drastically different estimates on the approximation error for the approximation of  \emph{the same class of functions}, depending upon whether the proof is constructive or merely existential.
Most approximation error estimates in the literature primarily establish the existence of a model that achieves a desirable level of approximation error, often highlighting dimension-independent bounds, which are particularly appealing. 
However, such results lack practical utility unless they are constructive. 
If constructive proofs are available, the reliance on empirical risk minimization becomes unnecessary. 
This is not to suggest that optimization should be  avoided entirely, but rather that alternative objectives may provide greater value than directly optimizing the expected value of a loss functional.
We have already demonstrated that in approximation on manifolds, finding the right quadrature formulas is a more fruitful quest.
\item Apart from the usual problems of optimization theory such as speed or even lack of convergence, local minima, etc., it is not clear that the result of the empirical minimization process will approximate the actual target function. 
It is shown in \cite{di2023neural,jin2020quantifying} that if the weights of a deep network are not initialized correctly, then the network may be ``dead on arrival''; i.e., produce a constant output regardless of the input data.
Another well known problem is the so called \emph{spectral bias}; the optimization process learns low frequency parts first and has difficulties transiting to the high frequency parts \cite{xu2024overview,rahaman2019spectral,fridovich2022spectral,fang2024addressing,cao2019towards}.
The use of appropriate quadrature formulas can mitigate this problem.
\item 
The generalization error, defined as an expected value, measures the error globally as an average. Intuitively, it is more desirable to estimate the probability that the difference between the actual target function and the constructed model becomes large, such as Theorems~\ref{theo:directsoltrig} and \ref{theo:probapprox}. Figure~\ref{fig:cospowquarter} illustrates how a global norm can be misleading as a measurement of the error.

\begin{figure}[ht]
    \centering
   \begin{minipage}{0.4\textwidth}
   \includegraphics[width=\textwidth]{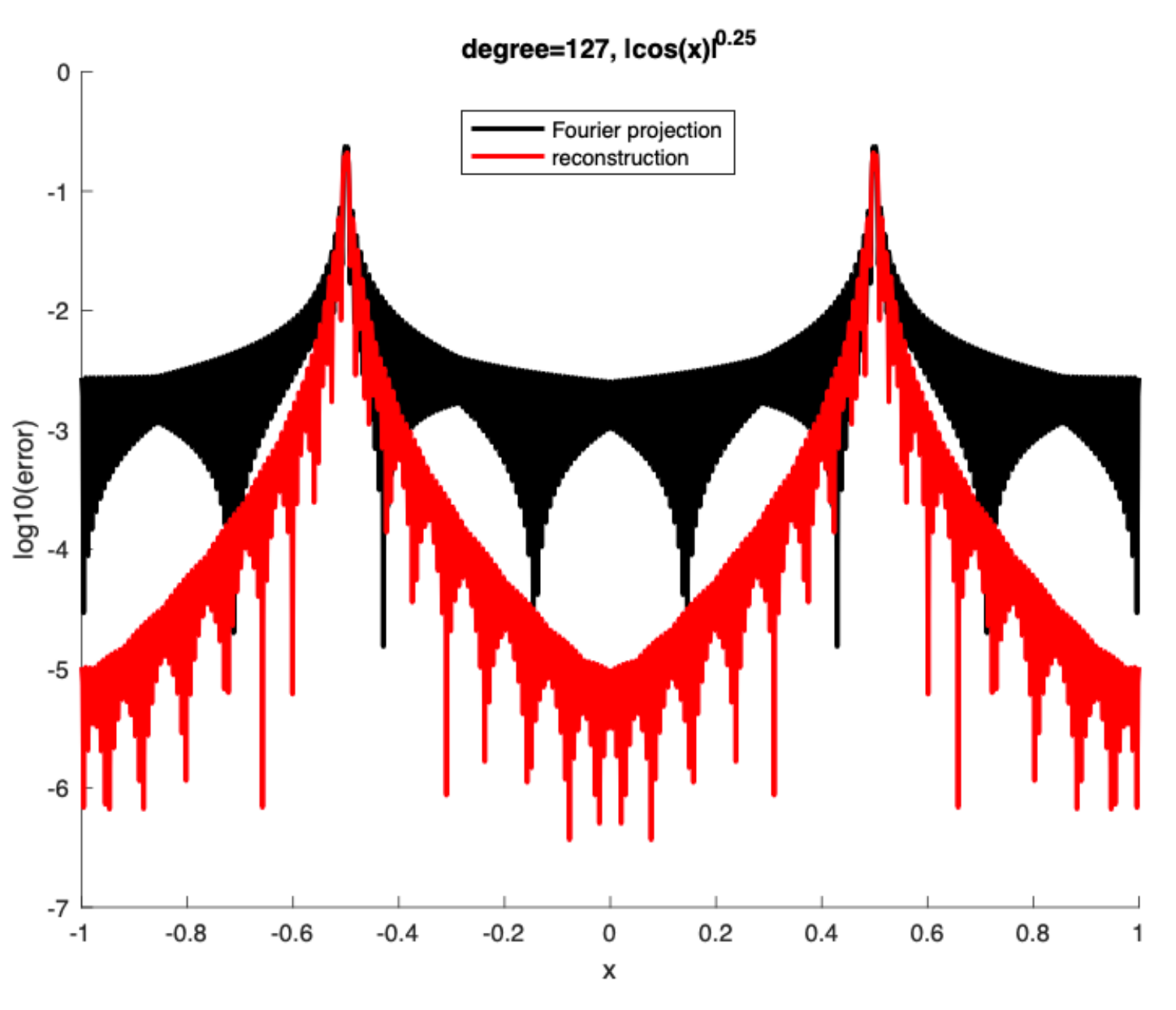} 
   \end{minipage}
    \begin{minipage}{0.4\textwidth}
   \includegraphics[width=\textwidth]{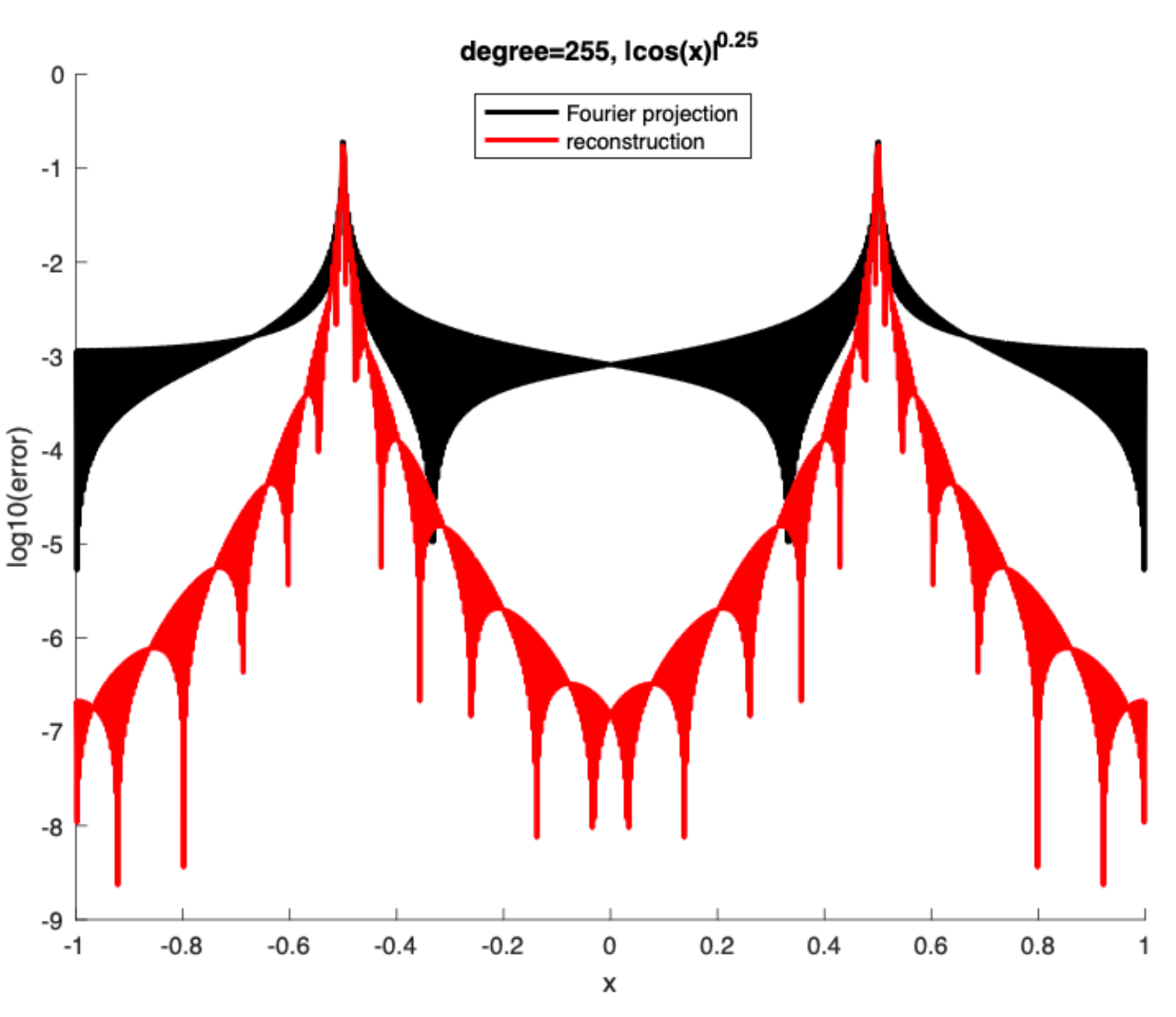} 
   \end{minipage}
   \caption{Illustration of local approximation of $f(x)=|\cos x|^{1/4}$. The interval $[-\pi,\pi]$ on the  $x$-axis, labeled with $x/\pi$. The log-plot of the absolute errors at each point on the $y$-axis. Left: The error plot for Fourier projection of degree $128$ in black, $\sigma_{128}(f)$ in red. With Fourier projection, the error is $\le 10^{-4}$ at only $1.56\%$ points, the corresponding percentage with the localized operator is $56.25\%$. Right: The error plot for Fourier projection of degree $256$ in black, $\sigma_{256}(f)$ in red.
    With Fourier projection, the error is $\le 10^{-4}$ at only $3.52\%$ points, the corresponding percentage with the localized operator is $79.30\%$.}
   \label{fig:cospowquarter}
\end{figure}
\end{enumerate}

\subsection{Limitations of classical approximation theory}\label{bhag:limitations}

Given that the core problem in machine learning is often framed as one of function approximation, it is reasonable to question why approximation theory is not employed directly to address this challenge. Instead, it is frequently used in a tangential and somewhat ad hoc manner, primarily to gain insights into the complexity of the required model.
There are several reasons.
\begin{enumerate}
\item  In classical approximation theory, the target function is typically sampled at carefully chosen points within the domain, for example, grid points to enable the use of the Fast Fourier Transform (FFT) for periodic functions. 
In contrast, a typical machine learning problem involves sampling the target function at data points obtained from natural processes, where there is no control over the sampling locations. 
Such input data is commonly referred to as \emph{scattered data}.
Our solution to solve this problem is given by quadrature formulas discussed above
\cite{frankbern,modlpmz,mnw1,mhaskar2020kernel}.
\item  In machine learning, the target function's values are often noisy. Consequently, beyond the challenge of approximating the function itself, it is essential to account for the noise in these values. Addressing this issue involves selecting appropriate localized kernels tailored to the core approximation problem and employing concentration inequalities to effectively manage the noise.
\item A significant limitation of directly applying classical approximation theory is its reliance on the assumption that data points are densely distributed over a known domain (e.g., a cube or a sphere). The manifold hypothesis addresses this limitation by proposing that the data resides on an \emph{unknown} low-dimensional manifold, thus providing a framework to work with non-dense and high-dimensional data distributions.

In Section~\ref{bhag:manifoldlearning}, we have described a few approaches, all very delicate, to learn enough about the manifold. 
Regardless of the sophistication of such techniques, the fundamental challenge remains that any data-driven approach is inherently ill-posed. Errors in approximating quantities like eigen-decompositions will inevitably propagate, contaminating the results of approximations based on these quantities.
Another significant issue with the manifold hypothesis is that, to learn the manifold, such as constructing a graph Laplacian, the entire dataset must be available in advance. This requirement imposes a constraint on the applicability of these methods in many practical scenarios.

Finally, the manifold hypothesis may not hold in its entirety; the data may not lie precisely on the manifold, even though a low-dimensional structure may still exist. For instance, the data could reside within a tubular neighborhood of an unknown manifold. 

\item  In classical approximation theory, we prove theorems about the approximation error typically for smooth functions.
The class label function is generally not continuous; only piecewise continuous.
For example, if we classify objects with norm less than one, we have the smoothest possible boundary, but the target function  is the characteristic function of a ball. 
When the classes are disjoint, Stein extension theorems \cite{stein2016singular} can be used to extend such functions to arbitrarily smooth functions in theory.
This can be a challenging problem constructively, especially when the class boundaries may not be smooth.
Existence theorems for approximation by deep networks for the case when the class boundaries are assumed to have various smoothness properties are investigated, for example, in \cite{petersen2018optimal}.

\end{enumerate}

In recent years, two significant insights have emerged regarding these issues. First, the classical Belkin-Niyogi theorem can be modified to enable direct approximation on an \textbf{unknown} manifold, without the need to learn anything about the manifold except for its dimension \cite{mhaskar2019deep}. 
These results have been applied to diverse problems, such as blood sugar prediction in continuous glucose monitoring devices \cite{mhas_sergei_maryke_diabetes2017} and gesture recognition from radio frequency data \cite{mason2021manifold}. In both applications, our approach outperformed deep networks typically used for these tasks. 

The second insight is that the classification problem can be approached similarly to the problem of signal separation. Instead of isolating the locations of signal sources, we focus on separating the supports of the constituent probability distributions that define the different classes \cite{cloninger2020cautious}. This concept was applied to hyperspectral imaging, where it yielded superior results compared to the prevailing techniques at the time.
A notable aspect of this work is its hierarchical approach to classification. Additionally, the algorithm does not rely on labels at the outset. Instead, it suggests the optimal locations at which to query for labels, a process known as cautious active learning. This leads to effective classification using only a small number of carefully selected labels. In theory, the number of labels queried is equal to the number of classes, and this number is determined by the algorithm itself.

Originally, these ideas were developed for approximation on submanifolds of a Euclidean space and for separating the supports of measures defined on a Euclidean space. However, in some applications, it may prove more beneficial to project the Euclidean space onto a unit sphere. 
For regression problems, one advantage of this approach is the ability to leverage the compactness of the sphere, which reduces the number of samples required for approximation and facilitates the encoding of the target function. In classification problems, using Chebyshev polynomials instead of Hermite or ultraspherical polynomials can lead to significant computational savings.
\subsection{Learning on manifolds without manifold learning}\label{bhag;newregression}

The material in this section summarizes the results in \cite{mhaskar2024learning}.
In this section, we let $\XX$ to be a submanifold of an ambient sphere $\SS^Q$, with the dimension of $\XX$ being $q$.
We recall the kernel $\Phi_{n,q}$ defined in \eqref{eq:sphlockern}.
We assume a data set $\mathcal{D}=\{(x_j,y_j)\}$ drawn from a probability distribution $\tau$, and as usual, define $f=\mathbb{E}_\tau (y|x)$.
The smoothness class $W_\gamma$ is defined by (local) transition to the tangent sphere at any point using the corresponding definition of local smoothness on that sphere as described earlier in Section~\ref{bhag:local}.
\yadi{$f_0$}{Denisty of the data distribution with respect to $\mu^*$}
Our main theorem is the following.
\begin{theorem}\label{thm:mainthm}
We assume that
\begin{equation}\label{eq:newballmeasure}
    \sup_{x\in \mathbb{X},r>0}\frac{\mu^*(\mathbb{B}(x,r))}{r^q}\lesssim 1.
\end{equation}
Let $\mathcal{D}=\{(x_j,y_j)\}_{j=1}^M$ be a random sample  from a joint distribution $\tau$. 
We assume that the  marginal distribution of $\tau$ restricted to $\mathbb{X}$ is absolutely continuous with respect to $\mu^*$ with density $f_0$, and that the random variable $y$ has a bounded range.
Let
\begin{equation}\label{eq:fdef}
    f(x)\coloneqq \mathbb{E}_\tau(y|x),
\end{equation}
and
\begin{equation}\label{eq:approximation}
    \mathcal{F}_{n}(\mathcal{D};x)\coloneqq \frac{1}{M}\sum_{j=1}^M y_j\Phi_{n,q}(x\cdot x_j),\qquad x\in \mathbb{S}^Q,
\end{equation}
where $\Phi_{n,q}$ is defined in \eqref{eq:sphlockern}.

Let $0<\gamma<2$, $ff_0\in W_\gamma(\mathbb{X})$.
 Then for every $n\geq 1$, $0<\delta<1/2$ and 
 \be\label{eq:Mncond}
  M\gtrsim n^{q+2\gamma}\log(n/\delta),
  \ee
   we have with $\tau$-probability $\geq 1-\delta$:
\begin{equation}\label{eq:neapproxest}
    \norm{ \mathcal{F}_n(\mathcal{D};\circ)-ff_0}_\mathbb{X}\lesssim \frac{\sqrt{\norm{f_0}_{\mathbb{X}}}\norm{y}_{\mathbb{X}\times \Omega}+\norm{ff_0}_{W_\gamma(\mathbb{X})}}{n^\gamma}.
\end{equation}
\end{theorem}
 
 We observe a few salient features of this theorem.
 \begin{enumerate}
 \item The construction of $\mathcal{F}_n$ is truly universal; it makes no assumption about the nature of the target function or noise.
 \item The construction is straightforward. There is no need to learn anything about the underlying unknown manifold, aside from its dimension, and the process does not involve any optimization or eigen-decomposition
 \item The inner product involved in the definition \eqref{eq:approximation} can involve $x\in\SS^Q$. 
 So, the approximation gives an immediate expression for an out-of-sample extension, in contrast to the approach described in Section~\ref{bhag:manifoldlearning}.
 \item In the case when $f\equiv 1$, we obtain an estimate of the probability density $f_0$.
An interesting feature of this approach is that the kernel $\Phi_{n,q}$ is not a positive kernel, which contrasts with the majority of probability density estimation techniques we are familiar with.
 The non-positivity allows that there is no saturation in the approximation error; especially in the case when the manifold is a $q$ dimensional equator of $\SS^Q$.
 \item The theorem assumes that the measure $f_0d\mu^*$ is a probability measure. 
 In light of the above remark, we could bypass this condition by replacing $\mathcal{F}_n$ by
 $$\frac{M\mathcal{F}_n(\mathcal{D};x)}{\sum_{j=1}\Phi_{n,q}(x\cdot y_j)}
$$
\item We may encode the target function $f$ by
$$
\hat{y}(\ell,k)=\frac{1}{M}\sum_{j=1}^M y_jY_{\ell,k}(x_j).
$$
Then $\mathcal{F}_n$ acts as a decoder given by
$$
F_n(\mathcal{D};x)=\sum_{\ell=0}^n \Gamma_{\ell,n}\sum_{k=1}^{\operatorname{dim}(\mathbb{H}_\ell^Q)}\hat{y}(\ell,k)Y_{Q,\ell,k}(x)\qquad x\in \mathbb{S}^Q,
$$
where
$$
\Gamma_{\ell,n}\coloneqq \frac{\omega_q\omega_{Q-1}}{\omega_Q\omega_{q-1}}\sum_{i=\ell}^n h\left(\frac{i}{n}\right)\frac{p_{q,i}(1)}{p_{Q,\ell}(1)}C_{Q,q}(\ell,i),
$$
and $C_{Q,q}(\ell,i)$ are coefficients independent of $f$.
 \end{enumerate}

\subsection{Classification as signal separation}\label{bhag:newclassificiation}
The material in this section is based on \cite{Mhaskar2025Sep,mhaskar2025active}. 
The \emph{blind source signal separation problem} (also known as \emph{parameter estimation in exponential sums}) is the following.
Let $K\ge 2$ be an integer, $\{\omega_j\}_{j=1}^K\subset \TT$, $\{a_j\}_{j=1}^K \subset \CC$, $\mu=\sum_{j=1}^K a_j\delta_{\omega_j}$.
Given samples
\be\label{eq:expparameter}
\hat{\mu}(\ell)=\sum_{j=1}^K a_j\exp(-i\ell\omega_j), \qquad |\ell|<N,
\ee
for some positive integer $N$, determine $K$, $\omega_j$'s and $a_j$'s.
Out of these, obviously, the determination of $K$ and $\omega_j$'s are the harder problems; the  $a_j$'s can then be
determined by a solution of a system of linear equations.
Using the univariate version of the trigonometric kernel $\Phi_n$ introduced in \eqref{eq:op_kerndef}, we note that
$$
\sigma_n(\mu)(x)=\sum_{|\ell|<N}h\left(\frac{|\ell|}{N}\right)\hat{\mu}(\ell)\exp(i\ell x)=\sum_{k=1}^K a_k\Phi_n(x-\omega_k)\approx \mu(x),
$$
where the last $\approx$ is understood in a weak star sense.
In practice, the power spectrum $ |\sigma_n(\mu)(x)|$ can be partitioned into exactly $K$ clusters through careful thresholding, with each cluster $\Gamma_k$ containing precisely one of the points $\omega_k$. 
With $\hat{\omega}_k=\argmax_{x\in\Gamma_k}|\sigma_n(\mu)(x)|$, we have \cite{mhaskar2024robust}
$$
|\hat{\omega}_k-\omega_k|\lesssim 1/n.
$$
 While a detailed discussion of this phenomenon is beyond the scope of this paper, it provides the foundation for our insights into classification.

Thus, rather than considering a linear combination of Dirac delta functions, we consider a convex combination $\mu$ of the measures $ \mu_k$, from which points with label $k$ are sampled. Although the measures $\mu_k$ are unknown, $\mu$ represents the probability distribution from which the data is sampled. A point $x$ lies in the support of $\mu_k$ if the label associated with $x$ is $k$.
Thus, if we can separate the supports of $\mu_k$ from the samples taken from $\mu$, then we need to query the label only at one point in each of these supports.
This is similar to the signal separation problem, where the support of each $\mu_k$ is the point $\omega_k$.

As a reality check, we reproduce an example from \cite{Mhaskar2025Sep}.
\begin{example}\label{uda:measureseparation} {\rm
We define a distribution $\mu$ on $\TT$ as a convex combination of:
\bit
\item A sum of two uniform distributions each supported on $ [-0.6,-0.4]$, with a weight of $1200/3900$.

\item A normal distribution with mean $0.05$ and variance $0.04$, with a weight of $2400/3900$.


\item Three point-mass distributions at $-2,0.4,1.5$, with weights of $60/3900, 120/3900, 120/3900$ respectively (anomaly).
\eit

We take 3900 samples from this distribution (the number of points from each part of the distribution corresponding to the numerator of the weight). 
The samples from the distribution are visualized in Figure~\ref{fig:histogram} (a) as a normalized histogram. 
Then, we apply $\sigma_{128}$ to get an estimation of the support, as seen in Figure~\ref{fig:histogram} (b). 
Not only do we get an idea of the support of the distribution by looking at $\sigma_{128}$, but also the amplitudes of the atomic components of the distribution. 
 Since we are dealing with finitely many samples, we in fact are only estimating the integral in the definition of $\sigma_n$ as a Monte-Carlo type summation. That is, with data $\{u_j\}_{j=1}^M$ sampled randomly from $\mu$, we estimate
$$
\sigma_n(t)\approx \frac{1}{M}\sum_{j=1}^M\Phi_n(t- u_j).
$$

\begin{figure}[!ht]
\begin{center}
\begin{minipage}{.45\textwidth}
\begin{center}
\includegraphics[width=\textwidth]{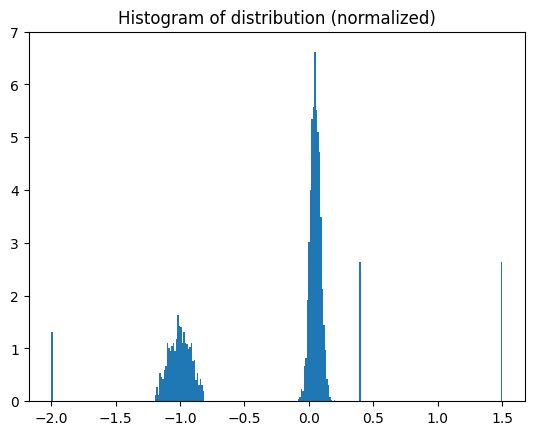}
\end{center}
\end{minipage}
\begin{minipage}{.45\textwidth}
\begin{center}
\includegraphics[width=\textwidth]{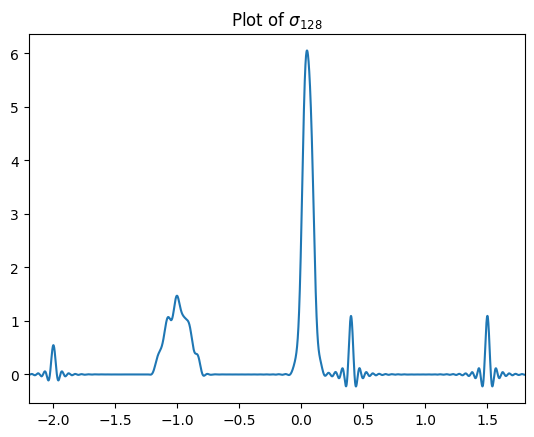}
\end{center}
\end{minipage}
\caption{Normalized histogram of the density of interest (left), paired with our density estimation by $\sigma_{128}$ based on $3900$ samples (right).} 
\label{fig:histogram}
\end{center}
\end{figure}
\qed}
\end{example}

There are several technical challenges in this context, beginning with the fact that we are observing samples drawn from $\mu$, rather than Fourier coefficients of $\mu$. Additionally, in the theory of signal separation, the minimal separation between the points $ \omega_k$ and the minimal amplitude $|a_k|$ are crucial for detecting the clusters. However, in the classification context, the supports of the measures $\mu_k$ may be continua, so that the minimal separation is $0$, and one needs to develop carefully the analogue of the minimal amplitude.

In this section, we describe our results in \cite{Mhaskar2025Sep}, where we propose a solution to these problems in the context of a metric space. 
Thus, we assume an ambient compact metric space $\mathbb{M}$, endowed with a metric $\rho$. 
For technical reasons, we assume that the diameter of $\mathbb{M}$ is $\pi$. 
The support of the data measure is a compact subset $\XX$ of $\mathbb{M}$.

We take a hierarchical classification approach.
We assume at each level $\eta$ of the minimal separation, there are $K_\eta$ classes of interest, which are separated by $\eta$, and an extra class $K_{\eta}+1$ of points in the overlapping classes. 
With some additional technical assumptions, we may then use the same theory developed for signal separation also for classification.

We now give some details.
In order to allow probabilities to taper off to $0$, we need a weaker version of the maximum and minimum amplitude as in the signal separation problem.
This is given in the following definition by \eqref{eq:maxamplitude} and \eqref{eq:minamplitude} respectively.
As usual, we denote the closed ball of radius $r$ centered at $x\in\mathbb{M}$ by $\BB(x,r)$.

\begin{definition}
\label{def:detectable}
We say a measure $\mu$ on $\mathbb{M}$ is \textbf{detectable} if there exist constants  $\kappa_1, \kappa_2, \alpha\geq 0$ such that for all $0< r\leq \pi$
\be\label{eq:maxamplitude}
\mu(\mathbb{B}(x,r))\le \kappa_1 r^\alpha\qquad x\in \mathbb{M}
\ee
and moreover, for some $r_0\le \pi$,
\be\label{eq:minamplitude}
\mu(\mathbb{B}(x,r))\ge \kappa_2 r^\alpha\qquad 0<r\le r_0.
\ee
\end{definition}

\begin{example}\label{ex:manifold}
{\rm
If $\mathbb{X}$ is a $\alpha$-dimensional, compact, connected, Riemannian manifold, then the normalized Riemannian volume measure is detectable with parameter $\alpha$. 
\qed}
\end{example}
The following definition makes precise the sentiments regarding minimal separation among classes.
In the following definition, $S_{k,\eta}$ is the set of points with label $k$ at minimal separation level $\eta$. 
To take into account overlapping class boundaries, we assume another set $S_{K_\eta+1,\eta}$ whose measure should tend to $0$ as $\eta\to 0$.
\begin{definition}
\label{def:finestructure}
We say a measure $\mu$ has a \textbf{fine structure} if there exists an $\eta_0$ such that for every $\eta\in (0,\eta_0]$ there is an integer $K_\eta$ and a partition $\mathbb{X}=\{S_{k,\eta}\}_{k=1}^{K_\eta+1}$ where each of the following are satisfied.
\begin{enumerate}

\item (\textbf{Cluster Minimal Separation}) For any $j,k=1,2,\dots,K_\eta$ with $j\neq k$ we have
\be
\mathsf{dist}(S_{j,\eta},S_{k,\eta})\geq 2\eta.
\ee

\item (\textbf{Exhaustion Condition}) We have
\be
\lim_{\eta\to 0^+} \mu(S_{K_\eta+1,\eta})=0.
\ee
\end{enumerate}
\end{definition} 
\begin{example}\label{ex:pointmass}
{\rm
Supposing that $\mu=\sum_{k=1}^K a_k \delta_{\omega_k}$ (as in the signal separation problem), then we see that $\mu$ is detectable with $\alpha=0$, $\kappa_1=\max_k |a_k|$, $\kappa_2=\min_k |a_k|$. 
It has fine structure in the classical sense  whenever $\eta<\min_{j\neq k} |\omega_j-\omega_k|$. In this sense, the theory presented in this paper is a generalization of results for signal separation in this regime.
\qed} 
\end{example}
In the remainder of this section, constants used in the notation such as $\lesssim$ will depend upon $\kappa_1, \kappa_2$ as well.
\begin{example}\label{ex:manifoldbis}
{\rm
Let the support $\XX_k$ of each $\mu_k$ be a manifold of dimension $q$, so that $\XX=\cup_{k=1}^K\XX_k$, and $\mu$ is a convex combination of $\mu_k$'s. 
Cleary, $\mu$ is a detectable measure.
If the $\XX_k$'s are disjoint, and $2\eta_0$ is the minimal separation among these, then we may take for each $\eta\le\eta_0$, $K_\eta=\emptyset$ and $\mathcal{S}_{k,\eta}=\XX_k$ for each $k$.
If we allow, say $\XX_1\cap\XX_2\not=\emptyset$, then for any $\eta$, we may take out an $\eta$ neighborhood of this intersection, resulting in $K_\eta=K$, $\mathcal{S}_{K_\eta+1,\eta}$ to be this intersection, and $\mathcal{S}_{k,\eta}=\XX_k\setminus \mathcal{S}_{K_\eta+1,\eta}$. 
Obviously, the classification problem is not well defined for points in $\mathcal{S}_{K_\eta+1,\eta}$.
The exhaustion condition requires that the measure of this intersection should become smaller and smaller as $\eta\to 0$.
\qed}
\end{example}
Since our primary objective is to identify the supports of the measures, rather than approximating the measures themselves, we require a positive localized kernel to prevent interference between measures. 

In this subsection, we define
\begin{equation}\label{eq:chebkernel}
\Psi_n(x,  y)= \left\{\sum_{\ell=0}^{n}h\left(\frac{\ell}{n}\right)\exp(i\ell\rho(x,y))\right\}^2.
\end{equation}

The main theorems are as follows. The first theorem demonstrates how the support of the data measure $\mu$ can be detected using the kernel $\Psi_n$.
We define the \textbf{measure support estimator} by
\be\label{eq:support_estimator_def}
F_n(x)\coloneqq \frac{1}{M}\sum_{j=1}^M \Psi_n(x,x_j),
\ee
and the \textbf{support estimator set} by
\be\label{eq:support_est_set_def}
\mathcal{G}_n(\Theta)= \left\{x\in \mathbb{M}: F_n(x)\geq \Theta \max_{1\leq k\leq M} F_n(x_k)\right\}.
\ee
\begin{theorem}\label{thm:fullsuportdet}
Let $\mu$ be detectable and suppose $M\gtrsim n^{\alpha}\log(n)$. Let $\{x_1,x_2,\dots,x_M\}$ be independent samples from $\mu$. There exists a constant $C>0$ such that if $\Theta<C<1$, then there exists $r(\Theta)\sim \Theta^{-1/(S-\alpha)}$ such that with probability at least $1-c_1/M^{c_2}$ we have
\be\label{eq:thm1}
\mathbb{X}\subseteq \mathcal{G}_n(\Theta)\subseteq \mathbb{B}\left(\mathbb{X},r(\Theta)/n\right).
\ee
\end{theorem}

The next theorem makes precise the sentiment that the support of $\mu$ should be separated into a number of clusters, each approximating the support of one class.

\begin{theorem}\label{thm:class_separation}
Suppose, in addition to the assumptions of Theorem~\ref{thm:fullsuportdet}, that $\mu$ has a fine structure, $n\gtrsim 1/(\eta\Theta^{1/(S-\alpha)})$, and $\mu(\mathbf{S}_{K_{\eta}+1,\eta})\lesssim \Theta n^{-\alpha}$. Define
\be\label{eq:thm2-1}
\mathcal{G}_{k,\eta,n}(\Theta)\coloneqq \mathcal{G}_n(\Theta)\cap \mathbb{B}(\mathbf{S}_{k,\eta},r(\Theta)/n).
\ee
Then, with probability at least $1-c_1/M^{c_2}$, $\{\mathcal{G}_{k,\eta,n}(\Theta)\}_{k=1}^{K_\eta}$ is a partition of $\mathcal{G}_{n}(\Theta)$ such that 
\be\label{eq:thm2-2}
\operatorname{dist}(\mathcal{G}_{j,\eta,n}(\Theta),\mathcal{G}_{k,\eta,n}(\Theta))\geq \eta\qquad j\neq k,
\ee 
and in this case, there exists $c<1$ such that
\be\label{eq:thm2-3}
\mathbb{X}\cap \mathbb{B}(\mathbf{S}_{k,\eta},cr(\Theta)/n)\subseteq \mathcal{G}_{k,\eta,n}(\Theta)\subseteq \mathbb{B}\left(\mathbf{S}_{k,\eta},r(\Theta)/n\right).
\ee
\end{theorem}

\begin{figure}[ht]
    \begin{center}
    \includegraphics[width=0.4\textwidth]{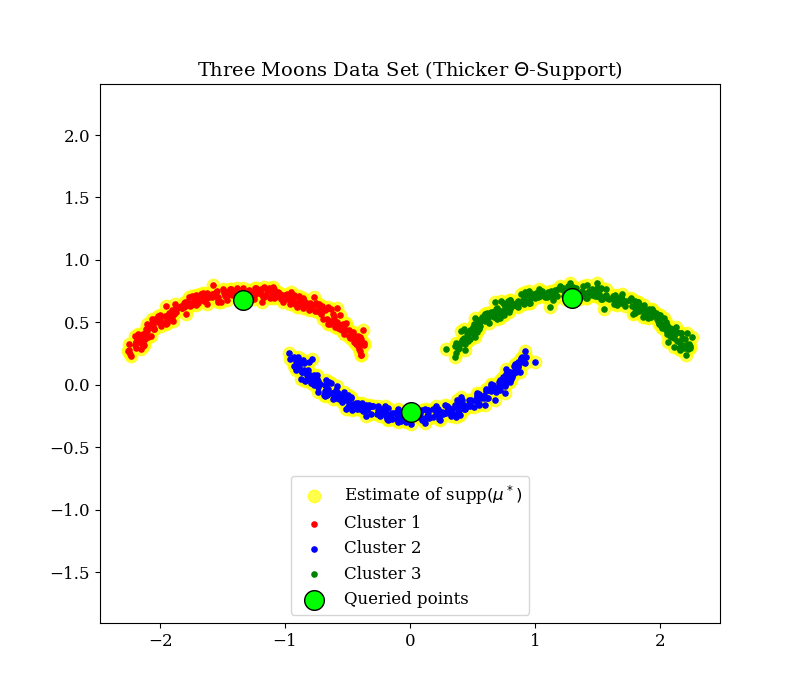}
    \end{center}
    \caption{Support estimation and active learning on a multi-class three-moons data set.
        This figure illustrates the geometric classification paradigm underlying
    the multi-class extension of Theorem~10.2.
    The yellow region represents the estimated support
    $\mathcal{G}_n(\Theta) \approx \operatorname{supp}(\mu)$ of the data distribution.
    The estimator correctly recovers three disjoint connected components of the support.
    By querying the label of a single point in each connected component
    (indicated by green circles), the active learning strategy propagates labels
    across the entire manifold using geometric connectivity.
    As a result, all points are correctly classified, achieving $100\%$ accuracy
    with only $k=3$ label queries, equal to the number of connected components.}
    \label{fig:three_moons}

\end{figure}

\begin{figure}
\begin{center}
\begin{minipage}[t]{.45\textwidth}
\begin{center}
\includegraphics[width=\textwidth]{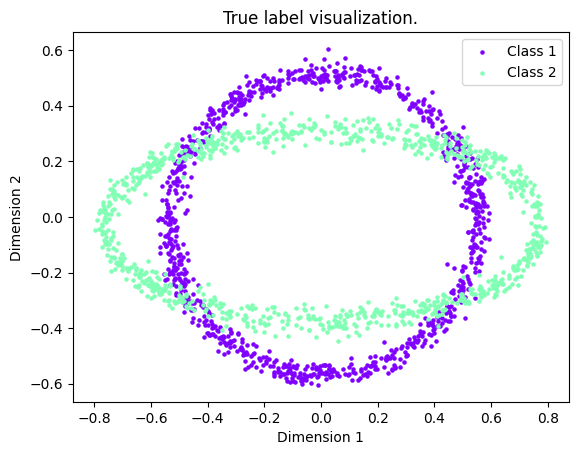}
\end{center}
\end{minipage}
\begin{minipage}[t]{.45\textwidth}
\begin{center}
\includegraphics[width=\textwidth]{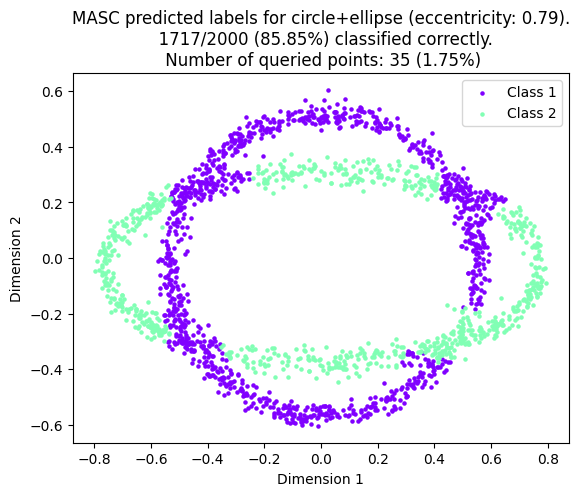}
\end{center}
\end{minipage}
\caption{This figure illustrates the result of applying MASC to a synthetic circle and ellipse data set. On the left are true labels of the given data, and on the right is the estimation attained by MASC. MASC predicted the labels with 35 queries, achieving 83\% accuracy.}
\label{fig:circleellipse}
\end{center}
\end{figure}

An algorithm called MASC to illustrate the use of these theorems in various real problems is described in detail in \cite{Mhaskar2025Sep}, together with a detailed discussion about the choice of the parameters. 
We will not go into those details here, but reproduce from 
\cite{Mhaskar2025Sep} an example to illustrate classification when the supports of the class measures intersect.
We generated a synthetic data set of 1000 points sampled along the arclength of a circle and another 1000 sampled along the arclength of an ellipse with eccentricity 0.79. 
For each data point, normal noise with standard deviation 0.05 was additively applied independently to both components. Figure~\ref{fig:circleellipse} shows the true class label for each of the points on the left and the estimated class labels on the right. 
We can see that the misclassifications are mostly localized to the area where the supports of the two measures overlap.
 Near the intersection points of the circle and ellipse the classification problem becomes extremely difficult due to a high probability that a data point could have been sampled from either the circle or ellipse.

\bhag{Transformers/attention mechanism and local kernels}\label{bhag:transformer}

In this section we express some thoughts on how the theory described so far can be applied to the theory of transformers.
According to \cite{geshkovski2023mathematical}, the main task of a tranformer is to predict the next token given a string of $q$ tokens, analogous in theory, but a far more complicated task than the prediction problem in time series.
The meaning of the term token depends upon the application, and is probably proprietary information.
With an appropriate embedding, a token may be considered as a vector in $\RR^D$. 

The transformer consists of a cascade of machines comprising pairs of attention and deep networks.
The purpose of the attention mechanism is  to figure out which token among the ``keys'' is closer to the  ``query'', and return the corresponding ``value''. 
The value might be a better feature selection for the tokens.
The deep network then finishes the task of predicting the next token.

The attention mechanism is formulated as follows. 
First, the softmax function is defined by
\be\label{eq:softmax}
\mathsf{softmax}(z_1,\cdots, z_q)=\left(\sum_{k=1}^q \exp(z_k)\right)^{-1}\left(\exp(z_1), \cdots, \exp(z_q)\right).
\ee
Let $\mathbf{X}=(\x_k,\cdots, \x_{k+m-1}, \x_{k+m})$ be a $D\times (m+1)$ matrix. With  $d\times D$ ``query'' matrices $\mathbf{Q}$, a $d\times D$  ``key'' matrix $\mathbf{K}$, and a $(m+1)\times (m+1)$ ``value'' matrix $\mathbf{V}$, the attention is defined by
\be\label{eq:attention}
\mathsf{att}(\mathbf{X})=\mathsf{softmax}\left(\beta (\mathbf{Q}\mathbf{X})^T\mathbf{K}\mathbf{X}\right)\mathbf{V}\mathbf{X},
\ee
where the softmax is evaluated column-wise.
In particular, if we denote the $\ell$-th column of  $\mathbf{K}\mathbf{X}$ by $\mathbf{y}_\ell$, and that of $\mathbf{V}\mathbf{X}$ by $\mathbf{v}_\ell$, and the $s$-th component of $\mathbf{Q}^T\mathbf{X}$ by $\mathbf{q}_s$, then assuming layer normalization, the formula \eqref{eq:attention} involves (up to a multiplicative factor) a sum of the form
\be\label{eq:attnsbf}
\sum_{\ell=1}^d \exp\left(\beta\mathbf{q}_s\cdot\mathbf{y}_\ell\right)\mathbf{v}_\ell.
\ee
We recognize this immediately as a spherical basis function (SBF) network, for which the approximation theory is well developed (Section~\ref{bhag:kernelbased}). 
This implies in particular, that for the objective of function approximation alone, there is no intrinsic need to have multiple heads or different architectures for deep networks and attention mechanisms. 
Of course, in practice, one may want to approximate different features from the data, and refine the approximations.
Likewise, there may be different functions to be approximated for different queries, and if the machine has to be prepared for all these queries, it makes sense to use different heads, each to handle a certain kind of queries.
One implication of this observation is that we need not distinguish between attention and transformer mechanisms; they both can be combined into a single deep network with each channel evaluating an SBF network as described.
More generally, to cure the high dimensionality of the tokens, one may treat the tokens as belonging to some unknown manifold, in which case the theory developed in Section~\ref{bhag;newregression} may be applied. 

Standard Transformer architectures are shown to be universal approximators for continuous functions on compact domains \cite{yun2019transformers}. Their expressive power is deeply linked to the self-attention mechanism, which allows the model to capture long-range dependencies and complex interactions that are often unattainable by bounded-depth convolutional or recurrent networks. The generalization capabilities of Transformers are often attributed to their ability to learn structured representations of data. Research suggests that the training dynamics of self-attention exhibit an implicit bias toward low-rank solutions or specific sparsity patterns, which helps the model generalize from finite training sets to unseen sequences \cite{li2023transformers}.

\bhag{Computational considerations}\label{bhag:compute}

In this paper, we have discussed a number of ideas for function approximation on different domains. 
Some of the suggested methods are probably not well known.
In this section, we comment upon the computation of localized kernels and MZ quadrature formulas.
In the context of trigonometric polynomials, the kernels can be computed easily using FFT or non-uniform FFT (NFFT) introduced by Keiner, Kunis, and Potts \cite{keiner2009using}.
Similar techniques are introduced for ultraspherical polynomials in  \cite{potts1998fast, potts2003fast}.
It is demonstrated in \cite{mhaskar2024learning} how the localized kernels on the sphere can be computed using Clenshaw algorithm implemented as a deep/recurrent network.
For the kernels required for the use of diffusion geometry methods, it is customary to compute the eigenfunctions approximately by computing the eigenvectors of  a graph Laplacian.
We are not aware of fast computations for these.

A most general algorithm for computing quadrature measures is given in \cite{quadconst}. 
This is done theoretically by constructing an orthogonal polynomial system on the scattered data. 
In practice, this amounts to finding a solution of an underdetermined system of linear equations.
In the context of approximation on the sphere, we have illustrated the use of our algorithm and the superior performance of our localized kernel approximation of a set of benchmark examples.
Fast but approximate constructions for quadrature formulas on a two dimensional sphere are described in \cite{pottssphquadconst}.
It is open problem to find similar constructions for higher dimensional spheres.
The general technique described in \cite{quadconst} have been applied to the case of the eigenfunctions on an unknown manifold in the diffusion geometry setting, but this effort is not very successful in practice.

\bhag{Summary and discussion}\label{bhag:summary}
This work explores the intersection of approximation theory and machine learning, highlighting key theoretical results and the existing gaps between the two fields. 
We begin by framing the central problem in machine learning as constructing a functional model that generalizes well from a given dataset, emphasizing the role of neural networks and kernel-based methods in achieving this goal. 
Despite the fundamental importance of function approximation, classical approximation theory has not been fully integrated into the theoretical foundations of machine learning, leading to challenges in understanding model generalization.
 This work first provides an overview of approximation theory, discussing smoothness classes, trigonometric approximations, and clarifying the often misunderstood notion of the curse of dimensionality. 
 It then explores the expressive power of neural networks, differentiating between shallow and deep architectures, and reviewing key results such as universal approximation theorems and the efficiency gains offered by deep networks. 
 While most of the approximation theorems are proved in known domains such as a high dimensional cube, this results in a gap between theory and practice, where the data does not fill up the whole domain. 
 The theory of manifold learning seeks to bridge this gap. 
 The authors therefore examine manifold learning, in particular, the theory of function approximation under the manifold assumption that the data distribution lies on a low-dimensional smooth manifold.  
 More generally, the authors examine the notion of data spaces, and study kernel based approximation in this general setting. 
 A novelty of this study is the notion of local approximation and wavelet-like expansions based on the systems developed for manifold learning. 
 The study then delves into function approximation on spheres in particular, and highlights the applications to the theory of ReLU networks and operator approximations in this context. 
 We also examined the error estimates associated with physics-informed neural surrogates, with a particular focus on Physics-Informed Neural Networks (PINNs). PINNs have been extensively utilized in solving a wide range of PDEs, encompassing both deterministic and stochastic formulations.
  Additionally, these networks have been effectively applied to nonlocal and fractional PDEs, further demonstrating their versatility in addressing complex mathematical models across various scientific and engineering domains. 
  Toward the end of the paper, we also discussed the limitations of the classical machine learning paradigm as well as those of the classical approximation theory in addressing these limitations. In particular, we review some recent work proposing a new paradigm that achieves function approximation on manifolds without first approximating characteristics of the manifold, such as an atlas or Laplace-Beltrami eigensystem. Another novelty of the new paradigm is to treat classification as a signal separation problem. This perspective is particularly relevant in cases where class boundaries are non-smooth or closely situated, making traditional classification frameworks theoretically insufficient. The review also touches upon emerging trends, such as the integration of transformers and attention mechanisms with local kernel methods, and concludes by identifying several open research questions that warrant further exploration. Overall, this article provides a thorough examination of approximation theory’s role in machine learning, offering insights into its theoretical foundations, practical limitations, and potential future directions.

\subsection{Some open questions}\label{bhag:openquest}
This theory encompasses numerous deep and open questions, a few of which we highlight below.
\begin{enumerate}
\item The entire theory depends upon working with the right features. 
It is believed widely that deep networks find the correct features automatically.
However, this seems to us a vicious circle: the networks work because they detect the right features. In turn, these must have been detected since the networks work.
Finding the right features with theoretical guarantees is an important question. 
For some results in this direction, we point to the work by Belkin and his collaborators \cite{radhakrishnan2024linear}.
\item Related to the question of detecting the right features is the design of a deep network to extract the right features in a compositional manner. 
If one can figure out the right features to be refined in an iterative manner, then one can use our constructive theory to design a deep network without explicit training.
\item Find converse theorems in the theory of neural networks.
For homogeneous activation functions, we know such a theory once we transfer the networks from the Euclidean space to the sphere, provided that the activation functions are positive definite.
ReLU$^\gamma$ functions do not satisfy this condition.
Since both dimension independent bounds as well as constructions leading to dimension dependent bounds are known, we conjecture that the width results in this context would be different from the converse theorems.
The converse theorem depends only on the degree of approximation, not the manner in which it is constructed; the width results depend upon the data being selected in a continuous manner.
\item  The width results deal with the number of parameters. 
As we have seen, the converse theorems for eignets depend upon the minimal separation among the centers.
We may think of the number of parameters or the minimal separation as the cost of doing approximation.
It would be desirable to develop a notion of widths that depends upon a cost that is not necessarily integer-valued.
A recent paper  \cite{siegel2024sharp} by Siegel describes a notion of spherical width, which might be a good starting point for such a theory.
\item The theory in Section~\ref{bhag:newclassificiation} describes classification as a clustering problem, proving the existence of clusters satisfying certain properties.
However, characterizing and finding these clusters provably remains an open question.
\item The theory described in Section~\ref{bhag:operatorapprox} depends upon having the right quadrature formula and the right localized kernel.
It is therefore important to devise algorithms that yield the quadrature formulas with the requisite properties in a tractable manner; i.e., where the number of samples is a tractable multiple of the dimension of the polynomial space.
It is also desirable to find an alternative expression for the kernel analogous to the one defined in \eqref{eq:dataspace_kern_gen}.
\end{enumerate} 

\section*{Acknowledgments}
The research of H. N. Mhaskar is supported in part by ONR grants N00014-23-1-2394 and N00014-23-1-2790. The research of E. Tsoukanis is supported in part by ONR grant N00014-23-1-2790.
We would like to thank Helmut B\"olcskei, Alexander Cloninger,  Phillipp Petersen, Jonathan Siegel, Felix Voigtl\"nder, Ding-Xuan Zhou for their valuable comments on the first draft of this paper.
We thank Quoc Thong Le Gia for supplying us with Figure~\ref{fig:sph_loc_kern} and the data discussed in Remark~\ref{rem:num_examples}. We also thank the reviewers for their critical assessment of our manuscript, which has significantly improved the revised version.

\bibliographystyle{abbrv}
\bibliography{Approxi15}

@article{ito1996nonlinearity,
  title={Nonlinearity creates linear independence},
  author={Ito, Yoshifusa},
  journal={Advances in Computational Mathematics},
  volume={5},
  number={1},
  pages={189--203},
  year={1996},
  publisher={Springer}
}

@book{abu2012learning,
  title={Learning from data},
  author={Abu-Mostafa, Yaser S and Magdon-Ismail, Malik and Lin, Hsuan-Tien},
  volume={4},
  year={2012},
  publisher={AMLBook New York}
}

@article{auricchio2024accuracy,
  title={On the accuracy of interpolation based on single-layer artificial neural networks with a focus on defeating the Runge phenomenon},
  author={Auricchio, Ferdinando and Belardo, Maria Roberta and Calabr{\`o}, Francesco and Fabiani, Gianluca and Pascaner, Ariel F},
  journal={Soft Computing},
  volume={28},
  number={20},
  pages={11767--11785},
  year={2024},
  publisher={Springer}
}

@article{huang2015trends,
  title={Trends in extreme learning machines: A review},
  author={Huang, Gao and Huang, Guang-Bin and Song, Shiji and You, Keyou},
  journal={Neural Networks},
  volume={61},
  pages={32--48},
  year={2015},
  publisher={Elsevier}
}

@article{abbasi2025challenges,
  title = {Challenges and advancements in modeling shock fronts with physics-informed neural networks: A review and benchmarking study},
   author = {Jassem Abbasi and Ameya D. Jagtap and Ben Moseley and Aksel Hiorth and P{\aa}l {\O}steb{\o}},
  journal = {Neurocomputing},
  volume = {657},
  pages = {131440},
  year = {2025},
  publisher={Elsevier}
}

@article{menon2025anant,
  title={Anant-Net: Breaking the Curse of Dimensionality with Scalable and Interpretable Neural Surrogate for High-Dimensional PDEs},
  author={Menon, Sidharth S and Jagtap, Ameya D},
  journal = {Computer Methods in Applied Mechanics and Engineering},
  volume = {447},
  pages = {118403},
  year = {2025},
  publisher={Elsevier}
}

@article{yun2019transformers,
  title={Are transformers universal approximators of sequence-to-sequence functions?},
  author={Yun, Chulhee and Bhojanapalli, Srinadh and Rawat, Ankit Singh and Reddi, Sashank J and Kumar, Sanjiv},
  journal={arXiv preprint arXiv:1912.10077},
  year={2019}
}

@article{wang2021loss,
  title={Understanding and mitigating gradient pathologies in physics-informed neural networks},
  author={Wang, Sifan and Teng, Yujun and Perdikaris, Paris},
  journal={SIAM Journal on Scientific Computing},
  volume={43},
  number={5},
  pages={A3055--A3081},
  year={2021}
}

@article{krishnapriyan2021failure,
  title={Characterizing possible failure modes in physics-informed neural networks},
  author={Krishnapriyan, Ameya and Gholami, Ali and Zhe, Shandian and Kirby, Robert and Mahoney, Michael},
  journal={Advances in Neural Information Processing Systems},
  volume={34},
  pages={26548--26560},
  year={2021}
}

@inproceedings{li2023transformers,
  title={Transformers as algorithms: Generalization and stability in in-context learning},
  author={Li, Yingcong and Ildiz, Muhammed Emrullah and Papailiopoulos, Dimitris and Oymak, Samet},
  booktitle={International conference on machine learning},
  pages={19565--19594},
  year={2023},
  organization={PMLR}
}

@article{lu2021learning45,
  title={Learning nonlinear operators via DeepONet based on the universal approximation theorem of operators},
  author={Lu, Lu and Jin, Pengzhan and Pang, Guofei and Zhang, Zhongqiang and Karniadakis, George Em},
  journal={Nature machine intelligence},
  volume={3},
  number={3},
  pages={218--229},
  year={2021},
  publisher={Nature Publishing Group UK London}
}

@article{de2022error,
  title={Error analysis for physics-informed neural networks (PINNs) approximating Kolmogorov PDEs},
  author={De Ryck, Tim and Mishra, Siddhartha},
  journal={Advances in Computational Mathematics},
  volume={48},
  number={6},
  pages={79},
  year={2022},
  publisher={Springer}
}

@article{de2024numerical,
  title={Numerical analysis of physics-informed neural networks and related models in physics-informed machine learning},
  author={De Ryck, Tim and Mishra, Siddhartha},
  journal={Acta Numerica},
  volume={33},
  pages={633--713},
  year={2024},
  publisher={Cambridge University Press}
}

@article{zeinhofer2024unified,
  title={A unified framework for the error analysis of physics-informed neural networks},
  author={Zeinhofer, Marius and Masri, Rami and Mardal, Kent--Andr{\'e}},
  journal={IMA Journal of Numerical Analysis},
  pages={drae081},
  year={2024},
  publisher={Oxford University Press}
}

@article{de2024errorEE,
  title={Error estimates for physics-informed neural networks approximating the Navier--Stokes equations},
  author={De Ryck, Tim and Jagtap, Ameya D and Mishra, Siddhartha},
  journal={IMA Journal of Numerical Analysis},
  volume={44},
  number={1},
  pages={83--119},
  year={2024},
  publisher={Oxford University Press}
}

@inproceedings{hillebrecht2022certified,
  title={Certified machine learning: A posteriori error estimation for physics-informed neural networks},
  author={Hillebrecht, Birgit and Unger, Benjamin},
  booktitle={2022 International Joint Conference on Neural Networks (IJCNN)},
  pages={1--8},
  year={2022},
  organization={IEEE}
}

@article{kharazmi2021hp,
  title={hp-VPINNs: Variational physics-informed neural networks with domain decomposition},
  author={Kharazmi, Ehsan and Zhang, Zhongqiang and Karniadakis, George Em},
  journal={Computer Methods in Applied Mechanics and Engineering},
  volume={374},
  pages={113547},
  year={2021},
  publisher={Elsevier}
}

@article{raissi2019physics,
  title={Physics-informed neural networks: A deep learning framework for solving forward and inverse problems involving nonlinear partial differential equations},
  author={Raissi, Maziar and Perdikaris, Paris and Karniadakis, George E},
  journal={Journal of Computational physics},
  volume={378},
  pages={686--707},
  year={2019},
  publisher={Elsevier}
}

@article{jagtap2020adaptive,
  title={Adaptive activation functions accelerate convergence in deep and physics-informed neural networks},
  author={Jagtap, Ameya D and Kawaguchi, Kenji and Karniadakis, George Em},
  journal={Journal of Computational Physics},
  volume={404},
  pages={109136},
  year={2020},
  publisher={Elsevier}
}

@article{jagtap2022deepknn,
  title={Deep Kronecker neural networks: A general framework for neural networks with adaptive activation functions},
  author={Jagtap, Ameya D and Shin, Yeonjong and Kawaguchi, Kenji and Karniadakis, George Em},
  journal={Neurocomputing},
  volume={468},
  pages={165--180},
  year={2022},
  publisher={Elsevier}
}

@article{jagtap2020conservative,
  title={Conservative physics-informed neural networks on discrete domains for conservation laws: Applications to forward and inverse problems},
  author={Jagtap, Ameya D and Kharazmi, Ehsan and Karniadakis, George Em},
  journal={Computer Methods in Applied Mechanics and Engineering},
  volume={365},
  pages={113028},
  year={2020},
  publisher={Elsevier}
}

@article{abbasi2025history,
  title={History-Matching of imbibition flow in fractured porous media Using Physics-Informed Neural Networks (PINNs)},
  author={Abbasi, Jassem and Moseley, Ben and Kurotori, Takeshi and Jagtap, Ameya D and Kovscek, Anthony R and Hiorth, Aksel and Andersen, P{\aa}l {\O}steb{\o}},
  journal={Computer Methods in Applied Mechanics and Engineering},
  volume={437},
  pages={117784},
  year={2025},
  publisher={Elsevier}
}

@article{jagtap2022deepWW,
  title={Deep learning of inverse water waves problems using multi-fidelity data: Application to Serre--Green--Naghdi equations},
  author={Jagtap, Ameya D and Mitsotakis, Dimitrios and Karniadakis, George Em},
  journal={Ocean Engineering},
  volume={248},
  pages={110775},
  year={2022},
  publisher={Elsevier}
}

@article{penwarden2023unified,
  title={A unified scalable framework for causal sweeping strategies for physics-informed neural networks (PINNs) and their temporal decompositions},
  author={Penwarden, Michael and Jagtap, Ameya D and Zhe, Shandian and Karniadakis, George Em and Kirby, Robert M},
  journal={Journal of Computational Physics},
  volume={493},
  pages={112464},
  year={2023},
  publisher={Elsevier}
}

@inproceedings{raonic2023convolutional,
  title={Convolutional neural operators},
  author={Raonic, Bogdan and Molinaro, Roberto and Rohner, Tobias and Mishra, Siddhartha and de Bezenac, Emmanuel},
  booktitle={ICLR 2023 Workshop on Physics for Machine Learning},
  year={2023}
}

@article{shin2020error,
  title={Error estimates of residual minimization using neural networks for linear PDEs},
  author={Shin, Yeonjong and Zhang, Zhongqiang and Karniadakis, George Em},
  journal={arXiv preprint arXiv:2010.08019},
  year={2020}
}

@article{shin2024theoretical,
  title={Theoretical foundations of physics-informed neural networks and deep neural operators: A brief review},
  author={Shin, Yeonjong and Zhang, Zhongqiang and Karniadakis, George Em},
  year={2024},
  publisher={Elsevier}
}

@article{mishra2021physics,
  title={Physics informed neural networks for simulating radiative transfer},
  author={Mishra, Siddhartha and Molinaro, Roberto},
  journal={Journal of Quantitative Spectroscopy and Radiative Transfer},
  volume={270},
  pages={107705},
  year={2021},
  publisher={Elsevier}
}

@article{mishra2022estimates,
  title={Estimates on the generalization error of physics-informed neural networks for approximating a class of inverse problems for PDEs},
  author={Mishra, Siddhartha and Molinaro, Roberto},
  journal={IMA Journal of Numerical Analysis},
  volume={42},
  number={2},
  pages={981--1022},
  year={2022},
  publisher={Oxford University Press}
}

@inproceedings{fanaskov2023spectral,
  title={Spectral neural operators},
  author={Fanaskov, Vladimir Sergeevich and Oseledets, Ivan V},
  booktitle={Doklady Mathematics},
  volume={108},
  number={Suppl 2},
  pages={S226--S232},
  year={2023},
  organization={Springer}
}

@article{shin2020convergence,
  title={On the convergence of physics informed neural networks for linear second-order elliptic and parabolic type PDEs},
  author={Shin, Yeonjong and Darbon, Jerome and Karniadakis, George Em},
  journal={arXiv preprint arXiv:2004.01806},
  year={2020}
}

@article{peyvan2024riemannonets,
  title={RiemannONets: Interpretable neural operators for Riemann problems},
  author={Peyvan, Ahmad and Oommen, Vivek and Jagtap, Ameya D and Karniadakis, George Em},
  journal={Computer Methods in Applied Mechanics and Engineering},
  volume={426},
  pages={116996},
  year={2024},
  publisher={Elsevier}
}

@article{menon2026scientific,
  title={On Scientific Foundation Models: Rigorous Definitions, Key Applications, and a Comprehensive Survey},
  author={Menon, Sidharth S and Mondal, Trishit and Brahmachary, Shuvayan and Panda, Aniruddha and Joshi, Subodh M and Kalyanaraman, Kaushic and Jagtap, Ameya D},
  journal={Neural Networks},
  volume = {198},
  pages={108567},
  year={2026},
  publisher={Elsevier}
}

@article{shukla2021parallel,
  title={Parallel physics-informed neural networks via domain decomposition},
  author={Shukla, Khemraj and Jagtap, Ameya D and Karniadakis, George Em},
  journal={Journal of Computational Physics},
  volume={447},
  pages={110683},
  year={2021},
  publisher={Elsevier}
}

@article{mao2020physics,
  title={Physics-informed neural networks for high-speed flows},
  author={Mao, Zhiping and Jagtap, Ameya D and Karniadakis, George Em},
  journal={Computer Methods in Applied Mechanics and Engineering},
  volume={360},
  pages={112789},
  year={2020},
  publisher={Elsevier}
}

@article{goswami2024learning,
  title={Learning stiff chemical kinetics using extended deep neural operators},
  author={Goswami, Somdatta and Jagtap, Ameya D and Babaee, Hessam and Susi, Bryan T and Karniadakis, George Em},
  journal={Computer Methods in Applied Mechanics and Engineering},
  volume={419},
  pages={116674},
  year={2024},
  publisher={Elsevier}
}

@article{shukla2021physics,
  title={A physics-informed neural network for quantifying the microstructural properties of polycrystalline nickel using ultrasound data: A promising approach for solving inverse problems},
  author={Shukla, Khemraj and Jagtap, Ameya D and Blackshire, James L and Sparkman, Daniel and Karniadakis, George Em},
  journal={IEEE Signal Processing Magazine},
  volume={39},
  number={1},
  pages={68--77},
  year={2021},
  publisher={IEEE}
}

@article{jagtap2022physics,
  title={Physics-informed neural networks for inverse problems in supersonic flows},
  author={Jagtap, Ameya D and Mao, Zhiping and Adams, Nikolaus and Karniadakis, George Em},
  journal={Journal of Computational Physics},
  volume={466},
  pages={111402},
  year={2022},
  publisher={Elsevier}
}

@article{hu2021extended,
  title={When do extended physics-informed neural networks (XPINNs) improve generalization?},
  author = {Hu, Zheyuan and Jagtap, Ameya D. and Karniadakis, George Em and Kawaguchi, Kenji},
  journal = {SIAM Journal on Scientific Computing},
  volume = {44},
  number = {5},
  pages = {A3158-A3182},
  year = {2022}
}

@article{jagtap2020locally,
  title={Locally adaptive activation functions with slope recovery for deep and physics-informed neural networks},
  author={Jagtap, Ameya D and Kawaguchi, Kenji and Em Karniadakis, George},
  journal={Proceedings of the Royal Society A},
  volume={476},
  number={2239},
  pages={20200334},
  year={2020},
  publisher={The Royal Society}
}

@article{jagtap2023important,
  title={How important are activation functions in regression and classification? A survey, performance comparison, and future directions},
  author={Jagtap, Ameya D and Karniadakis, George Em},
  journal={Journal of Machine Learning for Modeling and Computing},
  volume={4},
  number={1},
  year={2023},
  publisher={Begel House Inc.}
}

@article{zhang2026bubbleokan,
  title={BubbleOKAN: A physics-informed interpretable neural operator for high-frequency bubble dynamics},
  author={Zhang, Yunhao and Menon, Sidharth S and Cheng, Lin and Gnanaskandan, Aswin and Jagtap, Ameya D},
  journal={Computer Methods in Applied Mechanics and Engineering},
  volume={450},
  pages={118667},
  year={2026},
  publisher={Elsevier}
}

@article{jagtap2020extended,
  title={Extended physics-informed neural networks (XPINNs): A generalized space-time domain decomposition based deep learning framework for nonlinear partial differential equations},
  author={Jagtap, Ameya D and Karniadakis, George Em},
  journal={Communications in Computational Physics},
  volume={28},
  number={5},
  year={2020},
  publisher={Brown Univ., Providence, RI (United States)}
}

@article{tripura2023wavelet,
  title={Wavelet neural operator for solving parametric partial differential equations in computational mechanics problems},
  author={Tripura, Tapas and Chakraborty, Souvik},
  journal={Computer Methods in Applied Mechanics and Engineering},
  volume={404},
  pages={115783},
  year={2023},
  publisher={Elsevier}
}

@book{bartlett,
 author = {Anthony, Martin and Bartlett, Peter L.},
 title = {Neural Network Learning: Theoretical Foundations},
 year = {2009},
 isbn = {052111862X, 9780521118620},
 edition = {1st},
 publisher = {Cambridge University Press},
 address = {New York, NY, USA},
}

@article{aronszajn1950theory,
  title={Theory of reproducing kernels},
  author={Aronszajn, Nachman},
  journal={Transactions of the American mathematical society},
  volume={68},
  number={3},
  pages={337--404},
  year={1950}
}

@article{barron1993,
  title={Universal approximation bounds for superpositions of a sigmoidal function},
  author={Barron, A. R.},
  journal={Information Theory, IEEE Transactions on},
  volume={39},
  number={3},
  pages={930--945},
  year={1993},
  publisher={IEEE}
}

@article{belkinfound,
  title={Towards a theoretical foundation for {L}aplacian-based manifold methods},
  author={Belkin, M. and Niyogi, P.},
  journal={Journal of Computer and System Sciences},
  volume={74},
  number={8},
  pages={1289--1308},
  year={2008},
  publisher={Elsevier}
}

@book{bishop2006pattern,
  title={Pattern recognition and machine learning},
  author={Bishop, Christopher M and Nasrabadi, Nasser M},
  volume={4},
   year={2006},
  publisher={Springer}
}

@article{birkhoffpap,
  title={Minimum Sobolev norm interpolation of scattered derivative data},
  author={Chandrasekaran, Shivkumar and Gorman, CH and Mhaskar, Hrushikesh Narhar},
  journal={Journal of Computational Physics},
  volume={365},
  pages={149--172},
  year={2018},
  publisher={Elsevier}
}

@article{bdint,
  title={Minimum {S}obolev norm interpolation with trigonometric polynomials on the torus},
  author={Chandrasekaran, S. and Jayaraman, K. R. and Mhaskar, H. N.},
  journal={Journal of Computational Physics},
  volume={249},
  pages={96--112},
  year={2013},
  publisher={Elsevier}
}

@article{chen1995universal,
  title={Universal approximation to nonlinear operators by neural networks with arbitrary activation functions and its application to dynamical systems},
  author={Chen, Tianping and Chen, Hong},
  journal={IEEE Transactions on Neural Networks},
  volume={6},
  number={4},
  pages={911--917},
  year={1995},
  publisher={IEEE}
}

@article{achaspissue,
title={Special issue: Diffusion maps and wavelets},
author={Chui, C. K. and Donoho, D. L.},
journal={Appl. and Comput. Harm. Anal.},
volume={21},
number={1},
year={2006}
}

@article{chuili1992,
  title={Approximation by ridge functions and neural networks with one hidden layer},
  author={Chui, C. K. and Li, X.},
  journal={Journal of Approximation Theory},
  volume={70},
  number={2},
  pages={131--141},
  year={1992},
  publisher={Elsevier}
}

@article{chui1993realization,
  title={Realization of neural networks with one hidden layer},
  author={Chui, C. K. and Li, X.},
  journal={Multivariate Approximation: From CAGD to Wavelets, World Scientific, Singapore},
  pages={77--89},
  year={1993}
}

@article{chui1994neural,
  title={Neural networks for localized approximation},
  author={Chui, C. K. and Li, X. and Mhaskar, H. N.},
  journal={Mathematics of Computation},
  volume={63},
  number={208},
  pages={607--623},
  year={1994}
}

@article{chui1996limitations,
  title={Limitations of the approximation capabilities of neural networks with one hidden layer},
  author={Chui, C. K. and Li, X. and Mhaskar, H. N.},
  journal={Advances in Computational Mathematics},
  volume={5},
  number={1},
  pages={233--243},
  year={1996},
  publisher={Springer}
}

@ARTICLE{chui_deep,
  AUTHOR={Chui, Charles K. and Mhaskar, Hrushikesh N.},   
TITLE={Deep Nets for Local Manifold Learning},      
	JOURNAL={Frontiers in Applied Mathematics and Statistics},      
	VOLUME={4},      
PAGES={12},     	
YEAR={2018},       
URL={https://www.frontiersin.org/article/10.3389/fams.2018.00012},       	
DOI={10.3389/fams.2018.00012},      
ISSN={2297-4687}
}

@book{chung1997spectral,
  title={Spectral graph theory},
  author={Chung, F. R. K.},
  volume={92},
  year={1997},
  publisher={American Mathematical Soc.}
}

@article{cloninger2021deep,
  title={A deep network construction that adapts to intrinsic dimensionality beyond the domain},
  author={Cloninger, Alexander and Klock, Timo},
  journal={Neural Networks},
  volume={141},
  pages={404--419},
  year={2021},
  publisher={Elsevier}
}

@article{cloninger2020cautious,
  title={Cautious active clustering},
  author={Cloninger, Alexander and Mhaskar, Hrushikesh Narhar},
  journal={Applied and Computational Harmonic Analysis},
  volume={54},
  pages={44--74},
  year={2021},
  publisher={Elsevier}
}

@article{coifmanlafondiffusion,
  title={Diffusion maps},
  author={Coifman, R. R. and Lafon, S.},
  journal={Applied and computational harmonic analysis},
  volume={21},
  number={1},
  pages={5--30},
  year={2006},
  publisher={Elsevier}
}

@article{corominas1954condiciones,
  title={Condiciones para que una funcion infinitamente derivable sea un polinomio},
  author={Corominas, Ernesto and Balaguer, Ferran Sunyer},
  journal={Revista matem{\'a}tica hispanoamericana},
  volume={14},
  number={1},
  pages={26--43},
  year={1954},
  publisher={Real Sociedad Matem{\'a}tica Espa{\~n}ola}
}

@book{zhoubk_learning,
  title={Learning theory: an approximation theory viewpoint},
  author={Cucker, Felipe and Zhou, Ding Xuan},
  volume={24},
  year={2007},
  publisher={Cambridge University Press}
}

@article{cybenko1989,
  title={Approximation by superposition of sigmoidal functions},
  author={Cybenko, G.},
  journal={Mathematics of Control, Signals and Systems},
  volume={2},
  number={4},
  pages={303--314},
  year={1989}
}

@book{dai2013approximation,
  title={Approximation theory and harmonic analysis on spheres and balls},
  author={Dai, Feng and Xu, Yuan},
  volume={23},
  year={2013},
  publisher={Springer}
}

@article{devore1989optimal,
  title={Optimal nonlinear approximation},
  author={DeVore, R. A. and Howard, R. and Micchelli, C. A.},
  journal={Manuscripta mathematica},
  volume={63},
  number={4},
  pages={469--478},
  year={1989},
  publisher={Springer}
}

@book{devlorbk,
  title={Constructive approximation},
  author={DeVore, R. A. and Lorentz, G. G.},
  volume={303},
  year={1993},
  publisher={Springer Science \& Business Media}
}

@article{compbio,
  title={Locally Learning Biomedical Data Using Diffusion Frames},
  author={Ehler, M. and Filbir, F. and Mhaskar, H. N.},
  journal={Journal of Computational Biology},
  volume={19},
  number={11},
  pages={1251--1264},
  year={2012},
  publisher={Mary Ann Liebert, Inc. 140 Huguenot Street, 3rd Floor New Rochelle, NY 10801 USA}
}

@article{frankbern,
  title={A quadrature formula for diffusion polynomials corresponding to a generalized heat kernel},
  author={Filbir, F. and Mhaskar, H. N.},
  journal={Journal of Fourier Analysis and Applications},
  volume={16},
  number={5},
  pages={629--657},
  year={2010},
  publisher={Springer}
}

@article{modlpmz,
  title={Marcinkiewicz--{Z}ygmund measures on manifolds},
  author={Filbir, F. and Mhaskar, H. N.},
  journal={Journal of Complexity},
  volume={27},
  number={6},
  pages={568--596},
  year={2011},
  publisher={Elsevier}
}

@article{funahashi1989,
  title={On the approximate realization of continuous mappings by neural networks},
  author={Funahashi, K.-I.},
  journal={Neural networks},
  volume={2},
  number={3},
  pages={183--192},
  year={1989},
  publisher={Elsevier}
}

@article{geller2011band,
  title={Band-limited localized {P}arseval frames and {B}esov spaces on compact homogeneous manifolds},
  author={Geller, Daryl and Pesenson, Isaac Z},
  journal={Journal of Geometric Analysis},
  volume={21},
  number={2},
  pages={334--371},
  year={2011},
  publisher={Springer}
}

@article{geshkovski2023mathematical,
  title={A mathematical perspective on Transformers},
  author={Geshkovski, Borjan and Letrouit, Cyril and Polyanskiy, Yury and Rigollet, Philippe},
  journal={arXiv preprint arXiv:2312.10794},
  year={2023}
}

@article{girosi1990networks,
  title={Networks and the best approximation property},
  author={Girosi, F. and Poggio, T.},
  journal={Biological cybernetics},
  volume={63},
  number={3},
  pages={169--176},
  year={1990},
  publisher={Springer}
}

@book{goodfellow2016deep,
  title={Deep learning},
  author={Goodfellow, Ian and Bengio, Yoshua and Courville, Aaron and Bengio, Yoshua},
  volume={1},
  year={2016},
  publisher={MIT press Cambridge}
}

@article{pottssphquadconst,
  title={On the computation of nonnegative quadrature weights on the sphere},
  author={Gr{\"a}f, M. and Kunis, S. and Potts, D.},
  journal={Applied and Computational Harmonic Analysis},
  volume={27},
  number={1},
  pages={124--132},
  year={2009},
  publisher={Elsevier}
}

@article{golomb1959optimal,
  title={Optimal approximation and error bounds},
  author={Golomb, Michael and Weinberger, Hans F},
  journal={On numerical approximation},
  pages={117--190},
  year={1959}
}

@article{guliyev2016single,
  title={A single hidden layer feedforward network with only one neuron in the hidden layer can approximate any univariate function},
  author={Guliyev, Namig J and Ismailov, Vugar E},
  journal={Neural computation},
  volume={28},
  number={7},
  pages={1289--1304},
  year={2016},
  publisher={MIT Press}
}

@book{horn_johnson_book,
  title={Matrix analysis},
  author={Horn, R. A. and Johnson, C. R.},
  year={2012},
  publisher={Cambridge university press}
}

@article{hornik1989,
  title={Multilayer feedforward networks are universal approximators},
  author={Hornik, K. and Stinchcombe, M. and White, H.},
  journal={Neural networks},
  volume={2},
  number={5},
  pages={359--366},
  year={1989},
  publisher={Elsevier}
}

@inproceedings{irie1988,
  title={Capabilities of three-layered perceptrons},
  author={Irie, B. and Miyake, S.},
  booktitle={Neural Networks, 1988., IEEE International Conference on},
  pages={641--648},
  year={1988},
  organization={IEEE}
}

@article{jones2010universal,
  title={Universal local parametrizations via heat kernels and eigenfunctions of the {L}aplacian},
  author={Jones, P. W. and Maggioni, M. and Schul, R.},
  journal={Ann. Acad. Sci. Fenn. Math.},
volume={35},
pages={131--174},
  year={2010},
  publisher={Citeseer}
}

@article{keiner2009using,
  title={Using NFFT 3---a software library for various nonequispaced fast Fourier transforms},
  author={Keiner, Jens and Kunis, Stefan and Potts, Daniel},
  journal={ACM Transactions on Mathematical Software (TOMS)},
  volume={36},
  number={4},
  pages={19},
  year={2009},
  publisher={ACM}
}

@article{kim2023theoretical,
  title={Theoretical bounds of generalization error for generalized extreme learning machine and random vector functional link network},
  author={Kim, Meejoung},
  journal={Neural Networks},
  volume={164},
  pages={49--66},
  year={2023},
  publisher={Elsevier}
}

@ARTICLE{Barron2018,
  author={Klusowski, Jason M. and Barron, Andrew R.},
  journal={IEEE Transactions on Information Theory}, 
  title={Approximation by Combinations of ReLU and Squared ReLU Ridge Functions With  $\ell^1$  and  $\ell^0$  Controls}, 
  year={2018},
  volume={64},
  number={12},
  pages={7649-7656},
  doi={10.1109/TIT.2018.2874447}}

@article{kovachki2021universal,
  title={On universal approximation and error bounds for Fourier Neural Operators},
  author={Kovachki, Nikola and Lanthaler, Samuel and Mishra, Siddhartha},
  journal={Journal of Machine Learning Research},
  volume={22},
  pages={Art--No},
  year={2021},
  publisher={JMLR Press}
}

@article{kurkova1,
  title={Bounds on rates of variable basis and neural network approximation},
  author={K\r{u}rkov{\'a}, V. and Sanguineti, M.},
  journal={IEEE Transactions on Information Theory},
  volume={47},
  number={6},
  pages={2659--2665},
  year={2001}
}

@article{kurkova2,
  title={Comparison of worst case errors in linear and neural network approximation},
  author={K\r{u}rkov{\'a}, V. and Sanguineti, M.},
  journal={IEEE Transactions on Information Theory},
  volume={48},
  number={1},
  pages={264--275},
  year={2002}
}

@phdthesis{lafon,
  title={Diffusion maps and geometric harmonics},
  author={Lafon, S. S.},
  year={2004},
  school={Yale University},
  address={Yale}
}

@article{lai2021kolmogorov,
  title={The kolmogorov superposition theorem can break the curse of dimensionality when approximating high dimensional functions},
  author={Lai, Ming-Jun and Shen, Zhaiming},
  journal={arXiv preprint arXiv:2112.09963},
  year={2021}
}

@article{lai2024optimal,
  title={The optimal rate for linear KB-splines and LKB-splines approximation of high dimensional continuous functions and its application},
  author={Lai, Ming-Jun and Shen, Zhaiming},
  journal={arXiv preprint arXiv:2401.03956},
  year={2024}
}

@article{quadconst,
  title={Localized linear polynomial operators and quadrature formulas on the sphere},
  author={Le Gia, Q. T. and Mhaskar, H. N.},
  journal={SIAM Journal on Numerical Analysis},
  volume={47},
  number={1},
  pages={440--466},
  year={2009},
  publisher={SIAM}
}

@article{leshnolinpinkus,
  title={Multilayer feedforward networks with a nonpolynomial activation function can approximate any function},
  author={Leshno, M. and Lin, V. Ya. and Pinkus, A. and Schocken, S.},
  journal={Neural networks},
  volume={6},
  number={6},
  pages={861--867},
  year={1993},
  publisher={Elsevier}
}

@article{liu2024kan,
  title={Kan: Kolmogorov-arnold networks},
  author={Liu, Ziming and Wang, Yixuan and Vaidya, Sachin and Ruehle, Fabian and Halverson, James and Solja{\v{c}}i{\'c}, Marin and Hou, Thomas Y and Tegmark, Max},
  journal={arXiv preprint arXiv:2404.19756},
  year={2024}
}

@article{li2020fourier,
  title={Fourier neural operator for parametric partial differential equations},
  author={Li, Zongyi and Kovachki, Nikola and Azizzadenesheli, Kamyar and Liu, Burigede and Bhattacharya, Kaushik and Stuart, Andrew and Anandkumar, Anima},
  journal={arXiv preprint arXiv:2010.08895},
  year={2020}
}

@book{lorentz2005approximation,
  title={Approximation of functions},
  author={Lorentz, George G},
  volume={322},
  year={2005},
  publisher={American Mathematical Soc.}
}

@book{lorentz_advanced,
  title={Constructive approximation: advanced problems},
  author={Lorentz, G. G. and von Golitschek, M. and Makovoz, Y.},
  volume={304},
  year={1996},
  publisher={Springer Berlin}
}

@article{ma2022uniform,
  title={Uniform approximation rates and metric entropy of shallow neural networks},
  author={Ma, Limin and Siegel, Jonathan W and Xu, Jinchao},
  journal={Research in the Mathematical Sciences},
  volume={9},
  number={3},
  pages={46},
  year={2022},
  publisher={Springer}
}

@article{mauropap,
  title={Diffusion polynomial frames on metric measure spaces},
  author={Maggioni, M. and Mhaskar, H. N.},
  journal={Applied and Computational Harmonic Analysis},
  volume={24},
  number={3},
  pages={329--353},
  year={2008},
  publisher={Elsevier}
}

@article{makovoz1998uniform,
  title={Uniform approximation by neural networks},
  author={Makovoz, Yuly},
  journal={Journal of Approximation Theory},
  volume={95},
  number={2},
  pages={215--228},
  year={1998},
  publisher={Elsevier}
}

@article{mason2021manifold,
  title={A manifold learning approach for gesture identification from micro-Doppler radar measurements},
  author={Mason, Eric and Mhaskar, Hrushikesh Narhar and Guo, Adam},
journal={  Neural Networks},
volume={ 152},
year={2022},
pages={ 353--369},
note={arXiv preprint arXiv:2110.01670, 2021}
}

@article{mhaskar2024tractability,
  title={Tractability of approximation by general shallow networks},
  author={Mhaskar, Hrushikesh N and Mao, Tong},
  journal={Analysis and Applications},
  volume={22},
  number={03},
  pages={535--568},
  year={2024},
  publisher={World Scientific}
}

@article{mhaskar2025approximation,
  title={Approximation by non-symmetric networks for cross-domain learning},
  author={Mhaskar, Hrushikesh Narhar},
  journal={Neural Networks},
  vol={187},
  year={2025},
  pages={107282}
}

@article{mhaskar2023ryan,
  title={Local transfer learning from one data space to another},
  author={Mhaskar, Hrushikesh Narhar and O'Dowd, Ryan},
  journal={arXiv preprint arXiv:2302.00160},
  year={2023}
}

@article{mhaskar2024learning,
  title={Learning on manifolds without manifold learning},
  author={Mhaskar, Hrushikesh N and O’Dowd, Ryan},
  journal={Neural Networks},
  volume={181},
  pages={106759},
  year={2025},
  publisher={Elsevier}
}

@misc{Mhaskar2025Sep, 
title={Signal separation approach for classification},
    author ={Mhaskar, Hrushikesh N and  O'Dowd, Ryan },
    note ={In preparation},
}

@article{multilayer,
  title={Approximation properties of a multilayered feedforward artificial neural network},
  author={Mhaskar, Hrushikesh Narhar},
  journal={Advances in Computational Mathematics},
  volume={1},
  number={1},
  pages={61--80},
  year={1993},
  publisher={Springer}
}

@article{sphrelu,
  title={Function approximation with zonal function networks with activation functions analogous to the rectified linear unit functions},
  author={Mhaskar, Hrushikesh N},
  journal={Journal of Complexity}, 
  volume={51}, 
  year={April 2019},
   pages={ 1-19},
  }

@article{locsmooth,
  title={On the representation of smooth functions on the sphere using finitely many bits},
  author={Mhaskar, Hrushikesh Narhar},
  journal={Applied and Computational Harmonic Analysis},
  volume={18},
  number={3},
  pages={215--233},
  year={2005},
  publisher={Elsevier}
}

@article{eignet,
  title={Eignets for function approximation on manifolds},
  author={Mhaskar, Hrushikesh Narhar},
  journal={Applied and Computational Harmonic Analysis},
  volume={29},
  number={1},
  pages={63--87},
  year={2010},
  publisher={Elsevier}
}

@article{tractable,
  title={On the tractability of multivariate integration and approximation by neural networks},
  author={Mhaskar, Hrushikesh Narhar},
  journal={Journal of Complexity},
  volume={20},
  number={4},
  pages={561--590},
  year={2004},
  publisher={Elsevier}
}

@article{convtheo,
  title={When is approximation by {G}aussian networks necessarily a linear process?},
  author={Mhaskar, Hrushikesh Narhar},
  journal={Neural Networks},
  volume={17},
  number={7},
  pages={989--1001},
  year={2004},
  publisher={Elsevier}
}

@inproceedings{indiapap,
  title={Approximation theory and neural networks},
  author={Mhaskar, Hrushikesh Narhar},
  booktitle={Wavelet Analysis and Applications, Proceedings of the international workshop in Delhi},
  pages={247--289},
  year={1999}
}

@article{heatkernframe,
  title={A generalized diffusion frame for parsimonious representation of functions on data defined manifolds},
  author={Mhaskar, Hrushikesh Narhar},
  journal={Neural Networks},
  volume={24},
  number={4},
  pages={345--359},
  year={2011},
  publisher={Elsevier}
}

@article{tauberian,
author={Mhaskar, Hrushikesh Narhar},
title={A unified framework for  harmonic analysis of functions on directed graphs and changing data},
journal={ Appl. Comput. Harm. Anal. },
volume={ 44}, 
number={3}, 
year={2018}, 
pages={ 611-644}
}

@article{optneur,
  title={Neural networks for optimal approximation of smooth and analytic functions},
  author={Mhaskar, Hrushikesh Narhar},
  journal={Neural Computation},
  volume={8},
  number={1},
  pages={164--177},
  year={1996},
  publisher={MIT Press}
}

@incollection{sloanfest,
title={Approximate quadrature measures on data--defined spaces},
  author={Mhaskar, Hrushikesh N},
  booktitle={Festschrift for the 80th Birthday of Ian Sloan},
 year={2017},
 editor={Dick, Josef and Kuo, Frances Y and Wozniakowski, Henryk },
 publisher={Springer},
 pages={931-962},
 note={arXiv preprint arXiv:1612.02368}
 }

@article{mhaskar2019deep,
  title={A direct method for function approximation on data defined manifolds},
  author={Mhaskar, Hrushikesh Narhar},
  note={ArXiv preprint arXiv:1908.00156, doi:10.1016/j.neunet.2020.08.018},
  journal={Neural Networks},
  volume={132},
  pages={253-268},
  year={2020},
}

@article{mhaskar2019dimension,
  title={Dimension independent bounds for general shallow networks},
  author={Mhaskar, Hrushikesh N},
  journal={arXiv preprint arXiv:1908.09880},
  year={2019}
}

@article{mhaskar2020dimension,
  title={Dimension independent bounds for general shallow networks},
  author={Mhaskar, Hrushikesh N},
  journal={Neural Networks},
  volume={123},
  pages={142--152},
  year={2020},
  publisher={Elsevier}
}

@article{mhaskar2020kernel,
 AUTHOR={Mhaskar, Hrushikesh N.},   
TITLE={Kernel-Based Analysis of Massive Data},      
JOURNAL={Frontiers in Applied Mathematics and Statistics},      
VOLUME={6},      
PAGES={30},     
YEAR={2020},      
URL={https://www.frontiersin.org/article/10.3389/fams.2020.00030},       
DOI={10.3389/fams.2020.00030},      
ISSN={2297-4687}
}

@article{mhaskar2022local,
  title={Local approximation of operators},
  author={Mhaskar, Hrushikesh Narhar},
  journal={arXiv preprint arXiv:2202.06392},
  year={2022}
}

@article{mhaskar2023local,
  title={Local approximation of operators},
  author={Mhaskar, Hrushikesh N},
  journal={Applied and Computational Harmonic Analysis},
  year={2023},
  publisher={Elsevier}
}

@article{mhasmich,
  title={Approximation by superposition of sigmoidal and radial basis functions},
  author={Mhaskar, H. N. and Micchelli, C. A.},
  journal={Advances in Applied mathematics},
  volume={13},
  number={3},
  pages={350--373},
  year={1992},
  publisher={Elsevier}
}

@article{mhaskar1995degree,
  title={Degree of approximation by neural and translation networks with a single hidden layer},
  author={Mhaskar, H. N. and Micchelli, C. A.},
  journal={Advances in Applied Mathematics},
  volume={16},
  number={2},
  pages={151--183},
  year={1995},
  publisher={Elsevier}
}

@article{dimindbd,
  title={Dimension-independent bounds on the degree of approximation by neural networks},
  author={Mhaskar, H. N. and Micchelli, C. A.},
  journal={IBM Journal of Research and Development},
  volume={38},
  number={3},
  pages={277--284},
  year={1994},
  publisher={IBM}
}

@article{eugenenevai,
  title={Applications of classical approximation theory to periodic basis function networks and computational harmonic analysis},
  author={Mhaskar, H. N. and Nevai, P. and Shvarts, E.},
  journal={Bulletin of Mathematical Sciences},
  volume={3},
  number={3},
  pages={485--549},
  year={2013},
  publisher={Springer}
}

@article{mnw1,
  title={Spherical {M}arcinkiewicz-{Z}ygmund inequalities and positive quadrature},
  author={Mhaskar, H. N. and Narcowich, F. J. and Ward, J. D.},
  journal={Mathematics of computation},
  volume={70},
  number={235},
  pages={1113--1130},
  year={2001}
}

@article{approxint2002,
  title={Approximation with interpolatory constraints},
  author={Mhaskar, H. N. and Narcowich, F. J. and Sivakumar, N. and Ward, J. D.},
  journal={Proceedings of the American Mathematical Society},
  volume={130},
  number={5},
  pages={1355--1364},
  year={2002}
}

@ARTICLE{mhas_sergei_maryke_diabetes2017,
AUTHOR={Mhaskar, H. N. and Pereverzyev, Sergei V. and van der Walt, Maria D.},   
TITLE={A Deep Learning Approach to Diabetic Blood Glucose Prediction},      
JOURNAL={Frontiers in Applied Mathematics and Statistics},      
VOLUME={3},      
PAGES={14},     
YEAR={2017},      
URL={http://journal.frontiersin.org/article/10.3389/fams.2017.00014},       
DOI={10.3389/fams.2017.00014},      
ISSN={2297-4687}
}

@misc{matrixpap,
title={Data based construction of kernels for classification},
author={Mhaskar, H. N. and Pereverzyev, Sergei V. and Semenov, Vasyl Yu. and Semenova, Evgeniya V.},
howpublished={Submitted for publication, https://www.ricam.oeaw.ac.at/files/reports/18/rep18-25.pdf},
year={2018}
}

@article{dingxuanpap,
  title={Deep vs. shallow networks: An approximation theory perspective},
  author={Mhaskar, H. N. and Poggio, Tomaso},
  journal={Analysis and Applications},
  volume={14},
  number={06},
  pages={829--848},
  year={2016},
  publisher={World Scientific}
}

@article{mhaskar2018analysis,
crossref={mhaskar2019analysis},
  title={An analysis of training and generalization errors in shallow and deep networks},
  author={Mhaskar, H. N. and Poggio, Tomaso},
journal={arXiv preprint arXiv:1802.06266},
  year={2018}
}

@article{mhaskar2019analysis,
  title={An analysis of training and generalization errors in shallow and deep networks},
  author={Mhaskar, Hrushikesh N and Poggio, Tomaso},
  journal={Neural Networks},
  year={2020},
  publisher={Elsevier},
  volume={121},
  pages={229--241}
}

@article{poggio2016_cbmm58,
  title={Why and when can deep-but not shallow-networks avoid the curse of dimensionality: A review},
  author={Poggio, Tomaso and Mhaskar, H. N. and Rosasco, Lorenzo and Miranda, Brando and Liao, Qianli},
  journal={International Journal of Automation and Computing},
  pages={1--17},
  year={2017},
  publisher={Springer}
}

@article{trigwave,
  title={On the detection of singularities of a periodic function},
  author={Mhaskar, H. N. and Prestin, J.},
  journal={Advances in Computational Mathematics},
  volume={12},
  number={2-3},
  pages={95--131},
  year={2000},
  publisher={Springer}
}

@article{loctrigwave,
  title={On local smoothness classes of periodic functions},
  author={Mhaskar, H. N. and Prestin, J.},
  journal={Journal of Fourier Analysis and Applications},
  volume={11},
  number={3},
  pages={353--373},
  year={2005},
  publisher={Springer}
}

@article{mhaskar1997smooth,
  title={On smooth activation functions},
  author={Mhaskar, H. N.},
  journal={Mathematics of Neural Networks: Models, Algorithms and Applications},
  pages={275--279},
  year={1997},
  publisher={Springer}
}

@book{michel2012lectures,
  title={Lectures on constructive approximation: Fourier, spline, and wavelet methods on the real line, the sphere, and the ball},
  author={Michel, Volker},
  year={2012},
  publisher={Springer Science \& Business Media}
}

@book{murphy2022probabilistic,
  title={Probabilistic machine learning: an introduction},
  author={Murphy, Kevin P},
  year={2022},
  publisher={MIT press}
}

@article{petersen2018optimal,
  title={Optimal approximation of piecewise smooth functions using deep ReLU neural networks},
  author={Petersen, Philipp and Voigtlaender, Felix},
  journal={Neural Networks},
  volume={108},
  pages={296--330},
  year={2018},
  publisher={Elsevier}
}

@book{pinkusbk,
  title={$N$-widths in Approximation Theory},
  author={Pinkus, Allan},
  volume={7},
  year={2012},
  publisher={Springer Science \& Business Media}
}

@article{pinkus1999approximation,
  title={Approximation theory of the MLP model in neural networks},
  author={Pinkus, Allan},
  journal={Acta Numerica},
  volume={8},
  pages={143--195},
  year={1999},
  publisher={Cambridge University Press}
}

@article{potts1998fast,
  title={Fast algorithms for discrete polynomial transforms},
  author={Potts, Daniel and Steidl, Gabriele and Tasche, Manfred},
  journal={Mathematics of Computation},
  volume={67},
  number={224},
  pages={1577--1590},
  year={1998}
}

@article{potts2003fast,
  title={Fast algorithms for discrete polynomial transforms on arbitrary grids},
  author={Potts, Daniel},
  journal={Linear algebra and its applications},
  volume={366},
  pages={353--370},
  year={2003},
  publisher={Elsevier}
}

@article{schmidt2021kolmogorov,
  title={The Kolmogorov--Arnold representation theorem revisited},
  author={Schmidt-Hieber, Johannes},
  journal={Neural networks},
  volume={137},
  pages={119--126},
  year={2021},
  publisher={Elsevier}
}

@article{schmidt2020nonparametric,
  title={Nonparametric regression using deep neural networks with {ReLU} activation function},
  author={Schmidt-Hieber, Johannes},
   journal={The Annals of Statistics},
  volume={48},
  number={4},
  pages={1875--1897},
  year={2020}
}

@article{schmidt2019deep,
  title={Deep {ReLU} network approximation of functions on a manifold},
  author={Schmidt-Hieber, Johannes},
  journal={arXiv preprint arXiv:1908.00695},
  year={2019}
}

@article{siegel2025optimal,
  title={Optimal approximation of zonoids and uniform approximation by shallow neural networks},
  author={Siegel, Jonathan W},
  journal={Constructive Approximation},
  pages={1--29},
  year={2025},
  publisher={Springer}
}

@book{stein2016singular,
  title={Singular integrals and differentiability properties of functions (PMS-30)},
  author={Stein, Elias M},
  volume={30},
  year={2016},
  publisher={Princeton university press}
}

@article{siegel2022sharp,
  title={Sharp bounds on the approximation rates, metric entropy, and n-widths of shallow neural networks},
  author={Siegel, Jonathan W and Xu, Jinchao},
  journal={Foundations of Computational Mathematics},
  pages={1--57},
  year={2022},
  publisher={Springer}
}

@article{szabados1978some,
  title={On some convergent interpolatory polynomials},
  author={Szabados, J},
  journal={Fourier Analysis and Approximation Theory, Coll. Math. Soc. J{\'a}nos Bolyai},
  volume={19},
  pages={805--815},
  year={1978}
}

@book{timanbk,
  title={Theory of Approximation of Functions of a Real Variable: International Series of Monographs on Pure and Applied Mathematics},
  author={Timan, Aleksandr Filippovich},
  volume={34},
  year={2014},
  publisher={Elsevier}
}

@book{vapnik2013nature,
  title={The nature of statistical learning theory},
  author={Vapnik, V.},
  year={2013},
  publisher={Springer Science \& Business Media, New York}
}

@article{varadhan1967behavior,
  title={On the behavior of the fundamental solution of the heat equation with variable coefficients},
  author={Varadhan, Sathamangalam R Srinivasa},
  journal={Communications on Pure and Applied Mathematics},
  volume={20},
  number={2},
  pages={431--455},
  year={1967},
  publisher={Wiley Online Library}
}

@book{watanabe2009algebraic,
  title={Algebraic geometry and statistical learning theory},
  author={Watanabe, Sumio},
  volume={25},
  year={2009},
  publisher={Cambridge university press}
}

@article{xu2020finite,
  title={The finite neuron method and convergence analysis},
  author={Xu, Jinchao},
  journal={arXiv preprint arXiv:2010.01458},
  year={2020}
}

@article{yarotsky2016error,
crossref={yarotsky2017error},
  title={Error bounds for approximations with deep {ReLU} networks},
  author={Yarotsky, Dmitry},
  journal={arXiv preprint arXiv:1610.01145},
  year={2016}
}

@article{yarotsky2017error,
  title={Error bounds for approximations with deep {ReLU} networks},
  author={Yarotsky, Dmitry},
  journal={Neural Networks},
  volume={94},
  pages={103--114},
  year={2017},
  publisher={Elsevier}
}

@article{zhou2020universality,
  title={Universality of deep convolutional neural networks},
  author={Zhou, Ding-Xuan},
  journal={Applied and computational harmonic analysis},
  volume={48},
  number={2},
  pages={787--794},
  year={2020},
  publisher={Elsevier}
}

@article{doi:10.1137/20M133607,
author = {Belkin, Mikhail and Hsu, Daniel and Xu, Ji},
title = {Two Models of Double Descent for Weak Features},
journal = {SIAM Journal on Mathematics of Data Science},
volume = {2},
number = {4},
pages = {1167-1180},
year = {2020},
doi = {10.1137/20M1336072},
URL = {https://doi.org/10.1137/20M1336072},
}

@article{belkin2019reconciling,
  title={Reconciling modern machine-learning practice and the classical bias--variance trade-off},
  author={Belkin, Mikhail and Hsu, Daniel and Ma, Siyuan and Mandal, Soumik},
  journal={Proceedings of the National Academy of Sciences},
  volume={116},
  number={32},
  pages={15849--15854},
  year={2019},
  publisher={National Acad Sciences}
}

@article{mukherjee2006learning,
  title={Learning theory: stability is sufficient for generalization and necessary and sufficient for consistency of empirical risk minimization},
  author={Mukherjee, Sayan and Niyogi, Partha and Poggio, Tomaso and Rifkin, Ryan},
  journal={Advances in Computational Mathematics},
  volume={25},
  pages={161--193},
  year={2006},
  publisher={Springer}
}

@article{devore1997nonlinear,
  title={Nonlinear approximation in finite-dimensional spaces},
  author={DeVore, Ronald A and Temlyakov, Vladimir N},
  journal={Journal of Complexity},
  volume={13},
  number={4},
  pages={489--508},
  year={1997},
  publisher={Citeseer}
}

@book{trigub2012fourier,
  title={Fourier analysis and approximation of functions},
  author={Trigub, Roald M and Belinsky, Eduard S},
  year={2012},
  publisher={Springer Science \& Business Media}
}

@book{Singer1970BestApproximation,
  title={Best Approximation in Normed Linear Spaces by Elements of Linear Subspaces},
  author={Ivan Singer},
  year={1970},
  publisher={Springer}
}

@article{mao2023approximating,
  title={Approximating functions with multi-features by deep convolutional neural networks},
  author={Mao, Tong and Shi, Zhongjie and Zhou, Ding-Xuan},
  journal={Analysis and Applications},
  volume={21},
  number={01},
  pages={93--125},
  year={2023},
  publisher={World Scientific}
}

@article{bolcskei2019optimal,
  title={Optimal approximation with sparsely connected deep neural networks},
  author={Bolcskei, Helmut and Grohs, Philipp and Kutyniok, Gitta and Petersen, Philipp},
  journal={SIAM Journal on Mathematics of Data Science},
  volume={1},
  number={1},
  pages={8--45},
  year={2019},
  publisher={SIAM}
}

@article{singer2006graph,
  title={From graph to manifold Laplacian: The convergence rate},
  author={Singer, Amit},
  journal={Applied and Computational Harmonic Analysis},
  volume={21},
  number={1},
  pages={128--134},
  year={2006},
  publisher={Elsevier}
}

@article{von2008consistency,
  title={Consistency of spectral clustering},
  author={Von Luxburg, Ulrike and Belkin, Mikhail and Bousquet, Olivier},
  journal={The Annals of Statistics},
  pages={555--586},
  year={2008},
  publisher={JSTOR}
}

@inproceedings{hein2005intrinsic,
  title={Intrinsic dimensionality estimation of submanifolds in Rd},
  author={Hein, Matthias and Audibert, Jean-Yves},
  booktitle={Proceedings of the 22nd international conference on Machine learning},
  pages={289--296},
  year={2005}
}

@book{micchelli1984interpolation,
  title={Interpolation of scattered data: distance matrices and conditionally positive definite functions},
  author={Micchelli, Charles A},
  year={1984},
  publisher={Springer}
}

@article{park1991universal,
  title={Universal approximation using radial-basis-function networks},
  author={Park, Jooyoung and Sandberg, Irwin W},
  journal={Neural computation},
  volume={3},
  number={2},
  pages={246--257},
  year={1991},
  publisher={MIT Press}
}

@article{belkin2021fit,
  title={Fit without fear: remarkable mathematical phenomena of deep learning through the prism of interpolation},
  author={Belkin, Mikhail},
  journal={Acta Numerica},
  volume={30},
  pages={203--248},
  year={2021},
  publisher={Cambridge University Press}
}

@article{chen1993approximations,
  title={Approximations of continuous functionals by neural networks with application to dynamic systems},
  author={Chen, Tianping and Chen, Hong},
  journal={IEEE Transactions on Neural networks},
  volume={4},
  number={6},
  pages={910--918},
  year={1993},
  publisher={IEEE}
}

@article{jin2020quantifying,
  title={Quantifying the generalization error in deep learning in terms of data distribution and neural network smoothness},
  author={Jin, Pengzhan and Lu, Lu and Tang, Yifa and Karniadakis, George Em},
  journal={Neural Networks},
  volume={130},
  pages={85--99},
  year={2020},
  publisher={Elsevier}
}

@article{di2023neural,
  title={Neural operator prediction of linear instability waves in high-speed boundary layers},
  author={Di Leoni, Patricio Clark and Lu, Lu and Meneveau, Charles and Karniadakis, George Em and Zaki, Tamer A},
  journal={Journal of Computational Physics},
  volume={474},
  pages={111793},
  year={2023},
  publisher={Elsevier}
}

@article{xu2024overview,
  title={Overview frequency principle/spectral bias in deep learning},
  author={Xu, Zhi-Qin John and Zhang, Yaoyu and Luo, Tao},
  journal={Communications on Applied Mathematics and Computation},
  pages={1--38},
  year={2024},
  publisher={Springer}
}

@inproceedings{rahaman2019spectral,
  title={On the spectral bias of neural networks},
  author={Rahaman, Nasim and Baratin, Aristide and Arpit, Devansh and Draxler, Felix and Lin, Min and Hamprecht, Fred and Bengio, Yoshua and Courville, Aaron},
  booktitle={International conference on machine learning},
  pages={5301--5310},
  year={2019},
  organization={PMLR}
}

@article{fridovich2022spectral,
  title={Spectral bias in practice: The role of function frequency in generalization},
  author={Fridovich-Keil, Sara and Gontijo Lopes, Raphael and Roelofs, Rebecca},
  journal={Advances in Neural Information Processing Systems},
  volume={35},
  pages={7368--7382},
  year={2022}
}

@article{fang2024addressing,
  title={Addressing Spectral Bias of Deep Neural Networks by Multi-Grade Deep Learning},
  author={Fang, Ronglong and Xu, Yuesheng},
  journal={arXiv preprint arXiv:2410.16105},
  year={2024}
}

@article{cao2019towards,
  title={Towards understanding the spectral bias of deep learning},
  author={Cao, Yuan and Fang, Zhiying and Wu, Yue and Zhou, Ding-Xuan and Gu, Quanquan},
  journal={arXiv preprint arXiv:1912.01198},
  year={2019}
}

@article{filbir2008polynomial,
  title={Polynomial approximation on the sphere using scattered data},
  author={Filbir, Frank and Themistoclakis, Woula},
  journal={Mathematische Nachrichten},
  volume={281},
  number={5},
  pages={650--668},
  year={2008},
  publisher={Wiley Online Library}
}

@book{rivlin2020chebyshev,
  title={Chebyshev polynomials},
  author={Rivlin, Theodore J},
  year={2020},
  publisher={Courier Dover Publications}
}

@article{adcock2020deep,
  title={Deep neural networks are effective at learning high-dimensional Hilbert-valued functions from limited data},
  author={Adcock, Ben and Brugiapaglia, Simone and Dexter, Nick and Moraga, Sebastian},
  journal={arXiv preprint arXiv:2012.06081},
  year={2020}
}

@inproceedings{adcock2021learning,
  title={Learning High-Dimensional Hilbert-Valued Functions With Deep Neural Networks From Limited Data.},
  author={Adcock, Ben and Brugiapaglia, Simone and Dexter, Nick C and Moraga, Sebastian},
  booktitle={AAAI Spring Symposium: MLPS},
  year={2021}
}

@article{sule2023sharp,
  title={Sharp error estimates for target measure diffusion maps with applications to the committor problem},
  author={Sule, Shashank and Evans, Luke and Cameron, Maria},
  journal={arXiv preprint arXiv:2312.14418},
  year={2023}
}

@article{lanthaler2022error,
  title={Error estimates for deeponets: A deep learning framework in infinite dimensions},
  author={Lanthaler, Samuel and Mishra, Siddhartha and Karniadakis, George E},
  journal={Transactions of Mathematics and Its Applications},
  volume={6},
  number={1},
  pages={tnac001},
  year={2022},
  publisher={Oxford University Press}
}

@inproceedings{irion2015applied,
  title={Applied and computational harmonic analysis on graphs and networks},
  author={Irion, Jeff and Saito, Naoki},
  booktitle={Wavelets and Sparsity XVI},
  volume={9597},
  pages={336--350},
  year={2015},
  organization={SPIE}
}

@article{radhakrishnan2024linear,
  title={Linear Recursive Feature Machines provably recover low-rank matrices},
  author={Radhakrishnan, Adityanarayanan and Belkin, Mikhail and Drusvyatskiy, Dmitriy},
  journal={arXiv preprint arXiv:2401.04553},
  year={2024}
}

@article{siegel2024sharp,
  title={Sharp lower bounds on the manifold Widths of Sobolev and Besov spaces},
  author={Siegel, Jonathan W},
  journal={Journal of Complexity},
  pages={101884},
  year={2024},
  publisher={Elsevier}
}

@article{perekrestenko2018universal,
  title={The universal approximation power of finite-width deep ReLU networks},
  author={Perekrestenko, Dmytro and Grohs, Philipp and Elbr{\"a}chter, Dennis and B{\"o}lcskei, Helmut},
  journal={arXiv preprint arXiv:1806.01528},
  year={2018}
}

@article{radhakrishnan2024mechanism,
  title={Mechanism for feature learning in neural networks and backpropagation-free machine learning models},
  author={Radhakrishnan, Adityanarayanan and Beaglehole, Daniel and Pandit, Parthe and Belkin, Mikhail},
  journal={Science},
  volume={383},
  number={6690},
  pages={1461--1467},
  year={2024},
  publisher={American Association for the Advancement of Science}
}

@article{burkholz2021existence,
  title={On the existence of universal lottery tickets},
  author={Burkholz, Rebekka and Laha, Nilanjana and Mukherjee, Rajarshi and Gotovos, Alkis},
  journal={arXiv preprint arXiv:2111.11146},
  year={2021}
}

@article{shaham2018provable,
  title={Provable approximation properties for deep neural networks},
  author={Shaham, Uri and Cloninger, Alexander and Coifman, Ronald R},
  journal={Applied and Computational Harmonic Analysis},
  volume={44},
  number={3},
  pages={537--557},
  year={2018},
  publisher={Elsevier}
}

@article{fefferman2016testing,
  title={Testing the manifold hypothesis},
  author={Fefferman, Charles and Mitter, Sanjoy and Narayanan, Hariharan},
  journal={Journal of the American Mathematical Society},
  volume={29},
  number={4},
  pages={983--1049},
  year={2016}
}

@article{petrova2023lipschitz,
  title={Lipschitz widths},
  author={Petrova, Guergana and Wojtaszczyk, Przemys${\l}$aw},
  journal={Constructive Approximation},
  volume={57},
  number={2},
  pages={759--805},
  year={2023},
  publisher={Springer}
}

@article{mao2021theory,
  title={Theory of deep convolutional neural networks III: Approximating radial functions},
  author={Mao, Tong and Shi, Zhongjie and Zhou, Ding-Xuan},
  journal={Neural Networks},
  volume={144},
  pages={778--790},
  year={2021},
  publisher={Elsevier}
}

@article{fefferman2023fitting,
  title={Fitting a manifold of large reach to noisy data},
  author={Fefferman, Charles and Ivanov, Sergei and Lassas, Matti and Narayanan, Hariharan},
  journal={Journal of Topology and Analysis},
  pages={1--82},
  year={2023},
  publisher={World Scientific}
}

@article{cohen2022optimal,
  title={Optimal stable nonlinear approximation},
  author={Cohen, Albert and DeVore, Ronald and Petrova, Guergana and Wojtaszczyk, Przemyslaw},
  journal={Foundations of Computational Mathematics},
  volume={22},
  number={3},
  pages={607--648},
  year={2022},
  publisher={Springer}
}

@article{chen2023deep,
  title={Deep operator learning lessens the curse of dimensionality for PDEs},
  author={Chen, Ke and Wang, Chunmei and Yang, Haizhao},
  journal={arXiv preprint arXiv:2301.12227},
  year={2023}
}

@article{liu2024deep,
  title={Deep nonparametric estimation of operators between infinite dimensional spaces},
  author={Liu, Hao and Yang, Haizhao and Chen, Minshuo and Zhao, Tuo and Liao, Wenjing},
  journal={Journal of Machine Learning Research},
  volume={25},
  number={24},
  pages={1--67},
  year={2024}
}

@article{song2023approximation,
  title={Approximation of smooth functionals using deep ReLU networks},
  author={Song, Linhao and Liu, Ying and Fan, Jun and Zhou, Ding-Xuan},
  journal={Neural Networks},
  volume={166},
  pages={424--436},
  year={2023},
  publisher={Elsevier}
}

@article{mhaskar1997neural,
  title={Neural networks for functional approximation and system
identification},
  author={Mhaskar, Hrushikesh Narhar and Hahm, Nahmwoo},
  journal={Neural Computation},
  volume={9},
  number={1},
  pages={143--159},
  year={1997},
  publisher={MIT Press One Rogers Street, Cambridge, MA 02142-1209,
USA journals-info~…}
}

@article{lafon2006diffusion,
  title={Diffusion maps and coarse-graining: A unified framework for dimensionality reduction, graph partitioning, and data set parameterization},
  author={Lafon, Stephane and Lee, Ann B},
  journal={IEEE transactions on pattern analysis and machine intelligence},
  volume={28},
  number={9},
  pages={1393--1403},
  year={2006},
  publisher={IEEE}
}

@article{mhaskar2025active,
  title={Active Learning Classification from a Signal Separation Perspective},
  author={Mhaskar, Hrushikesh Narhar and O'Dowd, Ryan and Tsoukanis, Efstratios},
  journal={arXiv preprint arXiv:2502.16425},
  year={2025}
}

@article{siegel2023optimal,
  title={Optimal approximation rates for deep relu neural networks on sobolev and besov spaces},
  author={Siegel, Jonathan W},
  journal={Journal of Machine Learning Research},
  volume={24},
  number={357},
  pages={1--52},
  year={2023}
}

@article{mhaskar2024robust,
  title={Robust and tractable multidimensional exponential analysis},
  author={Mhaskar, Hrushikesh Narhar and Kitimoon, S and Raj, Raghu G},
  journal={arXiv preprint arXiv:2404.11004},
  year={2024}
}

@article {Bel_Niy_2007,
    AUTHOR = {Belkin, Mikhail and Niyogi, Partha},
     TITLE = {Convergence of Laplacian eigenmaps},
   JOURNAL = {Adv. neur. inform. process. 19},
  FJOURNAL = {Advances in neural information processing 19},
    VOLUME = {129},
      YEAR = {2007},
     PAGES = {},
      ISSN = {},
   MRCLASS = {68T05 (62H30)},
  MRNUMBER = {2274444},
}

@article{filom2020pde,
  title={PDE constraints on smooth hierarchical functions computed by
neural networks},
  author={Filom, Khashayar and Kording, Konrad Paul and Farhoodi, Roozbeh},
  journal={arXiv preprint arXiv:2005.08859},
  year={2020}
}

@article{kohler2021rate,
  title={On the rate of convergence of fully connected deep neural
network regression estimates},
  author={Kohler, Michael and Langer, Sophie},
  journal={The Annals of Statistics},
  volume={49},
  number={4},
  pages={2231--2249},
  year={2021},
  publisher={JSTOR}
}

@article{kohler2022estimation,
  title={Estimation of a function of low local dimensionality by deep
neural networks},
  author={Kohler, Michael and Krzy{\.z}ak, Adam and Langer, Sophie},
  journal={IEEE transactions on information theory},
  volume={68},
  number={6},
  pages={4032--4042},
  year={2022},
  publisher={IEEE}
}

@article{kohler2023estimation,
  title={Estimation of a regression function on a manifold by fully
connected deep neural networks},
  author={Kohler, Michael and Langer, Sophie and Reif, Ulrich},
  journal={Journal of Statistical Planning and Inference},
  volume={222},
  pages={160--181},
  year={2023},
  publisher={Elsevier}
}
\end{document}